%% file: elsarticle-template.tex
\documentclass[3p,twocolumn,authoryear]{elsarticle}

\usepackage{lineno,hyperref}

\usepackage{multirow}
\usepackage{amsmath}
\usepackage{amssymb}
\usepackage{graphicx}
\usepackage{tabularx}
\usepackage{booktabs}
\usepackage{microtype}
\usepackage{tikz,pgfplots}
\usepackage{siunitx}
\usepackage{algorithm}
\usepackage[babel]{csquotes}
\usepackage{algpseudocode}
\usetikzlibrary{spy}
\usepackage{array}
\usepackage{subfig}
\definecolor{mycolor2}{rgb}{0.00000,1.00000,1.00000}%

\newlength \figureheight
\newlength \figurewidth
\def\argmin{\mathop{\mathrm{argmin}}}

\newcommand{\ie}{i.\,e.\,\,}
\newcommand{\wrt}{w.r.t.\,\,}
\newcommand{\sref}[1]{Section~\ref{#1}}
\renewcommand{\vec}[1]{\mathbf{#1}}

\newcolumntype{R}[1]{>{\raggedleft\arraybackslash\hspace{0pt}}p{#1}}
\newcolumntype{L}[1]{>{\raggedright\arraybackslash\hspace{0pt}}p{#1}}
\newcolumntype{C}[1]{>{\centering\hspace{0pt}}p{#1}}

\definecolor{gray4}{rgb}{0.2431,0.2431,0.2431}
\definecolor{gray3}{rgb}{0.3098,0.3098,0.3098}
\definecolor{gray2}{rgb}{0.4863,0.4863,0.4863}
\definecolor{gray1}{rgb}{0.7725,0.7725,0.7725}

\journal{Medical Image Analysis}

\usepackage{numcompress}\biboptions{authoryear}

\begin{document}

\begin{frontmatter}

\title{Temporal and Volumetric Denoising via Quantile Sparse Image Prior}

\author{Franziska~Schirrmacher$^{1,*}$, Thomas~K\"ohler$^{1,4,*}$, J{\"u}rgen Endres$^{1,2}$, Tobias Lindenberger$^{1}$, Lennart~Husvogt$^{1}$, James~G.~Fujimoto$^{3}$, Joachim~Hornegger$^{1}$, Arnd D{\"o}rfler$^{2}$, Philip Hoelter$^{2}$, and~Andreas~K.~Maier$^{1}$}
\address{$^{1}$~Pattern Recognition Lab, Friedrich-Alexander-Universit{\"a}t Erlangen-N{\"u}rnberg, Germany

$^{2}$Department of Neuroradiology, Universit{\"a}tsklinikum Erlangen, Germany 

$^{3}$~Research Laboratory of Electronics, Massachusetts Institute of Technology, Cambridge, USA 

$^{4}$e.solutions GmbH, Erlangen, Germany
\\[0.2cm]
$^*$ These authors contributed equally to this work.}

\begin{abstract}
This paper introduces an universal and structure-preserving regularization term, called quantile sparse image (QuaSI) prior. The prior is suitable for denoising images from various medical imaging modalities. We demonstrate its effectiveness on volumetric optical coherence tomography (OCT) and computed tomography (CT) data, which show different noise and image characteristics. OCT offers high-resolution scans of the human retina but is inherently impaired by speckle noise. CT on the other hand has a lower resolution and shows high-frequency noise. For the purpose of denoising, we propose a variational framework based on the QuaSI prior and a Huber data fidelity model that can handle 3-D and 3-D+t data. Efficient optimization is facilitated through the use of an alternating direction method of multipliers (ADMM) scheme and the linearization of the quantile filter. Experiments on multiple datasets emphasize the excellent performance of the proposed method.

\end{abstract}


\end{frontmatter}


\section{Introduction}
\label{sec:introduction}

The reliable reduction of image noise poses a constantly recurring problem in today’s imaging systems. In healthcare, noise may limit the reliability of medical image data for subsequent clinical workflows. For instance, in radiology using computed tomography (CT) or related morphological imaging modalities, noise affects the analysis of anatomical structures and thus impedes diagnostic applications. In optical coherence tomography (OCT) for retinal imaging as another example use case, noise limits the measurement of structural features in the human eye, e.\,g. retinal layer properties. Apart from diagnostic applications, noise reduction is also a major theme for different interventional imaging modalities like fluoroscopically guided procedures. Low dose radiation exposure for patient safety leads to noisy and low-contrast fluroscopic sequences \citep{Amiot2016}.

To mitigate these limitations, \textit{denoising} can be either implemented by means of customized hardware or via postprocessing of captured image data. While hardware-based denoising often leads to increased system complexities, image-based postprocessing facilitates denoising in a cost-effective way using computational methods. Despite the great progress in developing general denoising schemes for natural images, adopting them for medical data poses several challenges. First and foremost, there is a narrow ridge between achieving sufficient noise reduction and unwanted distortions of meaningful medical structures. Moreover, noise distributions in medical data often deviate from the commonly employed models for natural images like additive, white Gaussian noise (AWGN). For example, noise can follow multiplicative models or structured patterns related to acquisition parameters like in CT. General denoising algorithms have been mainly developed for 2-D data, e.\,g. color photographs, but denoising in medical imaging also needs to handle time-resolved and/or volumetric data. These requirements desire enhanced and robust denoising methods to be applicable within medical workflows.

\begin{figure*}[!tp]
\scriptsize
\centering
\includegraphics[width = 0.93\textwidth]{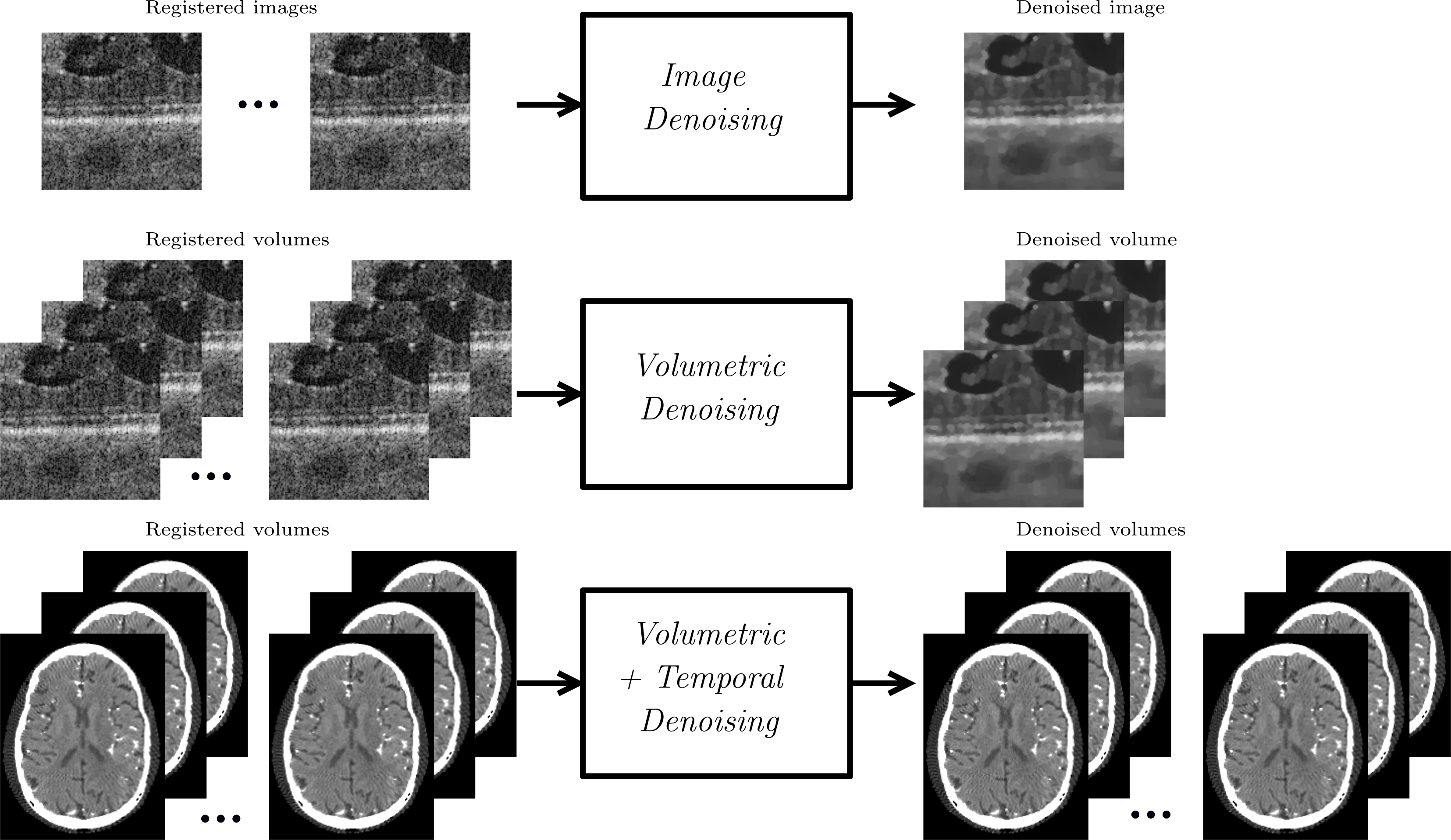}
\caption{We propose three modi of our spatio-temporal denoising algorithm. In the first modus (top), hereinafter called \textit{image denoising}, single images or a sequence of registered images are processed. The second modus (middle) processes volumes as well as a sequence of registered volumes and is called \textit{volumetric denoising}. The third modus (bottom) processes volumes as well as a sequence of registered volumes, outputs a sequence of volumes, and is called \textit{volumetric + temporal denoising}.}
\label{fig:graphicalAbstract}
\end{figure*}

In this paper, we propose denoising for medical image data within a variational framework. As the key contribution, we introduce the class of \textit{quantile sparse image} (QuaSI) priors to model the appearance of noise-free medical data. Specifically, we propose a median filter based regularizer that is based on the QuaSI prior using the 0.5 quantile. This follows the idea that noise-free data should be a fixed point of the median filter and we show that this model facilitates structure-preserving denoising. To approach the resulting non-linear and non-convex optimization problem, we present an alternating direction method of multipliers (ADMM) scheme. Our algorithm can handle \textit{spatio-temporal} denoising by processing either single images or sequences of consecutive images. Furthermore, it enables denoising of volumetric data. Thus, it can be adjusted to the clinical needs within a target application. 

This paper is an extension of our prior work in \cite{Schirrmacher2017} and makes the following additional contributions:
\begin{itemize}
	\item The algorithm as well as the QuaSI prior are extended to process volumetric medical data. 
	\item An investigation of the convergence and parameter sensitivity of our algorithm is conducted.	
	\item An extension of our algorithm is presented to process volumetric data in C-arm CT imaging.
\end{itemize}

The remainder of this paper is organized as follows. In \sref{sec:relatedWork}, we review related work on spatial and temporal denoising. \sref{sec:background} comprises the objective function of the energy minimization problem. In \sref{sec:quantileSparseImagePrior} the QuaSI prior is introduced. The numerical optimization of our denoising framework is derived in \sref{sec:deployingQuaSIForOCTDenoising}. In \sref{sec:experimentsAndResults}, an experimental evaluation of our method on publicly available benchmark data, clinical OCT scans as well as CT data is reported. Finally, section~\ref{sec:conclusion} contains our conclusion. 

\begin{figure}[!tb]
	\begin{center}
		\includegraphics[width = 0.45\textwidth]{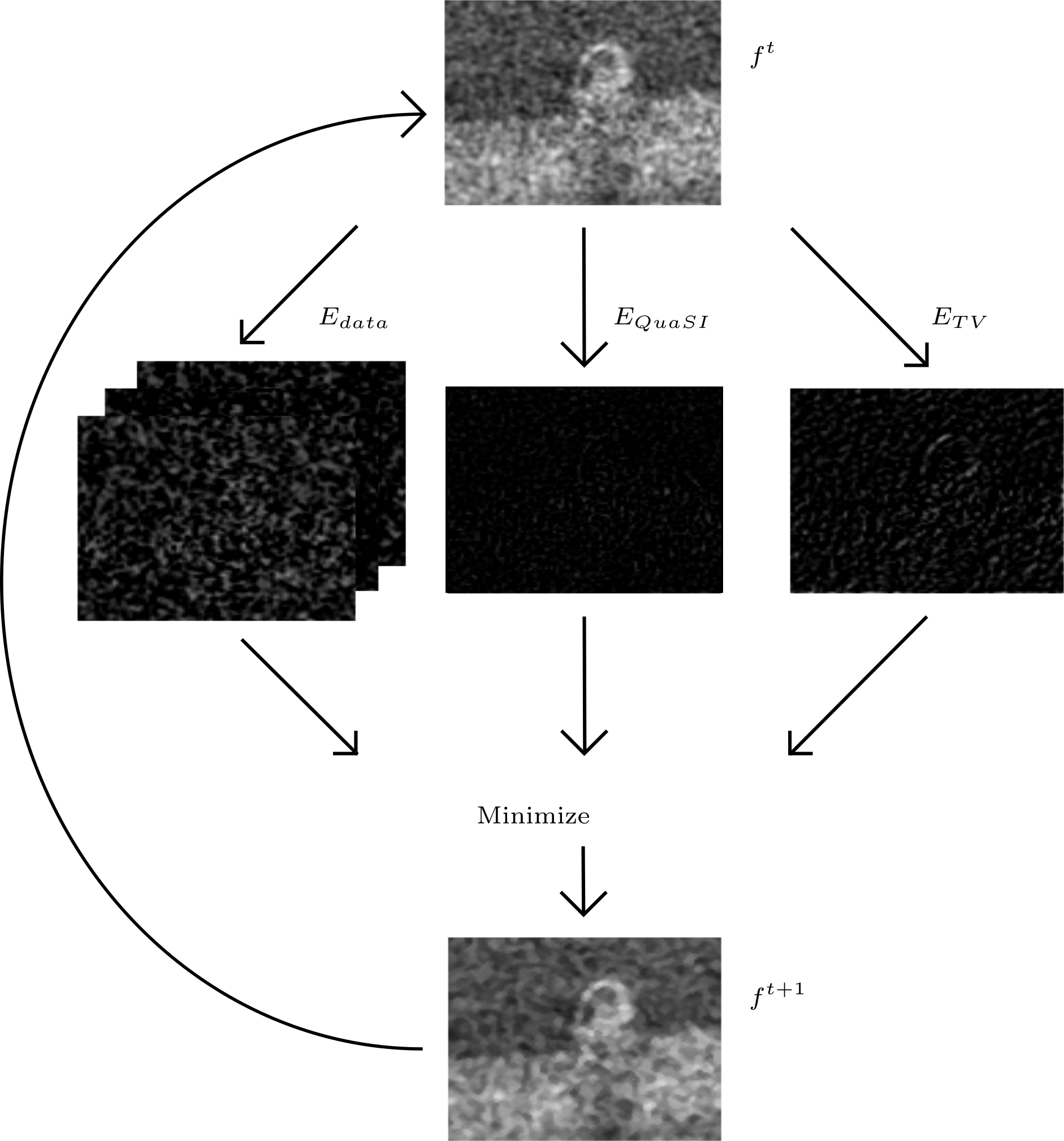}
	\end{center}
	\caption{Method overview: The proposed spatio-temporal denoising algorithm 
		is based on an energy minimization formulation with three terms.}
	\label{fig:methodOverview}
\end{figure}

\section{Related Work}
\label{sec:relatedWork}

The image-based denoising techniques can be divided into two groups.  

\subsection{Spatial Denoising Methods}
\label{sec:spatialDenoisingMethods}

\textit{Spatial} or \textit{single-image} denoising has been extensively studied in the image processing community and various approaches emerged over the past decades. Local image filters perform smoothing of noisy images possibly in an adaptive way to preserve image structures \citep{Tomasi1998}. Non-local filtering also exploits the statistics of similar and repeating patches within images. One representative from this class is the successful BM3D method by \cite{Dabov2007}. However, these methods have been mainly designed for natural images under simplified assumptions like additive white Gaussian noise, which is inappropriate to describe speckle noise that is multiplicative in nature. Learning-based denoising, e.\,g. based on multilayer neural networks \citep{Burger2012}, hold the potential to handle speckle noise by learning noise distributions from training data. However, large-scale training data required for such methods is barely available for OCT.

Some spatial filters that have been adopted for OCT denoising are the hybrid median filter, Lee filter, Wiener filter, or wavelet thresholding as investigated by \cite{Ozcan2007}. Global denoising methods for OCT have been introduced by \cite{Salinas2007} using non-linear diffusion and later by \cite{Duan2016} using second-order total generalized variation. \cite{Wong2010} have proposed structure-adaptive Bayesian estimation to handle speckle noise. One interesting approach has been proposed by \cite{Fang2012}, where dictionary learning based on B-scans with high signal-to-noise ratio (SNR) is used to denoise low SNR B-sans.

Single-image denoising offers great flexibility in clinical applications of OCT as few assumptions on the scanning protocol are made. However, the noise reduction is limited as such methods can utilize single B-scans only.

\subsection{Temporal Denoising Methods}
\label{sec:temporalDenoisingMethods}

\textit{Temporal} or \textit{multi-image} denoising methods consider coherence of consecutive images to improve noise reduction over single-image denoising. Such methods have been widely investigated for OCT and exploit sets of B-scans that are acquired sequentially from the same location or nearby positions. A popular approach in commercial systems is to register multiple of these B-scans and to average the registered scans to cancel out random noise. Averaging is computationally efficient but requires many repetitive acquisitions to effectively reduce speckle noise. \cite{Mayer2012} enhance simple averaging based on wavelet decompositions of B-scans to estimate local image structures and noise. Denoising is conducted in the wavelet domain by weighted averaging of wavelet coefficients according to the local image structure. \cite{Cheng2014} formulate OCT denoising from multiple scans as a low-rank matrix completion problem. \cite{Thapa2015} follow a similar notion and exploit the low-rank property on a patch-based level of multiple B-scans using weighted nuclear norm minimization. \cite{Bian2015} have proposed inter-frame and intra-frame priors for denoising using convex optimization. BM4D is an extension of the popular BM3D method to process volumetric data \citep{maggioni2013nonlocal}.

All of these multi-image methods have in common that they require multiple input scans. This increases the overall acquisition time and therefore might lead to a higher patient discomfort. Also, they perform denoising on a B-scan level but ignore coherence of nearby B-scans within volumetric OCT data. If denoising of entire volumes is desired, simple consecutive processing of individual B-scans can lead to suboptimal results. In this paper, we mitigate both limitations by proposing a unified approach to handle denoising on a B-scan or volumetric level based on single or multiple scans. 

\section{Background}
\label{sec:background}

This section presents the variational framework for denoising volumetric data. Figure~\ref{fig:graphicalAbstract} illustrates three modi of this framework, namely image denoising, volumetric denoising, and volumetric+temporal denoising. The pipelines differ in the number of outputs and are therefore divided into multiple-input single-output (MISO) denoising and multiple-input multiple-output (MIMO) denoising.

Throughout this paper, we use the following nomenclature. We denote a volume as a vector $\vec{g} \in \mathbb{R}^{N_z N_{xy}}$ composed of $N_z$ images $\vec{g}_z$, $z = 1, \ldots, N_z $ of size $N_{xy} = N_x N_y$ pixels. For the sake of convenience, 2-D images of size $N_x \times N_y$ are reshaped to vector notation using a row-wise scanning. A sequence of volumes is denoted as vector $\vec{G} \in \mathbb{R}^{N_t  N_z  N_{xy}}$, where $N_t$ is the number of volumes in the sequence. The input to the proposed framework is a sequence of $T$ volumes, where $1 \leq T \leq N_t$. For volumetric as well as volumetric+temporal denoising, we employ $Z$ consecutive images per volume ($1 < Z \leq N_z$), while image denoising is based on a single image in each volume ($Z = 1$).


\subsection{Noise Model}

In this paper, we consider several denoising applications with two different underlying noise models. In an \textit{additive} noise model, a noise-free volume $\vec{f} = (\vec{f}_1, \ldots, \vec{f}_Z)^\top$ is related to a noisy volume $\vec{g} = (\vec{g}_1, \ldots, \vec{g}_Z)^\top$ according to:
\begin{equation}
	\label{eqn:noiseAdd}
	\vec{g} = \vec{f} + \vec{n},
\end{equation}
where $\vec{n} = (\vec{n}_1, \ldots, \vec{n}_Z)^\top$ denotes an additive noise term. Common instances of this model are AWGN with stationary distribution of $\vec{n}$ or Poisson noise, where the variance of $\vec{n}$ depends on the measured image data.

In a \textit{multiplicative} noise model, each captured volume $\vec{g}$ is related to a respective noise-free volume $\vec{f}$ according to:
\begin{equation}
	\label{eqn:noiseMult}
	\vec{g} = \vec{f} \odot \vec{n},
\end{equation}
where $\odot$ is the Hadamard (element-wise) product. We can turn the multiplicative model in \eqref{eqn:noiseMult} to the additive one in \eqref{eqn:noiseAdd} by transforming it to a logarithmic measurement domain. One common instance of this model is speckle noise that appears in OCT imaging \citep{Wong2010,Duan2016}.

\subsection{Energy Minimization Formulation}

Given a sequence of $T$ volumes $\vec{g}^{(t)}$ with $t = 1, \ldots, T$ that are either captured from the same position or from nearby positions and registered to each other, we propose MIMO and MISO denoising. 

In MISO denoising, we aim at estimating one noise-free volume $\hat{\vec{f}}$. We formulate denoising as the minimization of the objective function:
\begin{equation}
	\begin{split}
		\hat{\vec{f}} &= \argmin_{\vec{f}} \sum_{t = 1}^{T}  \rho \big( \vec{f} - \vec{g}^{(t)} \big)
		+ \lambda R_{\mathrm{QuaSI}}(\vec{f}) + \mu \Vert\nabla \vec{f}\Vert_{1}.
	\end{split}
	\label{eqn:objective}
\end{equation}
The first term in \eqref{eqn:objective} denotes the data fidelity of $\vec{f}$ \wrt the input volumes $\vec{g}^{(t)}$. The second term is the proposed quantile sparse image (QuaSI) prior weighted by $\lambda \geq 0$. The third term denotes anisotropic total variation (TV) weighted by $\mu \geq 0$, which regularizes the spatial gradient $\nabla \vec{f} =~ (\nabla_{x}\vec{f}, \nabla_{y}\vec{f},\nabla_{z}\vec{f})^{\top}$. It is worth noting that the general denoising framework in \eqref{eqn:objective} can handle both noise reduction for entire volumes in 3-D as well as for individual images in 2-D by constraining the domain of both regularization terms.

MIMO denoising follows a similar approach but aims at estimating a sequence of volumes $\hat{\vec{F}}$. We formulate MIMO denoising as the minimization of the objective function:
\begin{equation}
	\begin{split}
		\hat{\vec{F}} = \argmin_{\vec{F}} &\rho \big( \vec{F} - \vec{G} \big)
		+ \lambda R_{\mathrm{QuaSI}}(\vec{F}) + \mu \Vert\nabla \vec{F}\Vert_{1}\\
		&+ \omega \Vert \nabla_t \vec{F} \Vert_1,
	\end{split}
	\label{eqn:objective3Dt}
\end{equation}
where $\nabla_{t} \vec{F}$ denotes the gradient of $\vec{F}$ in temporal direction and the associated TV regularization is weighted by $\omega \geq 0$.

In \eqref{eqn:objective} and \eqref{eqn:objective3Dt}, the data fidelity terms use the loss function $\rho: \mathbb{R}^N \rightarrow \mathbb{R}_0^+$ to formulate the image formation. In general, the image formation needs to consider a mixture of noise, potential misalignments between the input volumes, or motion artifacts. Following prior work on mixed noise models in image restoration \citep{Kohler2015c}, we propose to use the Huber loss \citep{Ochs2015}: 
\begin{equation}
	\rho(\vec{l}) = \sum_{i = 1}^N \phi(l_i),  
\end{equation}
where:
\begin{equation}
	\phi(l) =
	\begin{cases}
		 \frac{1}{2} l^2 & \text{if}~ l \leq \epsilon\\
		 \epsilon \left( |l| - \frac{1}{2}\epsilon \right) & \text{otherwise},
	\end{cases}
\end{equation}
and $\epsilon > 0$ denotes the threshold of the Huber loss. This leads to an outlier-insensitive model while the underlying data fidelity is a convex term.

\begin{figure*}[!tb]
  \scriptsize
  \centering
  \setlength\figurewidth{0.37\textwidth}
  \setlength\figureheight{0.5\figurewidth}
  
    \subfloat[Noisy B-scan $\vec{f}_{noisy}$ ]{
     \includegraphics[width = 0.37\textwidth]{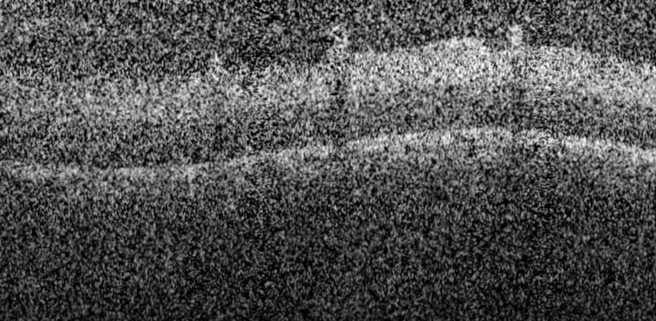}
     \label{fig:visualizeNoisy}
   }
  \qquad
   \subfloat[Gold standard B-scan $\vec{f}_{gold}$ ]{
     \includegraphics[width = 0.37\textwidth]{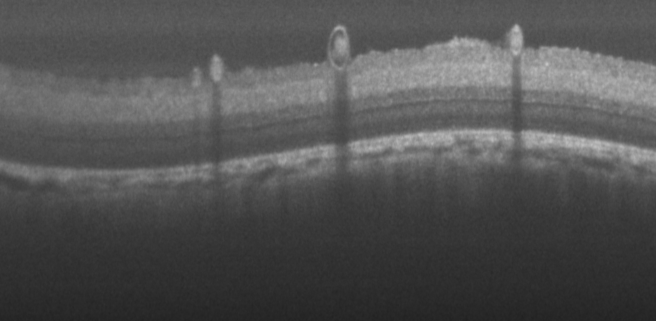}
     \label{fig:visualizeGold}
   } 
  
  \subfloat[$\vec{r}_{noisy}$]{
     \includegraphics[width = 0.37\textwidth]{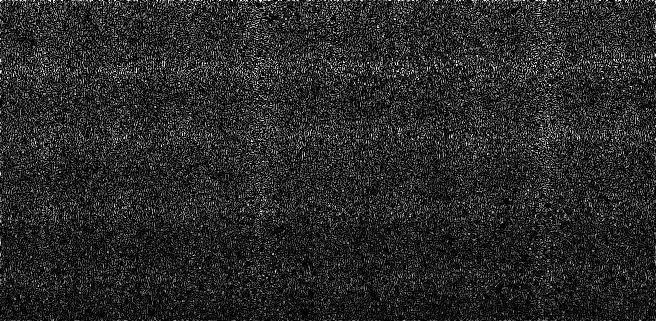}
     \label{fig:visualizeQuaSIInitial}
   }
  \qquad
   \subfloat[$\vec{r}_{gold}$]{
     \includegraphics[width = 0.37\textwidth]{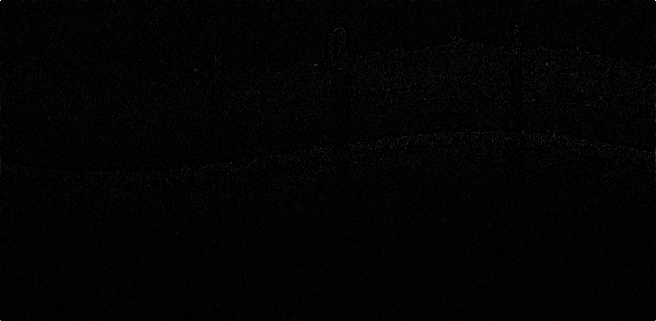}
     \label{fig:visualizeQuaSIGold}
   } 
  	
  \subfloat[Histogram of $\vec{r}_{noisy} $]{
    \input{images/QuaSIInitialHisto.tikz}
    \label{fig:histogramQuaSIInitial}
  }
  \qquad  
  \subfloat[Histogram of $\vec{r}_{gold} $]{
    \input{images/QuaSIGoldHisto.tikz}
     \label{fig:histogramQuaSIGold}
  }
  \caption{Analysis of our proposed QuaSI prior using median filtering $Q(\cdot)$ to model the appearance of OCT B-scans. (a) and (b) depict a noisy B-scan along with the respective gold standard taken from the pig eye dataset \cite{Mayer2012}. (c) and (d) show the residual $\vec{r} = \vec{f} - Q(\vec{f})$ of the QuaSI regularization term, where brighter pixels express higher residuals (contrast enhanced for visualization). (e) and (f) depict the corresponding histograms of the both residuals, where the histogram for the gold standard is sparse. Our QuaSI prior exploits the sparsity of $\vec{r} = \vec{f} - Q(\vec{f})$ for regularization in our variational denoising framework.}
  \label{fig:histogram}
\end{figure*}

\begin{figure*}[!tb]
	\centering
	
	\includegraphics[width = 0.8\textwidth]{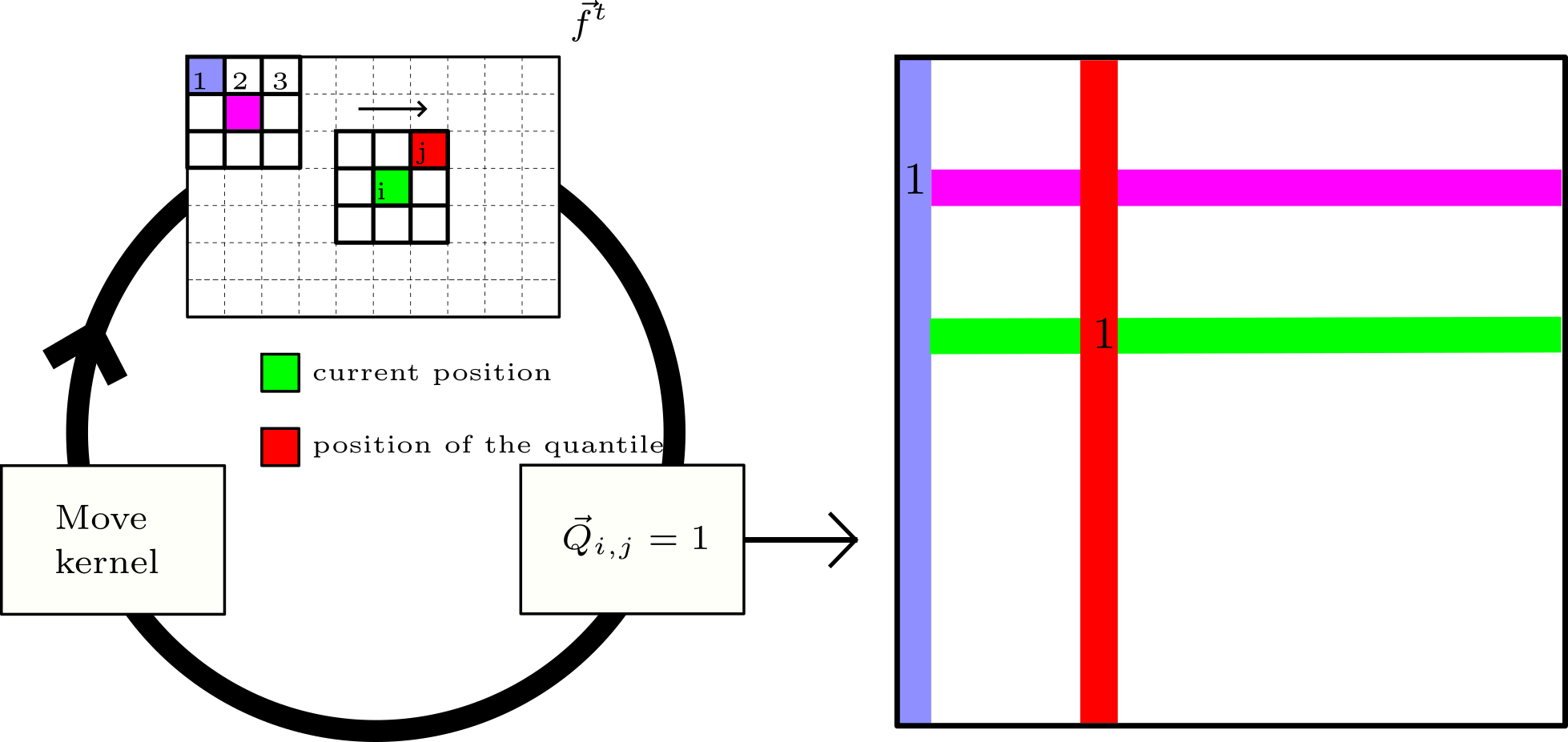}
	\caption{Construction of the binary matrix to approximate the quantile filter $Q(\vec{f}) = \vec{Q}\vec{f}^t$.}
	\label{fig:lookUp}
\end{figure*}
\section{Quantile Sparse Image (QuaSI) Prior}
\label{sec:quantileSparseImagePrior}

A robust and efficient regularization term is of importance to achieve results with a high signal-to-noise ratio (SNR). The better the regularization term is able to model natural or medical images, the better the result of the optimization. Structure preservation is a sensitive issue when dealing with medical data. The images might contain small morphological structures that need to be preserved for the purpose of diagnosis. In order to tackle the challenges referred to above, the so called quantile sparse image (QuaSI) prior is introduced. 

\subsection{Definition of the Prior}

The QuaSI prior is based on quantile filtering, where the quantile filter is denoted as $\tilde{\vec{f}} = Q(\vec{f})$. The $p$-quantile with $p \in [0, 1]$ is determined within a local neighborhood $\mathcal{N}(i)$. The local neighborhood consists of $d^3$ voxel, where $d$ denotes the width of the cubic filter kernel. For the $i$-th voxel in $\vec{f}$ we filter according to $\tilde{f}_i = \mathrm{quantile}_{\mathcal{N}(i)} (f_i, p)$. Inspired by the regularization by denoising priors by \cite{Romano2016a}, the denoised volume is a fixed point under the quantile filter. In this way:
\begin{equation}
	R_{\mathrm{QuaSI}}(\vec{f}) = ~\left|\left| \vec{f} - Q(\vec{f}) \right|\right|_1.
	\label{eqn:quasiPrior}
\end{equation}
Specifically, regularization according to \eqref{eqn:quasiPrior} enforces sparsity of the residual $\vec{f} - Q(\vec{f})$. This offers a general model for regularization and -- depending on the application -- various types of statistics can be chosen for $Q(\vec{f})$. In this paper, we propose the median filter, where $\tilde{f}_i = \mathrm{median}_{\mathcal{N}(i)} (f_i)$. This follows the rationale that median filtering facilitates structure-preserving denoising under non-Gaussian noise. Further applications including erosion and dilation are not covered in this paper. In the literature \citep{Rohkohl2011}, quantiles are used to obtain a reference image to estimate non-periodic motion. Those examples are suitable applications that the QuaSI prior can handle.

To validate the QuaSI prior using median filter regularization for denoising, we study its behavior under real measurement noise. For this purpose, we use the publicly available pig eye dataset by \cite{Mayer2012}, which provides a gold standard OCT B-scan obtained from the average of 455 registered noisy OCT B-scans. We compare a noisy OCT B-scan $\vec{f}_{noisy}$ with the gold standard $\vec{f}_{gold}$ in Fig.~\ref{fig:visualizeNoisy} and Fig.~\ref{fig:visualizeGold}. The residuals $\vec{r} = \vec{f} - \textit{Q}(\vec{f})$ of the QuaSI regularization term are illustrated in Fig.~\ref{fig:visualizeQuaSIInitial} for the noisy B-scan and in Fig.~\ref{fig:visualizeQuaSIGold} for the gold standard. Compared to the gold standard, the noisy B-scan yields a less sparse signal as shown in the histograms of both residuals in Fig.~\ref{fig:histogramQuaSIInitial} and Fig.~\ref{fig:histogramQuaSIGold}. Notice that the QuaSI regularization does not penalize image discontinuities. The histogram using the noisy B-scan contains less zero elements, while the histogram for the gold standard is sparse. Our proposed QuaSI prior exploits these observations for structure-preserving regularization in our variational denoising framework.

\subsection{Linearization}

In order to deal with the non-linearity of the quantile operator $\textit{Q}(\vec{f})$ the linearization $\textit{Q}(\vec{f}) = \vec{Q}\vec{f}$, similar to the work of \cite{Pan2016}, is performed.  The binary matrix $\vec{Q}$ is assembled element-wise according to:

\begin{equation}
Q_{ij} =~
\begin{cases}
1 & \mathrm{if} \  j=q, \\
0 & \mathrm{otherwise},
\end{cases}
\label{eqn:linearizationQuasi}
\end{equation}
 where $q = \arg\mathrm{quantile}_{r \in \mathcal{N}(i)} f_r$. This operation filters the $i$-th pixel according to the $p$-quantile in its local neighborhood $\mathcal{N}(i)$. For $\vec{f}^\prime = \vec{f}$ the linearization fullfills $\textit{Q}(\vec{f}^\prime) = \vec{Q} \vec{f}^\prime$, while otherwise $\vec{Q}$ serves as an approximation of the quantile filter.
 
Figure~\ref{fig:lookUp} illustrates the construction of the binary matrix $\vec{Q}$ in 2-D. Each pixel is replaced by the quantile within its local neighborhood. The position of the quantile is stored in the binary matrix. In this example, the quantile is at position $j$. Thus, the $i$-th row of the matrix contains a one in the $j$-th column and zeros otherwise. The multiplication $\vec{Q}\vec{f}$ yields the quantile filtered result.

\section{Deploying QuaSI for Denoising}
\label{sec:deployingQuaSIForOCTDenoising}

In this section, we show how the proposed QuaSI prior can be deployed for volumetric and temporal denoising. We derive two numerical optimization algorithms for denoising based on a MISO and a MIMO mode.

\subsection{Multiple-Input Single-Output (MISO) Mode}
\label{sec:MISO}

MISO denoising in our framework is based on the energy minimization formulation in \eqref{eqn:objective}. In order to handle the non-smooth $L_1$ norm terms, we adopt ADMM optimization \citep{Goldstein2009}. To this end, \eqref{eqn:objective} is reformulated to the constrained optimization problem:
\begin{equation}
\begin{split}
	\hat{\vec{f}} =~ &\argmin_{\vec{f}} \sum_{t = 1}^{T}  \rho \big( \vec{f} - \vec{g}^{(t)} \big)  + \lambda \Vert \vec{u} \Vert_1 + \mu \Vert\vec{v}\Vert_{1} \\
	& \mathrm{\ such\ that\ } \ \vec{u} = ~\vec{f} - \textit{Q}(\vec{f}),~\vec{v} = ~ \nabla \vec{f},
\end{split}
\label{eqn:forCT}
\end{equation}
where $\vec{u}$ and $\vec{v}$ are auxiliary variables. Then, an unconstrained optimization problem is obtained from \eqref{eqn:forCT} using quadratic penalty functions according to:
\begin{equation}
\begin{split}
	\hat{\vec{f}} = ~&\argmin_{\vec{f}} \sum_{t = 1}^{T}  \rho \big( \vec{f} - \vec{g}^{(t)} \big)
	+ \mu\Vert \vec{v} \Vert_{1} + \lambda\Vert \vec{u} \Vert_{1} \\
	 & + \dfrac{\alpha}{2} \Vert \vec{u} - \vec{f} + \textit{Q}(\vec{f})\Vert_{2}^{2} 
		 + \dfrac{\beta}{2} \Vert \vec{v} - \nabla \vec{f} \Vert_{2}^{2}.
\end{split}
\label{eqn:forCT2}
\end{equation}

The Lagrangian multipliers $\alpha > 0$ and $\beta > 0$ enforce the constraints $\vec{u} = f - \textit{Q}(\vec{f})$ and $\vec{v} = \nabla \vec{f}$. If $\alpha,~\beta \rightarrow \infty$, we end up at the original problem \eqref{eqn:objective}. In order to strictly enforce the constraint, the Bregman variables $\vec{b}_u$ and $\vec{b}_v$ are introduced. Then, we minimize the augmented Lagrangian:
\begin{equation}
	\begin{split}
		&\mathcal{L}_{\mathrm{AL}}(\vec{f}, \vec{u}, \vec{v}, \vec{b}_u, \vec{b}_v) 
		= ~\sum_{t = 1}^{T}  \rho \big( \vec{f} - \vec{g}^{(t)} \big)\\
		& + \dfrac{\alpha}{2} \Vert \vec{u} - \vec{f} + \textit{Q}(\vec{f}) - \vec{b}_{u} \Vert_{2}^{2} + \lambda\Vert \vec{u} \Vert_{1}\\ 
		 &+ \dfrac{\beta}{2} \Vert \vec{v} - \nabla \vec{f} - \vec{b}_v\Vert_{2}^{2} + \mu\Vert \vec{v} \Vert_{1}.
	\end{split}
	\label{eqn:augmentedLagrangian}
\end{equation}
We iteratively optimize \eqref{eqn:augmentedLagrangian} by alternating minimization \wrt the individual parameters. Hence, three subproblems emerge, where the $L_1$-Norm is decoupled from the $L_2$-Norm.

The minimization of the augmented Lagrangian \eqref{eqn:augmentedLagrangian} \wrt $\vec{f}$ can be solved in a least square sense. Therefore, the binary matrix $\vec{Q}$ is constructed using the result $\vec{f}^k$ from the previous iteration, where $k$ denotes the iteration index. In order to cope with the Huber loss, iteratively re-weighted least squares (IRLS) is applied. Solving the resulting least squares problem leads to the linear system:
\begin{align}
	\vec{A}\vec{f}^{k+1} &= \vec{b} 
		\label{eqn:cgEquationSystem}\\
	\vec{A} &= \sum_{t=1}^{T} \vec{W}^{(t)}  + \beta  \nabla^{\top}\nabla + \alpha \vec{M}^\top \vec{M} 
		\label{eqn:cgEquationSystem:matA}\\
	\begin{split}
		\vec{b} &= \sum_{t=1}^{T} \vec{W}^{(t)} \vec{g}^{(t)}\\
		&+ \beta \nabla^\top( \vec{v} - \vec{b}_{v}) + \alpha \vec{M}^{\top}( \vec{u}- \vec{b}_{u}),
	\end{split}
		\label{eqn:cgEquationSystem:vecB}
\end{align}
where $\vec{M} = \vec{I} - \vec{Q}$ with the identity matrix $\vec{I}$. In \eqref{eqn:cgEquationSystem} - \eqref{eqn:cgEquationSystem:vecB}, $\vec{W}^{(t)}$ are diagonal weight matrices constructed from $\vec{f}^k$. Using the intermediate result $\vec{f}^{k}$, we can compute the weights for IRLS according to:
\begin{equation}
	W_{ii}^{(t)} = ~ \dfrac{\phi^\prime \left(f^k_i - g_i^{(t)} \right)}{\left|f^k_i - g_i^{(t)}\right|}, \label{eqn:huberWeight}
\end{equation}
where $\phi^\prime(l)$ is the derivative of the Huber loss. The threshold of the Huber loss is set to $\epsilon = 1.345\sigma$ to achieve a 95-percent efficiency of the estimator under Gaussian noise \citep{Ochs2015}. We use the median absolute deviation (MAD) rule to obtain a consistent estimate of the standard deviation according to $\sigma = 1.4826 \cdot \mathrm{MAD}(f^k_i - g_i^{(t)})$ \citep{Rousseeuw1987}. To solve the linear system \eqref{eqn:cgEquationSystem}, conjugate gradient (CG) iterations are used. 

The minimization of the augmented Lagrangian \eqref{eqn:augmentedLagrangian} \wrt the auxiliary variables can be done by exploiting the separability of the problem. Given the estimate for the intermediate result $\vec{f}^{k+1}$, this leads to the element-wise updates:
\begin{align}
	u_i^{k+1} = ~ \mathrm{shrink} (&[\vec{f}^{k+1} - \vec{Q} \vec{f}^{k+1} + \vec{b}^{k}_{u}]_i, \lambda/\alpha) \label{eqn:updateAuxU},\\	
	v_i^{k+1} = ~\mathrm{shrink} (&[\nabla \vec{f}^{k+1} + \vec{b}^{k}_v]_i, \mu / \beta ).\label{eqn:updateAuxV},	
\end{align}
where $\mathrm{shrink}(z, \gamma) = \mathrm{sign}(z) \max(z - \gamma, 0)$ denotes the shrinkage operator \citep{Goldstein2009}. 

Given an estimate for the intermediate result $\vec{f}^{k+1}$ as well as the auxiliary variables $\vec{u}^{k+1}$ and $\vec{v}^{k+1}$, the Bregman variables are updated according to:
\begin{align}
	\vec{b}^{k+1}_u &=~ \vec{b}^{k}_u + (\vec{f}^{k+1} - \vec{Q} \vec{f}^{k+1} - \vec{u}^{k+1}), \label{eqn:updateBregmanU}\\
	\vec{b}^{k+1}_v &=~ \vec{b}_v^{k} + (\nabla\vec{f}^{k+1} - \vec{v}^{k+1}).
	\label{eqn:updateBregmanV}
\end{align}

Algorithm \ref{alg:denoisingAlgorithm2D} summarizes the proposed ADMM based iteration scheme. Overall, we use two nested optimization loops to solve \eqref{eqn:forCT}. We use the mean of the input images as an initial guess \smash{$\vec{f}^1$} as well as \smash{$\vec{u}^1 = \vec{v}^1 = \vec{0}$, $\vec{b}_u^1 = \vec{b}_v^1 = \vec{0}$}. The weight matrices for IRLS are updated at every iteration. 

The linearization $\vec{Q}$ of the quantile filter is updated every $K_{\text{inner}}$ iterations, assuming the position of the quantile does not change within the next $K_{\text{inner}}$ iterations.  This assumption speeds up the algorithm, as the construction of the matrix is time-consuming. Note that $K_{\text{inner}}$ should not be chosen too large in order to avoid a bad approximation of the quantile filter. A proper evaluation of the convergence of the algorithm is presented in Sect.~\ref{sec:convergenceAndParameterSensitivity}.

\begin{algorithm}[!t]
	\caption{MISO denoising with QuaSI prior}
	\label{alg:denoisingAlgorithm2D}
	\begin{algorithmic}
		\State Set $\vec{u}^1 = \vec{v}^1 = \vec{b}_u^1 = \vec{b}_v^1 = \vec{0}$, $\vec{f}^1 = \frac{1}{T} \sum_{t = 1}^T \vec{g}^{(t)}$		
		\For{$k = 1, \ldots, K_{\mathrm{outer}}$}
			\State Assemble $\vec{Q}$ from $\vec{f}^k$ according to \eqref{eqn:linearizationQuasi}
			\For{$i = 1, \ldots, K_{\mathrm{inner}}$}
				\State Update weights $\vec{W}^{(t)}$ using \eqref{eqn:huberWeight}
				\State Update $\vec{f}^{k+1}$ using CG for \eqref{eqn:cgEquationSystem}
				\State Update $\vec{u}^{k+1}$ and $\vec{v}^{k+1}$ using \eqref{eqn:updateAuxU} - \eqref{eqn:updateAuxV}
				\State Update $\vec{b}_u^{k+1}$ and $\vec{b}_v^{k+1}$ using \eqref{eqn:updateBregmanU} - \eqref{eqn:updateBregmanV}
			\EndFor
		\EndFor
	\end{algorithmic}
\end{algorithm}

\begin{algorithm}[!t]
	\caption{MIMO denoising with QuaSI prior}
	\label{alg:MIMODenoising}
	\begin{algorithmic}
		\State Set $\vec{F}^1 = \vec{G}$, $\vec{U}^1 = \vec{V}^1 = \vec{D}^1 = \vec{B}^1_U = \vec{B}^1_V = \vec{B}^1_D = \vec{0}$
		\For{$k = 1, \ldots, K_{\mathrm{outer}}$}
			\State Assemble $\vec{Q}$ from $\vec{F}^{k}$ according to \eqref{eqn:linearizationQuasi}
			\For{$i = 1, \ldots, K_{\mathrm{inner}}$}
				\State Update weights $\vec{W}^{(t)}$ using \eqref{eqn:huberWeight}
				\State Update $\vec{F}^{k+1}$ using CG for \eqref{eqn:cgEquationSystemMIMO}
				\State Update $\vec{U}^{k+1}$, $\vec{V}^{k+1}$, $\vec{D}^{k+1}$ using \eqref{eqn:updateU} - \eqref{eqn:updateD}
				\State Update $\vec{B}_U^{k+1}$, $\vec{B}_V^{k+1}$, $\vec{B}_D^{k+1}$ using \eqref{eqn:updateBU} - \eqref{eqn:updateBD}
			\EndFor
		\EndFor
	\end{algorithmic}
\end{algorithm}

\subsection{Multiple-Input Multiple-Output (MIMO) Mode}
\label{sec:MIMO}

MIMO denoising follows a similar optimization approach and is based on the energy minimization formulation in \eqref{eqn:objective3Dt}. To this end, the augmented Lagrangian is given by:
\begin{align}
	\begin{aligned}
		&\mathcal{L}_{\mathrm{AL}}(\vec{F}, \vec{U}, \vec{V},\vec{D}, \vec{B}_{U}, \vec{B}_{V}, \vec{B}_{D}) = \rho(\vec{F}-\vec{G})\\
		&+ \frac{\alpha}{2} \left| \left| \vec{U} - \vec{F} + Q(\vec{F}) - \vec{B}_{u} \right|\right|_{2}^{2} + \lambda\|\vec{U}\|_{1} \\
		&+ \frac{\beta}{2} \left| \left| \vec{V} - \nabla_{x,y,z}\vec{F} - \vec{B}_V \right|\right|_{2}^{2} + \mu \|\vec{V}\|_{1} \\
		&+ \frac{\gamma}{2} \left| \left| \vec{D} - \nabla_{t}\vec{F} - \vec{B}_{D} \right|\right|_{2}^{2} + \omega \|\vec{D}\|_{1},
	\end{aligned}
	\label{objective_2_reformat}
\end{align}
where $\vec{U}$, $\vec{V}$, and $\vec{D}$ denote auxiliary variables with the respective Bregman variables $\vec{B}_{U}$, $\vec{B}_{V}$, and $\vec{B}_{D}$ to enforce the constraints of spatial TV, QuaSI, and temporal TV regularization, respectively.

Following MISO denoising as presented in \sref{sec:MISO}, we linearize the non-linear quantile operator $Q(\vec{F}) = (Q(\vec{f}_1), \ldots, Q(\vec{f}_T))^\top$ using \eqref{eqn:linearizationQuasi}. Then, we have $Q(\vec{F}) = \vec{Q}\vec{F}$, where $\vec{Q} = (\vec{Q}_1, \ldots, \vec{Q}_T)^\top$ and for each volume $\vec{f}_t$ in the sequence $\vec{F}$ we have $Q(\vec{f}_t) = \vec{Q}_t \vec{f}_t$. Based on this linearization, we solve \eqref{objective_2_reformat} with an alternating scheme by minimizing \wrt the individual parameters. The minimization \wrt $\vec{F}$ leads to the linear system:
\begin{align}
	\vec{A}\vec{F}^{k+1} &= \vec{b} 
		\label{eqn:cgEquationSystemMIMO}\\
	\vec{A} &= \vec{W} + \beta \nabla_{x,y,z}^\top\nabla_{x,y,z} + \gamma \nabla_t^\top\nabla_t + \alpha \vec{M}^\top\vec{M}
		\label{eqn:cgEquationSystem:MIMOA}\\
	\begin{split}
		\vec{b} &= 2 \vec{W}\vec{G} + \beta\nabla_{x,y,z}^\top(\vec{D}_{x,y,z}-\vec{B}_{x,y,z}) \\
		 & + \gamma\nabla_t^\top(\vec{D}_t - \vec{B}_t) + \alpha \vec{M}^\top(\vec{U} - \vec{B}_u),
	\end{split}
		\label{eqn:cgEquationSystem:MOMOb}
\end{align}
where $\vec{W}$ is a diagonal weight matrix associated with $\vec{F}^k$ and constructed from the Huber loss according to \eqref{eqn:huberWeight}. We then solve \eqref{eqn:cgEquationSystem:MOMOb} using CG iterations. 

The auxiliary variables $\vec{U}$, $\vec{V}$, and $\vec{D}$ are updated element-wise according to:
\begin{align}
	U_i^{k+1} = ~ \mathrm{shrink} (&[\vec{F}^{k+1} - \vec{Q} \vec{F}^{k+1} + \vec{B}^{k}_{U}]_i, \lambda/\alpha)
		\label{eqn:updateU} \\	
	V_i^{k+1} = ~\mathrm{shrink} (&[\nabla_{x,y,z} \vec{F}^{k+1} + \vec{B}^{k}_V]_i, \mu / \beta ) \\
	D_i^{k+1} = ~\mathrm{shrink} (&[\nabla_t \vec{F}^{k+1} + \vec{B}^{k}_D]_i, \omega / \gamma ).
		\label{eqn:updateD}
\end{align}

Given the intermediate sequence $\vec{F}^{k+1}$ along with the auxiliary variables $\vec{U}^{k+1}$, $\vec{V}^{k+1}$, and $\vec{D}^{k+1}$, the Bregman variables are updated according to: 
\begin{align}
	\vec{B}^{k+1}_U &=~ \vec{B}^{k}_U + (\vec{F}^{k+1} - \vec{Q} \vec{F}^{k+1} - \vec{U}^{k+1}) \label{eqn:updateBU}\\
	\vec{B}^{k+1}_V &=~ \vec{B}_V^{k} + (\nabla_{x,y,z} \vec{F}^{k+1} - \vec{V}^{k+1}) \\
	\vec{B}^{k+1}_D &=~ \vec{B}_D^{k} + (\nabla_t \vec{F}^{k+1} - \vec{D}^{k+1}).\label{eqn:updateBD}
\end{align}
An illustration of the proposed optimization scheme is given in Algorithm \ref{alg:MIMODenoising}.

\section{Applications and Evaluation}
\label{sec:experimentsAndResults}

In order to show the applicability of the proposed framework for image, volumetric and volumetric+temporal denoising, we evaluate our framework in different diagnostic and interventional imaging workflows namely OCT as well as C-arm CT. Specifically, we benchmark our method on different datasets including comparisons to the state-of-the-art in the respective fields. 

\begin{figure*}[!tb]
 	\scriptsize
 	\centering
 	\setlength \figurewidth{0.35\textwidth}
 	\setlength \figureheight{0.75\figurewidth}	 	
 	\subfloat{
 		\includegraphics[width = 0.6\textwidth]{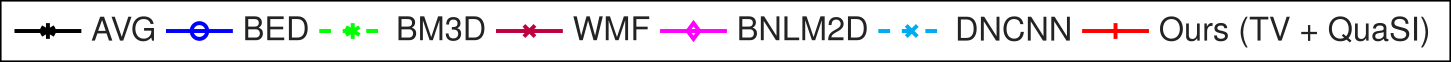}
 	} 	

 	\subfloat{
		\input{images/pigEyePSNR_color.tikz}
 	}
 	\qquad
 	\subfloat{
 		\input{images/pigEyeSSIM_color.tikz}
 	}
 	\caption{Quantification of noise reduction in terms of mean PSNR and SSIM for different denoising methods on the pig eye dataset for different numbers of input images. The points on the curves denote the average PSNR, and SSIM respectively, over the entire pig eye dataset using the number of input images denoted on the x-axis.}
 	\label{fig:pigEyeQualityMeasures}
\end{figure*}

\begin{figure*}[!tb]
 	\scriptsize
 	\centering
 	\subfloat[Noisy input image]{
 		\begin{tikzpicture}[spy using outlines={rectangle,orange,magnification=2.5, 
 			height=1cm, width = 1cm, connect spies, every spy on node/.append style={thick}}]
			\node {\pgfimage[width = 0.31\textwidth]{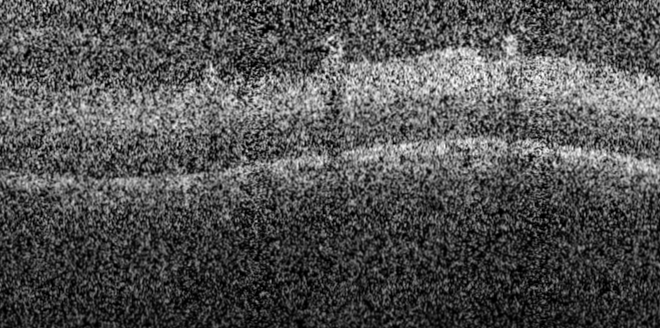}};
			\spy on (0.8,0.9) in node [left] at (2.55, -0.8);     
		\end{tikzpicture}\label{fig:pigEyeDataImages:noisy}
	} 
	\subfloat[AVG]{\begin{tikzpicture}[spy using outlines={rectangle,orange,magnification=2.5, height=1cm, width = 1cm, connect spies, every spy on node/.append style={thick}}]
			\node {\pgfimage[width = 0.31\textwidth]{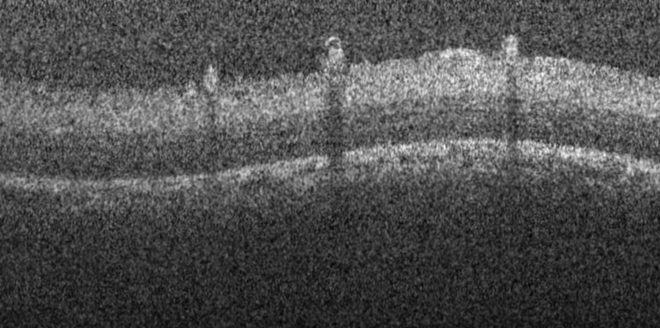}};
			\spy on (0.8,0.9) in node [left] at (2.55, -0.8);     
    \end{tikzpicture}\label{fig:pigEyeDataImages:AVG}
    }
	\subfloat[BED \citep{Wong2010}]{
		\begin{tikzpicture}[spy using outlines={rectangle,orange,magnification=2.5, 
			height=1cm, width = 1cm, connect spies, every spy on node/.append style={thick}}]
			\node {\pgfimage[width = 0.31\textwidth]{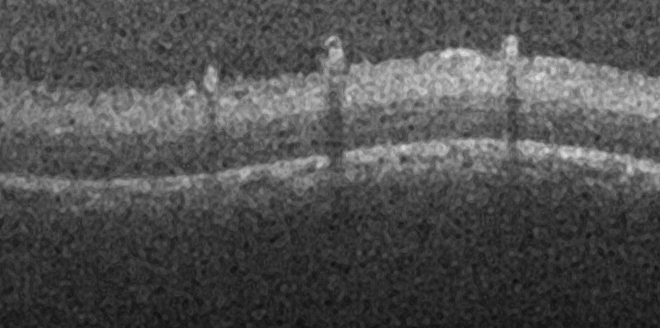}};
			\spy on (0.8,0.9) in node [left] at (2.55, -0.8);     
		\end{tikzpicture}
	}
	
	\subfloat[BM3D \citep{Dabov2007}]{
		\begin{tikzpicture}[spy using outlines={rectangle,orange,magnification=2.5, 
			height=1cm, width = 1cm, connect spies, every spy on node/.append style={thick}}]
			\node {\pgfimage[width = 0.31\textwidth]{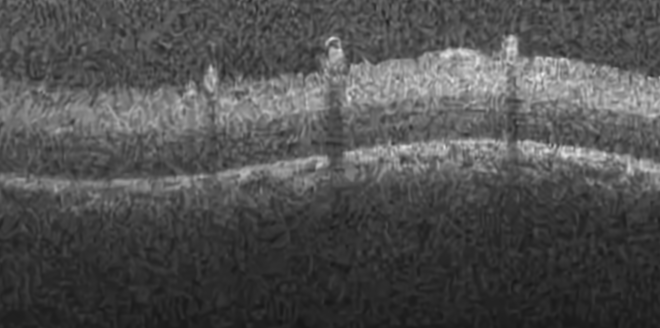}};
			\spy on (0.8,0.9) in node [left] at (2.55, -0.8);     
		\end{tikzpicture}
	} 
	\subfloat[WMF \citep{Mayer2012}]{
		\begin{tikzpicture}[spy using outlines={rectangle,orange,magnification=2.5, 
			height=1cm, width = 1cm, connect spies, every spy on node/.append style={thick}}]
			\node {\pgfimage[width = 0.31\textwidth]{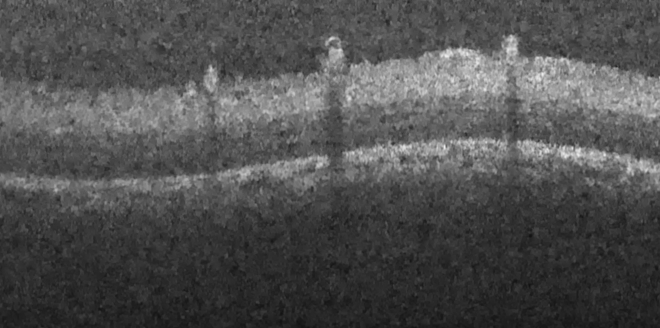}};
			\spy on (0.8,0.9) in node [left] at (2.55, -0.8);     
    	\end{tikzpicture}
    }
    	\subfloat[BNLM2D \citep{Coupe2009}]{
		\begin{tikzpicture}[spy using outlines={rectangle,orange,magnification=2.5, 
			height=1cm, width = 1cm, connect spies, every spy on node/.append style={thick}}]
			\node {\pgfimage[width = 0.31\textwidth]{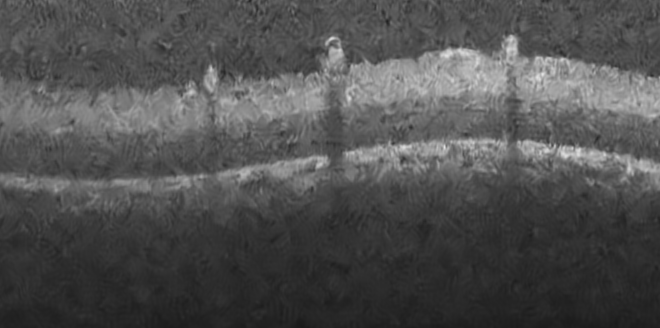}};
			\spy on (0.8,0.9) in node [left] at (2.55, -0.8);     
    	\end{tikzpicture}
    }
 
 	\subfloat[DnCNN \cite{Zhang2017}]{
		\begin{tikzpicture}[spy using outlines={rectangle,orange,magnification=2.5, 
			height=1cm, width = 1cm, connect spies, every spy on node/.append style={thick}}]
			\node {\pgfimage[width = 0.31\textwidth]{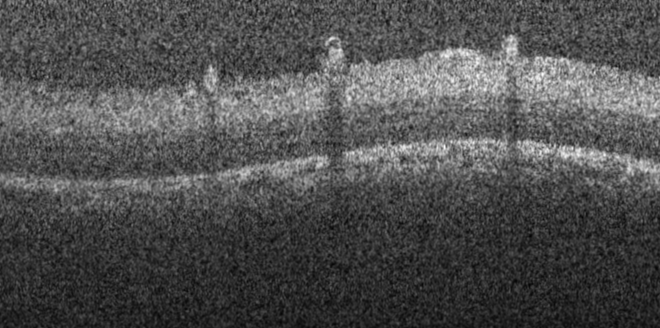}};
			\spy on (0.8,0.9) in node [left] at (2.55, -0.8);     
	     \end{tikzpicture}\label{fig:pigEyeDataImages:dncnn}
     }
	\subfloat[Ours (TV + QuaSI)]{
		\begin{tikzpicture}[spy using outlines={rectangle,orange,magnification=2.5, 
			height=1cm, width = 1cm, connect spies, every spy on node/.append style={thick}}]
			\node {\pgfimage[width = 0.31\textwidth]{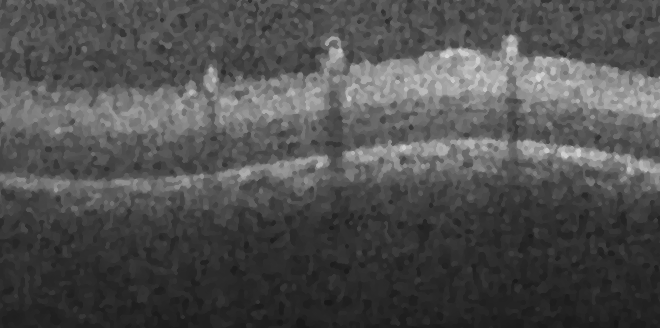}};
			\spy on (0.8,0.9) in node [left] at (2.55, -0.8);     
	     \end{tikzpicture}\label{fig:pigEyeDataImages:ours}
     }	
	\caption{Denoising on position 9 from the pig eye dataset using $5$ B-scans. \protect\subref{fig:pigEyeDataImages:noisy} Noisy image, \protect\subref{fig:pigEyeDataImages:AVG} -- \protect\subref{fig:pigEyeDataImages:ours} AVG, BED \citep{Wong2010}, BM3D \citep{Dabov2007}, WMF \citep{Mayer2012}, BNLM2D \citep{Coupe2009}, DnCNN \citep{Zhang2017} and the proposed method.}
	\label{fig:pigEye} 
\end{figure*}

\subsection{Optical Coherence Tomography Denoising}

Throughout all experiments on the OCT data, we adopted our framework to image and volumetric denoising. For denoising on a B-scan level, the parameters were set to $\mu = 0.075 \cdot T$, $\lambda = 5.0 \cdot T$, $\alpha = 100.0 \cdot T$, $\beta = 1.5 \cdot T$, $K_{\mathrm{outer}} = 20$ and $K_{\mathrm{inner}} = 2$ for $T$ B-scans and $3 \times 3$ median filtering to setup the QuaSI prior. 
In order to find appropriate standard parameter for the proposed method, we proceeded as follows. The parameter search was conducted on the pig eye dataset, using a clinical relevant image section of eye position 11 and 12 with 5 noisy B-scans each. First, the parameter of the proposed algorithm with pure TV regularization were set using a grid search approach for $\mu$ and $\beta$. To quantify the image quality, peak-signal-to-noise ratio (PSNR) and structural similarity index (SSIM) were evaluated in addition to a qualitative investigation. Second, the parameter of the proposed algorithm with QuaSI + TV regularization were set, using the optimal TV weights from the previous investigation.

For volumetric denoising based on $Z = 6$ adjacent B-scans, the parameters were set to $\mu = 0.0007 \cdot T$, $\lambda = 1.0 \cdot T$, $\alpha = 120.0 \cdot T$, $\beta = 0.05  \cdot T$, $K_{\mathrm{outer}} = 20$ and $K_{\mathrm{inner}} = 2$ for $T$ volumes and $3 \times 3 \times 3$ median filtering. The proposed algorithm for volumetric denoising was evaluated on clinical data only. The selection of standard parameters was performed in the same way as for denoising on a B-scan level. Using $Z = 6$ adjacent B-scans in $T = 5$ volumes from only 1 patient, the TV weights followed by the QuaSI weights were set.

\begin{figure*}[!tb]
	\scriptsize
	\centering
	\setlength \figurewidth{0.35\textwidth}
	\setlength \figureheight{0.65\figurewidth}
	\subfloat{
		\includegraphics[width = 0.6\textwidth]{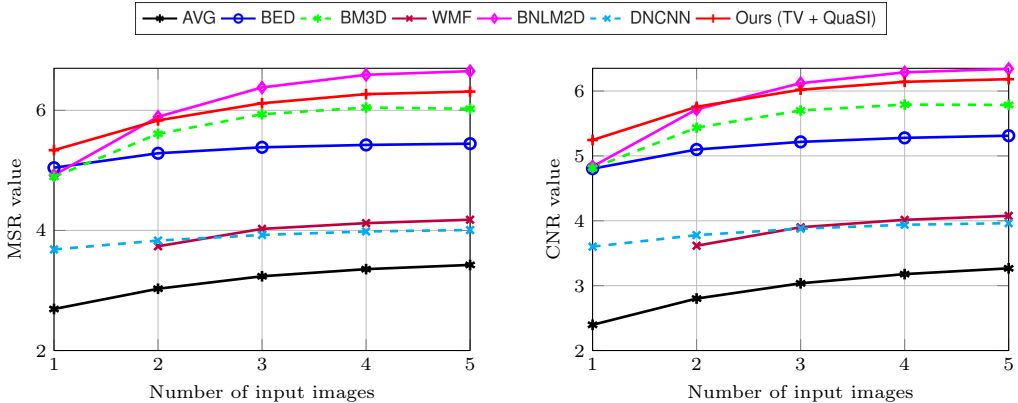}
	}
	
	\subfloat{
		\input{images/clinicalMSR_color.tikz}
	}
	\qquad
  	\subfloat{
  		\input{images/clinicalCNR_color.tikz}
  	}
  	\caption{Quantification of noise reduction in terms of mean MSR and CNR measures for denoising on a B-scan level on our clinical dataset for different numbers of input images. The plots illustrate the mean MSR and CNR of the whole clinical dataset and the 5 foreground regions. Each point on the curves denotes the mean MSR and CNR using the number of input images specified on the x-axis as input to state-of-the-art denoising methods and the proposed algorithm with the QuaSI prior.}
  	\label{fig:patientDataQualityMeasures}
\end{figure*}

\begin{figure*}[!p]
	\centering
	\scriptsize
	\subfloat[Noisy image (MSR: 2.68, CNR: 2.47)]{
		\begin{tikzpicture}[spy using outlines={rectangle,orange,magnification=2.8, 
			height= 3.55cm, width = 1.1cm, connect spies, every spy on node/.append style={thick}}]
			\node {\pgfimage[width = 0.35\textwidth]{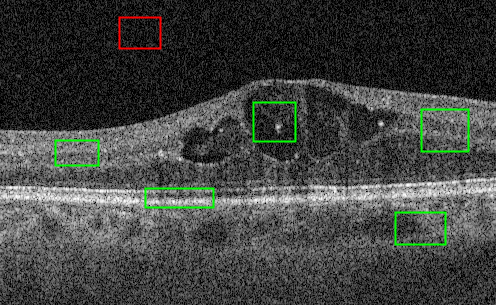}};
			\spy on (1.22,0.22) in node [left] at (4, 0);
		\end{tikzpicture}\label{fig:patientDataImages:noisy}
	}
	\subfloat[AVG (MSR: 3.17, CNR: 3.17)]{
		\begin{tikzpicture}[spy using outlines={rectangle,orange,magnification=2.8, 
			height= 3.55cm, width = 1.1cm, connect spies, every spy on node/.append style={thick}}]
			\node {\pgfimage[width = 0.35\textwidth]{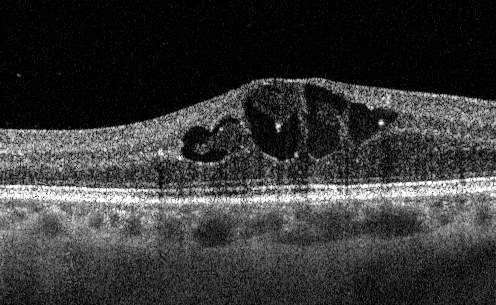}};
			\spy on (1.22,0.22) in node [left] at (4, 0);
    	\end{tikzpicture}\label{fig:patientDataImages:avg}
    }
    	
	\subfloat[BM3D \cite{Dabov2007} (MSR: 4.61, CNR: 4.85)]{
		\begin{tikzpicture}[spy using outlines={rectangle,orange,magnification=2.8, 
			height= 3.55cm, width = 1.1cm, connect spies, every spy on node/.append style={thick}}]
			\node {\pgfimage[width = 0.35\textwidth]{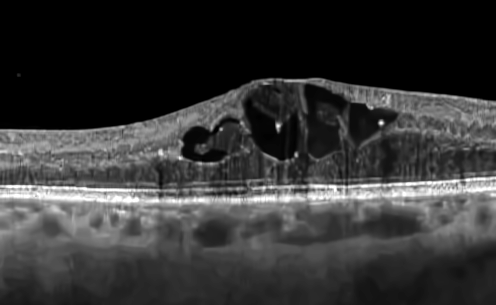}};
			\spy on (1.22,0.22) in node [left] at (4, 0);
    	\end{tikzpicture}\label{fig:patientDataImages:bm3d}
    }    		
	\subfloat[BED \cite{Wong2010} (MSR: 4.67, CNR: 4.85)]{
		\begin{tikzpicture}[spy using outlines={rectangle,orange,magnification=2.8, 
			height= 3.55cm, width = 1.1cm, connect spies, every spy on node/.append style={thick}}]
			\node {\pgfimage[width = 0.35\textwidth]{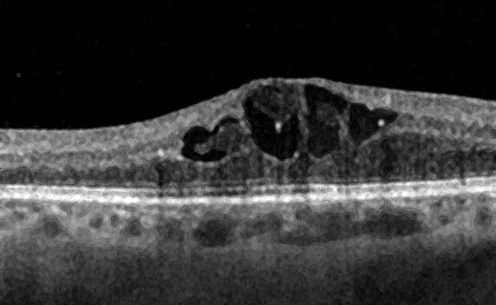}};
			\spy on (1.22,0.22) in node [left] at (4, 0);
    	\end{tikzpicture}\label{fig:patientDataImages:bed}
    }
    		
	\subfloat[WMF \cite{Mayer2012} (MSR: 3.67, CNR: 3.55)]{
		\begin{tikzpicture}[spy using outlines={rectangle,orange,magnification=2.8, 
			height= 3.55cm, width = 1.1cm, connect spies, every spy on node/.append style={thick}}]
			\node {\pgfimage[width = 0.35\textwidth]{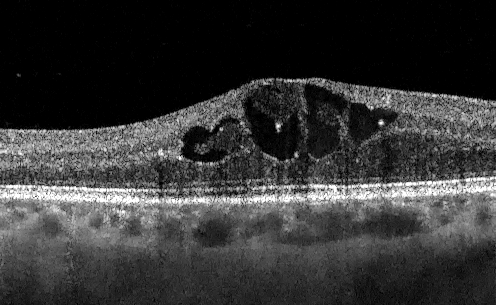}};
			\spy on (1.22,0.22) in node [left] at (4, 0);
    	\end{tikzpicture}\label{fig:patientDataImages:wmf}
    }
		\subfloat[DnCNN (MSR: 3.54, CNR: 3.70)]{
		\begin{tikzpicture}[spy using outlines={rectangle,orange,magnification=2.8, 
			height= 3.55cm, width = 1.1cm, connect spies, every spy on node/.append style={thick}}]
			\node {\pgfimage[width = 0.35\textwidth]{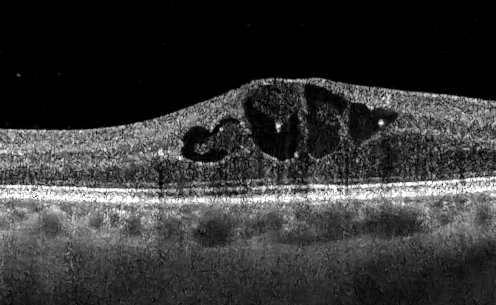}};
			\spy on (1.22,0.22) in node [left] at (4, 0);
    	\end{tikzpicture}\label{fig:patientDataImages:dncnn}
    }
	
	\subfloat[BNLM2D (MSR: 5.04, CNR: 5.32)]{
		\begin{tikzpicture}[spy using outlines={rectangle,orange,magnification=2.8, 
			height= 3.55cm, width = 1.1cm, connect spies, every spy on node/.append style={thick}}]
			\node {\pgfimage[width = 0.35\textwidth]{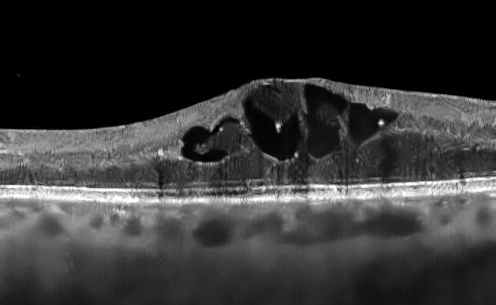}};
			\spy on (1.22,0.22) in node [left] at (4, 0);
    	\end{tikzpicture}\label{fig:patientDataImages:wnlm2D}
    }
  	\subfloat[Ours (MSR: 5.02, CNR: 5.36)]{
		\begin{tikzpicture}[spy using outlines={rectangle,orange,magnification=2.8, 
			height= 3.55cm, width = 1.1cm, connect spies, every spy on node/.append style={thick}}]
			\node {\pgfimage[width = 0.35\textwidth]{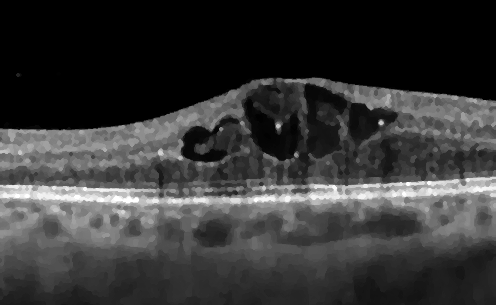}};
			\spy on (1.22,0.22) in node [left] at (4, 0);
    	\end{tikzpicture}\label{fig:patientDataImages:ours}
    } 
	
	\caption{Visual comparison of denoising results using our clinical dataset with the central B-scan of $T = 5$ volumes from a 46 years old male patient with diabetic retinopathy. \protect\subref{fig:patientDataImages:noisy} Noisy image with manually selected background (red) and foreground regions (green) to determine MSR and CNR. \protect\subref{fig:patientDataImages:avg} -- \protect\subref{fig:patientDataImages:ours} AVG, BM3D \citep{Dabov2007}, BED \citep{Wong2010}, WMF \citep{Mayer2012}, DnCNN \citep{Zhang2017}, BNLM2D \citep{Coupe2009}, and the proposed method.}
	\label{fig:patientDataImages}
\end{figure*}

\subsubsection{Datasets}
\label{sec:datasets}

To evaluate the performance of the proposed denoising algorithm, we conducted experiments on two different OCT datasets. This comprises ex-vivo benchmark data and real clinical data.


For an evaluation of denoising on B-scan level, we used the publicly available pig eye dataset provided by \cite{Mayer2012}. The dataset comprises 455 B-scans corresponding to 35 eye positions with 13 scans per position and was captured ex-vivo with a Spectralis HRA \& OCT. The published B-scans were registered to each other to compensate for geometric shifts. We apply denoising to sets of $T$ registered B-scans with $T \in [1,13]$ to demonstrate the influence of different numbers of input B-scans on the denoising result. The pig eye dataset provides a gold standard B-scan that was obtained by averaging all 455 registered scans. The quality of the denoising algorithm was evaluated by assessing the fidelity of a denoised B-scan \wrt the gold standard using the peak-signal-to-noise ratio (PSNR) as well as the structural similarity index (SSIM). 


In order to evaluate and compare B-scans with volumetric denoising, we use clinical data. A prototype ultrahigh-speed swept-source OCT system with 1050\,nm wavelength and a sampling rate of 400,000 A-scans per second \citep{Choi2013a} was used to acquire volumetric data of 14 human subjects. Proliferative and non-proliferative diabetic retinopathy, early age-related macular degeneration and one healthy subject were imaged on two volumes per subject, where each B-scan was acquired five times in immediate succession. We use 500 A-scans by 500 B-scans for a field size of $3 \times 3$\,mm.

For denoising on a B-scan level, the central B-scan of each volume is used, while volumetric denoising is performed on adjacent B-scans including the central one. As the clinical data does not provide a gold standard, we follow prior work by \cite{Fang2012,Ozcan2007,Wong2010} and measure the noise reduction using the mean-to-standard-deviation ratio (MSR) and the contrast-to-noise ratio (CNR) according to:
\begin{align}
	\mathrm{MSR} &= \frac{\mu_{f}}{\sigma_{f}}\\
	\mathrm{CNR} &= \frac{| \mu_{f} - \mu_{b} |} { \frac{1}{2} \sqrt{(\sigma_{f}^{2} + \sigma_{b}^{2})}},
\end{align}
where $\mu_{f}$ and $\mu_{b}$ as well as $\sigma_{f}$ and $\sigma_{b}$ are the means and standard deviations of the intensities in a foreground and a background region, respectively. The regions to determine MSR and CNR were manually selected for the central B-scan, see Fig.~\ref{fig:patientDataImages:noisy}.

\subsubsection{Comparison to the State-of-the-Art}

We compared our method against seven competing denoising approaches. As representatives of general-purpose methods, we evaluated BM3D \citep{Dabov2007} as well as a deep denoising CNN (DnCNN) \citep{Zhang2017}, which are state-of-the-art in the field of natural image denoising. We also evaluated non-local means-based speckle noise filtering (BNLM2D) that has been originally proposed for ultrasound image denoising \citep{Coupe2009}. In terms of spatial filters customized for OCT, we used Bayesian estimation denoising (BED) \citep{Wong2010}. In the field of temporal methods using multiple registered B-scans, we evaluate simple averaging (AVG) as a baseline as well as wavelet multi-frame denoising (WMF) \citep{Mayer2012}. To ensure fair comparisons between spatial and temporal methods, we provide the average of all B-scans as input for single-image denoising (BM3D, BNLM2D, DnCNN, and BED). In contrast, AVG and WMF are pure temporal approaches that process multiple registered B-scans. Notice that all of these methods can only operate on individual 2-D B-scans to denoise volumetric data and are therefore compared to our proposed method on a B-scan level. The parameters of the competing methods were set according to suggestions of the authors and adapted to the OCT data.

\begin{figure*}[!tb]
	\centering
	\scriptsize
		\subfloat{
			\begin{tikzpicture}[spy using outlines={rectangle,orange,magnification=2.5,
				 height=1.1cm, width = 1.1cm, connect spies, every spy on node/.append style={thick}}]
				\node {\pgfimage[width = 0.31\textwidth]{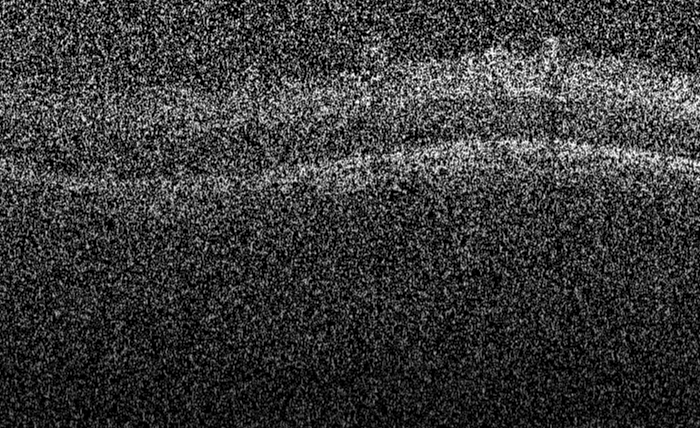}};
				\spy on (0.2,1.2) in node [left] at (1.88, -0.59);     
			\end{tikzpicture}
		}
		\subfloat{
			\begin{tikzpicture}[spy using outlines={rectangle,orange,magnification=2.5,
				 height=1.1cm, width = 1.1cm, connect spies, every spy on node/.append style={thick}}]
				\node {\pgfimage[width = 0.31\textwidth]{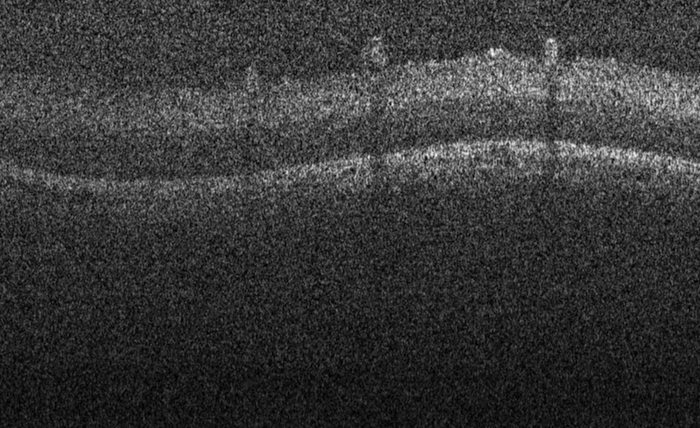}};
				\spy on (0.2,1.2) in node [left] at (1.88, -0.59);     
    		\end{tikzpicture}
    	}
		\subfloat{
			\begin{tikzpicture}[spy using outlines={rectangle,orange,magnification=2.5, 
				height=1.1cm, width = 1.1cm, connect spies, every spy on node/.append style={thick}}]
				\node {\pgfimage[width = 0.31\textwidth]{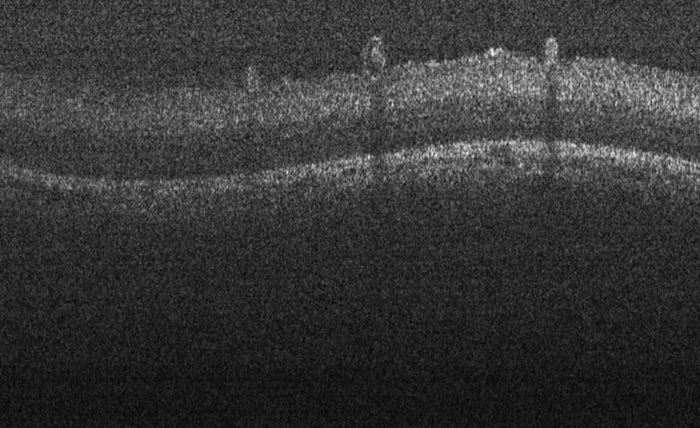}};
				\spy on (0.2,1.2) in node [left] at (1.88, -0.59);     
		\end{tikzpicture}
		}
		
		\setcounter{subfigure}{0}
		\subfloat{
			\begin{tikzpicture}[spy using outlines={rectangle,orange,magnification=2.5, 
				height=1.1cm, width = 1.1cm, connect spies, every spy on node/.append style={thick}}]
				\node {\pgfimage[width = 0.31\textwidth]{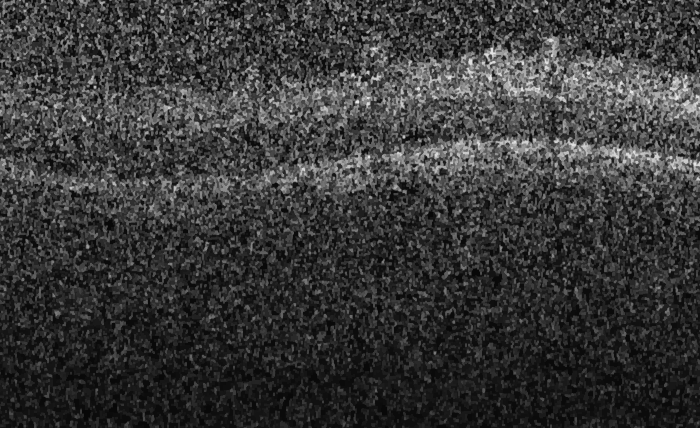}};
				\spy on (0.2,1.2) in node [left] at (1.88, -0.59);     
			\end{tikzpicture}
		}
		\subfloat{
			\begin{tikzpicture}[spy using outlines={rectangle,orange,magnification=2.5, 
				height=1.1cm, width = 1.1cm, connect spies, every spy on node/.append style={thick}}]
				\node {\pgfimage[width = 0.31\textwidth]{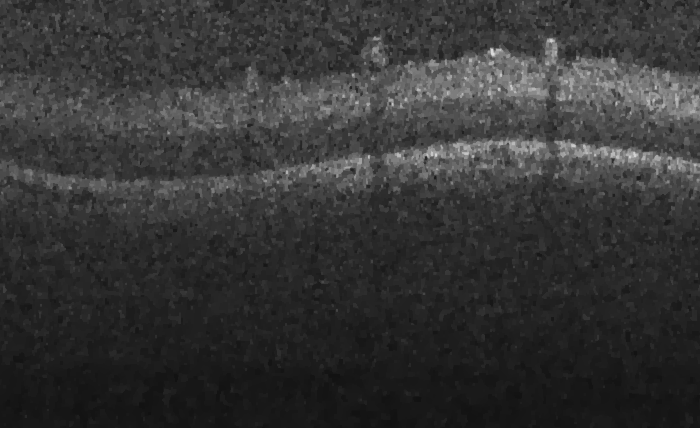}};
				\spy on (0.2,1.2) in node [left] at (1.88, -0.59);     
    	\end{tikzpicture}
    	}
		\subfloat{
			\begin{tikzpicture}[spy using outlines={rectangle,orange,magnification=2.5, 
				height=1.1cm, width = 1.1cm, connect spies, every spy on node/.append style={thick}}]
				\node {\pgfimage[width = 0.31\textwidth]{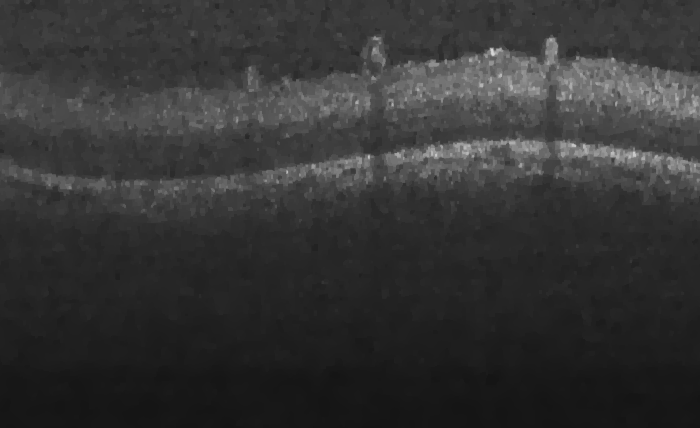}};
				\spy on (0.2,1.2) in node [left] at (1.88, -0.59);     
			\end{tikzpicture}
		}
		\setcounter{subfigure}{0}
		\subfloat[$T = 1$]{
			\begin{tikzpicture}[spy using outlines={rectangle,orange,magnification=2.5, 
				height=1.1cm, width = 1.1cm, connect spies, every spy on node/.append style={thick}}]
				\node {\pgfimage[width = 0.31\textwidth]{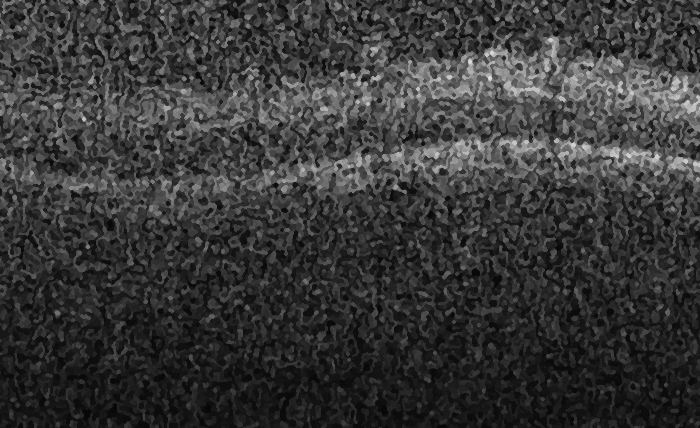}};
				\spy on (0.2,1.2) in node [left] at (1.88, -0.59);     
     		\end{tikzpicture}
     	}
		\subfloat[$T = 5$]{
			\begin{tikzpicture}[spy using outlines={rectangle,orange,magnification=2.5, 
				height=1.1cm, width = 1.1cm, connect spies, every spy on node/.append style={thick}}]
					\node {\pgfimage[width = 0.31\textwidth]{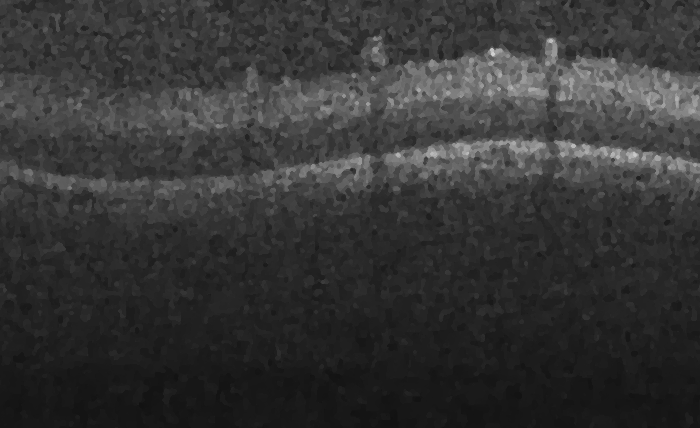}};
					\spy on (0.2,1.2) in node [left] at (1.88, -0.59);     
			\end{tikzpicture}
		}
		\subfloat[$T = 13$]{
			\begin{tikzpicture}[spy using outlines={rectangle,orange,magnification=2.5, 
				height=1.1cm, width = 1.1cm, connect spies, every spy on node/.append style={thick}}]
				\node {\pgfimage[width = 0.31\textwidth]{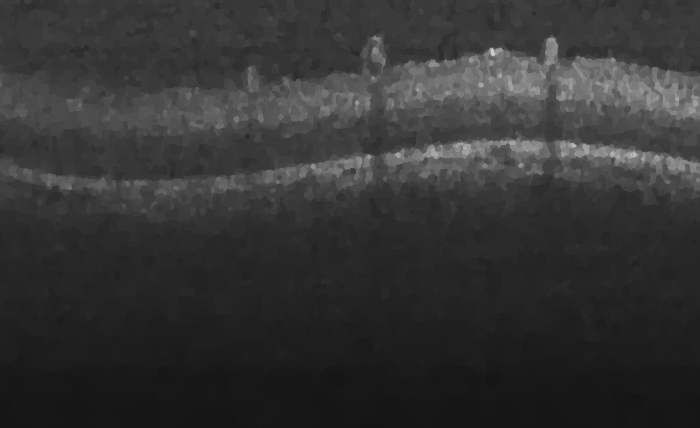}};
				\spy on (0.2,1.2) in node [left] at (1.88, -0.59);     
    		\end{tikzpicture}
    	}
	\caption{This comparison aims at demonstrating the improvement of the proposed spatio-temporal denoising with TV + QuaSI regularization (third row) compared to simple averaging of registered B-scans (top row) and the proposed spatio-temporal denoising with TV regularization only (second row) for different numbers of input images. For the comparison, dataset 27 from the pig eye dataset was used to evaluate the proposed algorithm with and without the QuaSI prior using the standard parameter.}
	\label{fig:pigEyeImages}
\end{figure*}

\begin{figure*}
\centering
\subfloat{
\includegraphics[width = 0.25\textwidth]{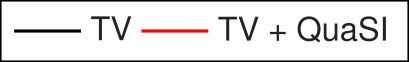}}\\
\subfloat{
\includegraphics[width=0.95\textwidth]{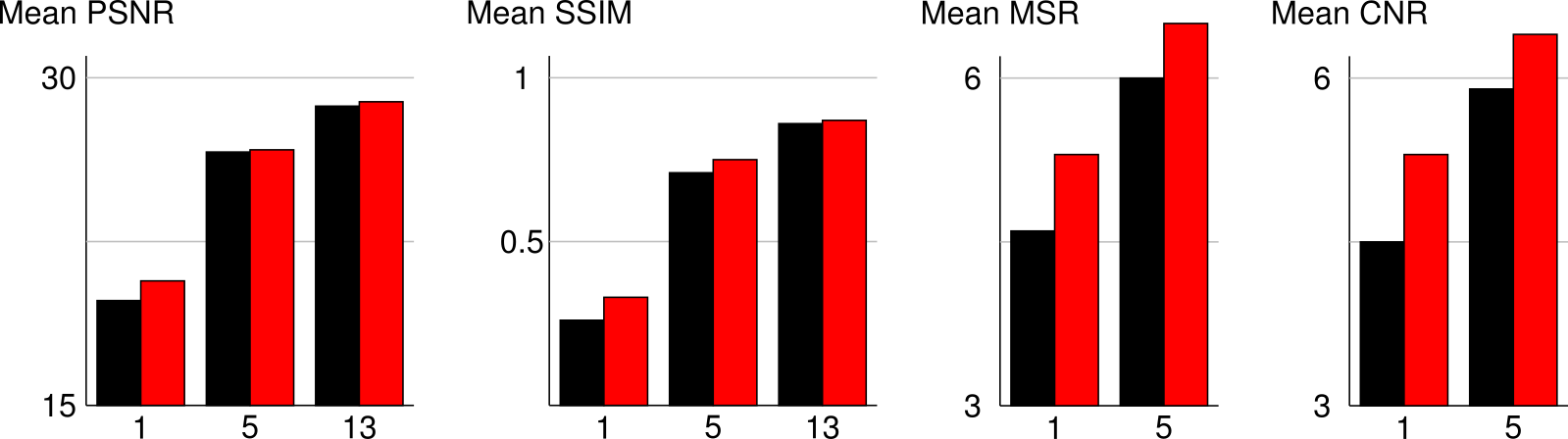}}
\caption{Mean PSNR, SSIM, MSR and CNR measures to quantify noise reduction with and without the QuaSI prior for 1, 5 and 13 input images. The two bar graphs on the left hand side illustrate the average PSNR and SSIM over the entire pig eye dataset using the proposed algorithm with and without QuaSI prior and the standard parameters. The average MSR and CNR over the entire clinical dataset is shown in the two bar graphs on the right hand side.}
\label{fig:TVvsQuaSImeasures}
\end{figure*}

First, we conducted experiments for denoising on B-scan level on the pig eye dataset. Figure~\ref{fig:pigEyeQualityMeasures} depicts the mean PSRN and SSIM of the competing denoising methods \wrt the gold standard for different numbers of input B-scans. We observed quantitatively that our proposed method consistently outperforms the competing BM3D, BED, and WMF denoising methods regardless of the number of input frames. Moreover, using only $T = 2$ input B-scans, our spatio-temporal method achieved comparable results to averaging $T = 5$ B-scans. The proposed method performs better than BNLM2D for $T < 5$ input B-scans. This reveals that our method is more economic regarding the number of required input scans. This property is essential for clinical applications, where acquiring more scans might lead to unacceptable long acquisition times. Figure~\ref{fig:pigEye} depicts qualitative results for $T = 5$ B-scans. Here, the proposed algorithm using the QuaSI prior achieved superior performance in terms of noise reduction, while anatomical structures like retinal layers are preserved. Comparable results are achieved by BNLM2D, but the latter suffers from small streak-like artifacts. DnCNN achieved comparable results to simple averaging both regarding quantitative measures and qualitative assessment.

Second, denoising on a B-scan level was studied on our clinical datasets using the non-reference MSR and CNR measures for a quantitative evaluation. Figure~\ref{fig:patientDataQualityMeasures} depicts the averaged MSR and CNR measures for different numbers of input images. Overall, we observed that BNLM2D and our proposed method achieved the best noise reduction expressed by both measures. Figure~\ref{fig:patientDataImages} compares the denoising performance on one example dataset. We found that AVG, WMF, and BED facilitate structure-preserving denoising but were prone to noise breakthroughs in homogeneous areas, which lowers their MSR and CNR. In contrast, BM3D achieved superior noise reduction but suffered from streak artifacts. Similar observations were made in related work on OCT denoising \citep{Fang2012} and can be explained by the assumption of additive white Gaussian noise used for BM3D. The proposed method achieved a decent tradeoff between noise reduction and structure preservation.

\subsubsection{Impact of the QuaSI Prior}

We used the pig eye dataset as well as clinical data to evaluate the performance of our spatio-temporal denoising algorithm with and without the QuaSI prior. Figure~\ref{fig:pigEyeImages} illustrates the impact of the QuaSI prior on the denoising result for the pig eye data compared to simple averaging and pure TV regularization. In terms of noise reduction, the proposed variational framework outperformed simple averaging. Especially in the enlarged region, a noticeable difference between averaging and the proposed denoising algorithm is shown. In homogeneous areas, the algorithm considerably suppressed speckle noise, while preserving important structures. The noise reduction was superior when using a combination of the QuaSI prior and the TV prior for regularization as shown for the retinal structures in the enlarged region. In addition, the QuaSI prior contributed to structure-preservation and avoided staircasing artifacts that typically appear in TV denoising.  

Figure~\ref{fig:TVvsQuaSImeasures} illustrates the impact of the QuaSI prior using PSNR and SSIM (for the pig eye data) as well as MSR and CNR (for clinical data) for different numbers of input scans. Here, our denoising framework with QuaSI prior outperformed TV denoising in terms of all measures.

\begin{figure*}[!tb]
	\centering
	\scriptsize
		\subfloat[BM4D $T=1$ (MSR: 5.14, CNR: 5,63)]{
			\begin{tikzpicture}[spy using outlines={rectangle,orange,magnification=2.8,
				 height= 3.55cm, width = 1.1cm, connect spies, every spy on node/.append style={thick}}]
			\node {\pgfimage[width = 0.35\textwidth]{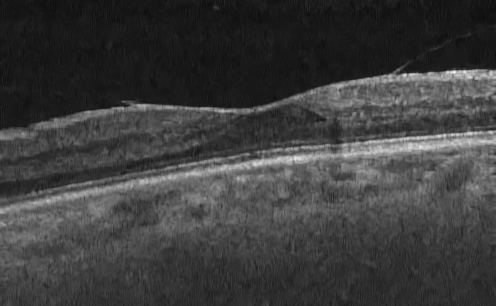}};
			\spy on (-0.75,0.12) in node [left] at (4, 0);
			\end{tikzpicture}\label{fig:patientDataImagesK1:BM4D:1}
		}	
		\subfloat[BM4D $T=5$ (MSR: 6.10, CNR: 5.65)]{
			\begin{tikzpicture}[spy using outlines={rectangle,orange,magnification=2.8, 
					height= 3.55cm, width = 1.1cm, connect spies, every spy on node/.append style={thick}}]
				\node {\pgfimage[width = 0.35\textwidth]{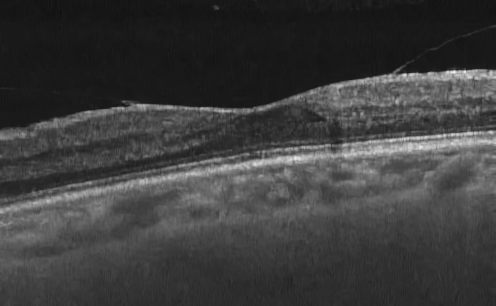}};
				\spy on (-0.75,0.12) in node [left] at (4, 0);
    		\end{tikzpicture}\label{fig:patientDataImagesK5:BM4D:5}
    	}

	 	\subfloat[B-scan denoising $T=1$ (MSR: 5.68, CNR: 5.25)]{
			\begin{tikzpicture}[spy using outlines={rectangle,orange,magnification=2.8, 
					height= 3.55cm, width = 1.1cm, connect spies, every spy on node/.append style={thick}}]
				\node {\pgfimage[width = 0.35\textwidth]{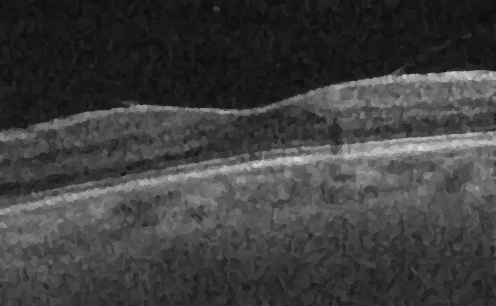}};
				\spy on (-0.75,0.12) in node [left] at (4, 0);
    		\end{tikzpicture}\label{fig:patientDataImagesK1:2D}
    	}
    	\subfloat[B-scan denoising $T=5$ (MSR: 7.78, CNR: 7.15)]{
			\begin{tikzpicture}[spy using outlines={rectangle,orange,magnification=2.8, 
					height= 3.55cm, width = 1.1cm, connect spies, every spy on node/.append style={thick}}]
				\node {\pgfimage[width = 0.35\textwidth]{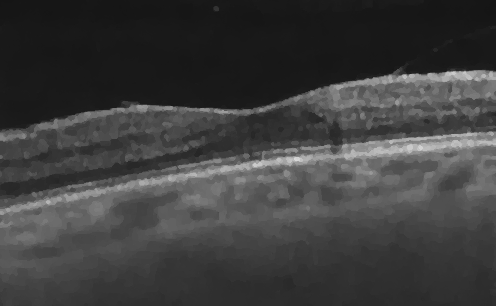}};
				\spy on (-0.75,0.12) in node [left] at (4, 0);
    		\end{tikzpicture}\label{fig:patientDataImagesK5:2D}
    	}
    	 	
		\subfloat[Volumetric denoising $T=1$ (MSR: 6.43, CNR: 5.99)]{
			\begin{tikzpicture}[spy using outlines={rectangle,orange,magnification=2.8,
				 height= 3.55cm, width = 1.1cm, connect spies, every spy on node/.append style={thick}}]
			\node {\pgfimage[width = 0.35\textwidth]{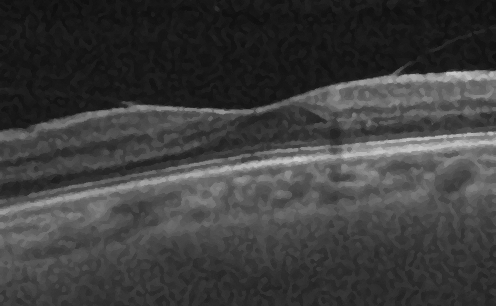}};
			\spy on (-0.75,0.12) in node [left] at (4, 0);
			\end{tikzpicture}\label{fig:patientDataImagesK1:3D}
		}	
		\subfloat[Volumetric denoising $T=5$ (MSR: 6.85, CNR: 6.51)]{
			\begin{tikzpicture}[spy using outlines={rectangle,orange,magnification=2.8,
				 height= 3.55cm, width = 1.1cm, connect spies, every spy on node/.append style={thick}}]
			\node {\pgfimage[width = 0.35\textwidth]{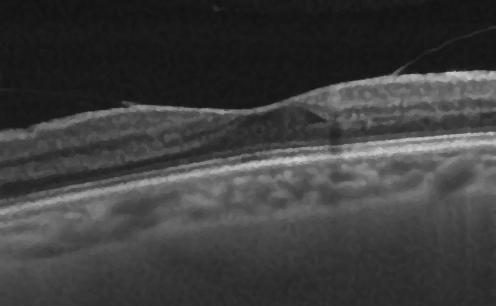}};
			\spy on (-0.75,0.12) in node [left] at (4, 0);
			\end{tikzpicture}\label{fig:patientDataImagesK5:3D}
		}	
	\caption{Denoising on the clincial dataset using $T = 5$ registered volumes from a 67 years old male patient with non-proliferative diabetic retinopathy. The left column illustrates the results of the proposed method on a B-scan level with $Z = 1$ scan \protect\subref{fig:patientDataImagesK1:2D} and on a volumetric level \protect\subref{fig:patientDataImagesK1:3D} as well as BM4D \protect\subref{fig:patientDataImagesK1:BM4D:1} with $Z = 6$ consecutive scans using $T = 1$ input volume. The right column illustrates the results of the proposed method on a B-scan level with $Z = 1$ scan  \protect\subref{fig:patientDataImagesK5:2D} and on a volumetric level \protect\subref{fig:patientDataImagesK5:3D} as well as BM4D \protect\subref{fig:patientDataImagesK5:BM4D:5} with $Z = 6$ consecutive scans using $T = 5$ registered input volumes.} 
	\label{fig:patientDataImages2Dvs3D}
\end{figure*}

\subsubsection{B-scan vs. Volumetric Denoising}

\begin{table*}[!tb] 
\centering
\small
\begin{tabular}{L{1cm}cccccc}\toprule
 &  \multicolumn{3}{c}{$T = 1$ volume} & \multicolumn{3}{c}{$T = 5$ volumes} \\
 & \multicolumn{1}{C{2cm}} {BM4D  \citep{maggioni2013nonlocal}}& \multicolumn{1}{C{2cm}} {B-scan denoising}&  \multicolumn{1}{C{2cm}} {Volumetric denoising}&  \multicolumn{1}{C{2cm}} {BM4D  \citep{maggioni2013nonlocal}}& \multicolumn{1}{C{2cm}} {B-scan denoising}&  \multicolumn{1}{C{2cm}} {Volumetric denoising} \\ \midrule
MSR &5.16&5.35&5.77 &5.38& 6.50 & 6.31 \\ \midrule
CNR &5.00&5.27&5.60 & 5.23&6.38 & 6.18 \\ \bottomrule
\end{tabular}
\caption{Mean MSR and CNR measures for 1 and 5 registered input volumes on the clinical data. For B-scan denoising, the central B-scan is used and for volumetric denoising 6 adjacent B-scans including the central one are used. The B-scan-wise average of $T = 5$ input volumes served as input to BM4D \citep{maggioni2013nonlocal}.}
\label{tab:2Dvs3D}
\end{table*} 

So far, we evaluated denoising of volumetric OCT data by simply processing individual B-scans. In order to evaluate the impact of true volumetric denoising to simple B-scan wise denoising in our proposed framework, we used our clinical dataset. Volumetric denoising processes 6 consecutive B-scans including the central one. That way, CNR and MSR measures from the previous experiments can be used for comparison. Table~\ref{tab:2Dvs3D} shows the mean MSR and CNR using $T = 1$ and $T = 5$ registered input volumes. The proposed method is compared to BM4D \citep{maggioni2013nonlocal} using $T = 1$ volume and the average of $T = 5$ volumues as an input. Here, we found that our volumetric denoising achieved better results in terms of noise reduction for $T = 1$ input volume, as adjacent B-scans affect denoising positively. For $T = 5$ input volumes, we found that our B-scan denoising achieved slightly better results in terms of noise reduction. However, as opposed to noise reduction, volumetric denoising achieved superior performance in structure preservation by exploiting coherence between adjacent B-scans. This is depicted in Fig.~\ref{fig:patientDataImages2Dvs3D}, where the retinal layers in the magnified region can be better distinguished. 

\begin{figure*}[!tb]
	\scriptsize
	\centering
	\setlength\figurewidth{0.255\textwidth}
	\setlength\figureheight{0.74\figurewidth}
	\subfloat{
   \includegraphics[width = 0.8\textwidth]{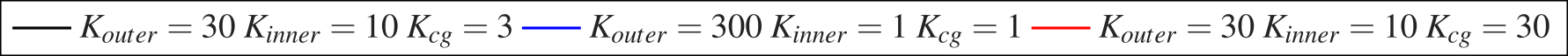}
	}\\
	\subfloat{
    \input{images/conv_PSNR_iter.tikz}
  }
	~
	\subfloat{
   \input{images/conv_SSIM_iter.tikz}
  }
  ~
  \subfloat{
    \input{images/conv_energy_iter.tikz}
  }
	\caption{Convergence analysis for our proposed optimization scheme in OCT B-scan denoising using different combinations of iteration numbers $K_{\mathrm{outer}}$, $K_{\mathrm{inner}}$ and $K_{\mathrm{cg}}$. For each combination, we depict the value of the energy function optimized by ADMM along the with PSNR of the intermediate denoised images over the iterations.}
  \label{fig:convergenceIterations}
\end{figure*}
\begin{figure*}[!tb]
 	\scriptsize
 	\centering
 	\setlength \figurewidth{0.35\textwidth}
 	\setlength \figureheight{0.64\figurewidth}
  \subfloat{
		\includegraphics[width = 0.4\textwidth]{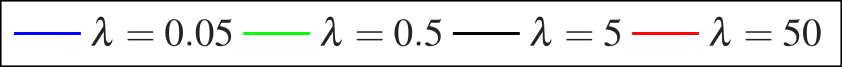}
	}\\
  \subfloat{
 		\input{images/conv_PSNR_lambda.tikz}
 	}
	\qquad
 	\subfloat{
 		\input{images/conv_SSIM_lambda.tikz}
 	}
	\caption{Convergence analysis for our proposed optimization scheme in OCT B-scan denoising using different QuaSI regularization weights $\lambda$. For each parameter setting, we depict the influence of $\lambda$ using the PSNR and SSIM of the intermediate denoised image over the iterations.}
	\label{fig:convergenceLambda}
\end{figure*}
\begin{figure*}[!tb]
 	\scriptsize
	\centering
	\setlength \figurewidth{0.35\textwidth}
	\setlength \figureheight{0.64\figurewidth}
    \subfloat{
		\includegraphics[width = 0.4\textwidth]{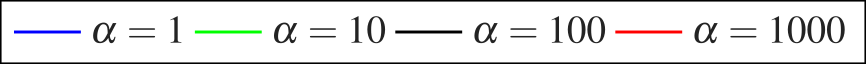}
	}\\
	\subfloat{
		\input{images/conv_PSNR_alpha.tikz}
	}
	\qquad
	\subfloat{
 		\input{images/conv_SSIM_alpha.tikz}
 	}
	\caption{Convergence analysis for our proposed algorithm in OCT B-scan denoising using different Lagrangian multiplier $\alpha$ for ADMM optimization. For each parameter setting, we depict the influence of $\alpha$ using the PSNR and SSIM of the intermediate denoised image over the iterations.}
	\label{fig:convergenceAlpha}
\end{figure*}

\subsubsection{Convergence and Parameter Sensitivity}
\label{sec:convergenceAndParameterSensitivity}

\begin{figure*}[!tb]
 	\scriptsize
 	\centering
 	\setlength \figurewidth{0.35\textwidth}
 	\setlength \figureheight{0.6\figurewidth}	 	
 	\subfloat{
 		\includegraphics[width = 0.35\textwidth]{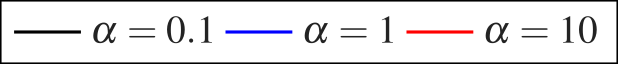}
 	} 	

 	\subfloat{
 		\input{images/sensitivityPSNR.tikz}
 	}
 	\qquad
 	\subfloat{
 		\input{images/sensitivitySSIM.tikz}
 	}
 	\caption{Parameter sensitivity analysis for the interplay of the QuaSI regularization weight $\lambda$ and the Lagrangian multiplier $\alpha$ used for ADMM to B-scan denoising. The PSNR and SSIM measures were evaluated for a clinical relevant region of position 11 from the pig eye dataset. Each measure was determined for different QuaSI parameters $\lambda$ and $\alpha$ while keeping the TV regularization weight $\mu = 0.075$ and the corresponding Lagrangian multiplier $\beta = 1.5$ fixed.}
 	\label{fig:sensitivityQuaSI}
\end{figure*}

The convergence of the proposed algorithm is shown experimentally on a B-scan level. By our definition, the algorithm converges if a stationary point of the objective function \eqref{eqn:objective} is reached. The value of the objective, hereinafter referred to as energy, is computed after every update of the intermediate image $\vec{f}^{k+1}$. In addition, PSNR and SSIM of the intermediate image are computed. Based on the optimal parameter setting  $\mu = 0.075 \cdot T$, $\lambda = 5.0 \cdot T$, $\alpha = 100.0 \cdot T$, $\beta = 1.5 \cdot T$, $K_{\mathrm{outer}} = 30$, $K_{\mathrm{inner}} = 10$ and $K_{\mathrm{cg}} = 3$ for B-scan denoising, we denoise the pig eye dataset 9 with $T = 8$ B-scans.

Figure~\ref{fig:convergenceIterations} shows the impact of $K_{\mathrm{outer}}$, $K_{\mathrm{inner}}$, and $K_{\mathrm{cg}}$ on the convergence using three different parameter settings, where $K_{\mathrm{outer}} \cdot K_{\mathrm{inner}}=300$ for a fair comparison. The approximation of the QuaSI prior is updated every $K_{\mathrm{inner}}$ iterations. We found that increasing numbers of inner iterations ($K_{\mathrm{inner}} = 10$) or CG iterations ($K_{\mathrm{cg}} = 30$) impair the convergence properties of the algorithm as shown by the peaks of the energy and the PSNR. This is mainly caused by the rare update of the linearization $\vec{Q}$. If the linearization is updated every iteration ($K_{\mathrm{inner}} = 1$), the convergence is improved as no approximation is necessary but the computational complexity is increased. The optimal setting ($K_{\mathrm{outer}} = 30$, $K_{\mathrm{inner}} = 10$, $K_{\mathrm{cg}} = 3$) provides an excellent tradeoff between stable convergence and low computational complexity.

Figure~\ref{fig:convergenceLambda} shows the influence of the QuaSI regularization weight $\lambda$ to the convergence of our algorithm. We found that with decreasing $\lambda$, the PSNR and SSIM measures increase slower due to the low impact of the QuaSI prior. For the optimal setting $\lambda = 5.0$, we observed a fast convergence of our iteration scheme. Notice that further increasing $\lambda$ does not affect the convergence, which underlines effectiveness of the proposed QuaSI prior and the robustness of our iteration scheme.

Figure~\ref{fig:convergenceAlpha} depicts the influence of the Lagrangian multiplier $\alpha$, which enforces the constraint $\vec{u} = \vec{f} - \vec{Q}(\vec{f})$ in our ADMM optimization. For $\alpha \rightarrow \infty$, the augmented Lagrangian \eqref{eqn:augmentedLagrangian} results in the objective function \eqref{eqn:objective}. Hence, decreasing $\alpha$ impairs the convergence as shown by the peaks in the PSNR and SSIM measures over the iterations. Choosing $\alpha$ too large resulted in slower convergence compared to the proposed parameter setting $\alpha = 100$.

In order to show the interplay of the QuaSI regularization weight $\lambda$ and the corresponding Lagrangian multiplier $\alpha$ used for ADMM, Fig.~\ref{fig:sensitivityQuaSI} depicts the influence of different configurations to B-scan denoising using fixed TV parameters ($\mu = 0.075$, $\beta = 1.5$). We evaluated the denoising performance in terms of the PSNR and SSIM measures for a clinical relevant region showing retinal layers. Overall, we observed that increasing $\lambda$ and thus the impact of QuaSI consistently improved denoising, whereas the sensitivity against $\alpha$ is lower over several orders of magnitudes. Notice that our QuaSI prior was insensitive against oversmoothing as shown by the convergence of PSNR and SSIM for large $\lambda$.

\subsection{C-Arm Computed Tomography Denoising} 

C-arm computed tomography (CT) denotes an imaging modality where an X-ray source and detector are mounted on opposing sides of a C-shaped gantry. That gantry is further able to rotate around a patient lying on a table, thus allowing to acquire CT-like projection images. Using image reconstruction techniques \citep{zeng2010medical, strobel20093d}, these projection images can finally be transformed into a volumetric representation of the object under consideration.

Clinically, C-arm CT is both used for acquiring single volumetric images as well as for acquiring sequences of volumes, as it is for example used in perfusion imaging for acute stroke diagnosis \citep{univis91367737}. While single volumes just provide static information about the morphology itself, the acquisition of volume sequences typically involves injection of contrast agent during the acquisition, thus making the volume sequences provide additional temporal information.

Similar to conventional CT, photon effects as well as patient movement and angular undersampling usually deteriorate the image quality by introducing both structured and unstructured noise, see Fig.~\ref{figure:ct_syn} b, Fig.~\ref{figure:ct_real} a.

For our experiments, the noise $\vec{n}$ in reconstructed CT volumes is modeled as additive noise according to \eqref{eqn:noiseAdd}, and is further composed of both shot noise $\vec{p}$ \cite{} and structured noise $\vec{s}$, \ie
\begin{equation}
	\vec{n} = \vec{p} + \vec{s}.
\end{equation}

While shot noise in the acquired projection data results from fluctuations measured by the sensor, various processing steps during the reconstruction process complicate an exact statistical description of the noise in the resulting volumetric data \citep{Fessler}. Structured noise comes in the form of high-frequent streak artifacts, causes by angular undersampling.

\subsubsection{Datasets}
\label{sec:CTData}

We applied the proposed denoising algorithm on simulated C-arm CT data as well as on acquired, real patient data.

For our application, this results in two cases: single volumes can be denoised using \textit{volumetric denoising} (for the sake of convenience, we further refer to this method as SISO - single volume input, single volume output), while sequences of volumes are processed using \textit{volumetric + temporal denoising} (MIMO - multiple volume input, multiple volume output), cf. Fig.~\ref{fig:graphicalAbstract}.

 order to evaluate the denosing, we particularly investigated simulated data since it provides a known ground truth. The simulated data is based on a digital brain CT phantom \citep{aichert2013realistic}, which was used in combination with a simulation framework mimicking the acquisition process of a C-arm CT system \citep{univis91387862}. We added Poisson noise and simulated minor patient movement during the generation of the simulated data by rotating the head up to a total of \ang{5} around z-axis between the individual scans. After reconstructing the generated projection data, the individual volumes are co-registered again to assert pixel correspondence between the volumes. Due to the slight different positions of the head within individual volumes, the resulting streak artifacts slightly differ between the co-registered individual volumes.

For a numerical comparison of different algorithms, we calculate the peak signal to noise ratio (PSNR) and the structured similarity index measure (SSIM) \citep{wang2004image} using the digital phantom data as ground truth.

In addition to the simulated data, we also apply the proposed methods to real patient data which was clinically acquired during a perfusion imaging procedure.

\subsubsection{Comparison to the State-of-the-Art}

Current approaches towards noise reduction in CT imaging are, for example, based on anisotropic filtering or rely on a heuristic detection of streaks and vessel structures \citep{univis91578504, univis91131693, univis91420770}.

\begin{table}[!tb]
\begin{center}
\small
\begin{tabular}{lcccc}
\toprule
	&	\multirow{2}{*}{Input} 	& 	\multirow{2}{*}{BM4D}  	& QuaSI     & QuaSI \\
    &	 	    & 	  	    & (SISO)    & (MISO)  \\
\midrule
PSNR &	32.105 	& 	32.485 	& 32.462    & 34.788 \\
\midrule
SSIM &	0.883 	& 	0.914 	& 0.925     & 0.943	 \\ 
\bottomrule
\end{tabular}
\caption{PSNR and SSIM for the input data, BM4D \cite{maggioni2013nonlocal} and the QuaSI methods.}
\label{table:measures}
\end{center}
\end{table}

\begin{figure}[!tb]
	\centering
	\subfloat[Ground truth]{
     \includegraphics[width = 0.228\textwidth]{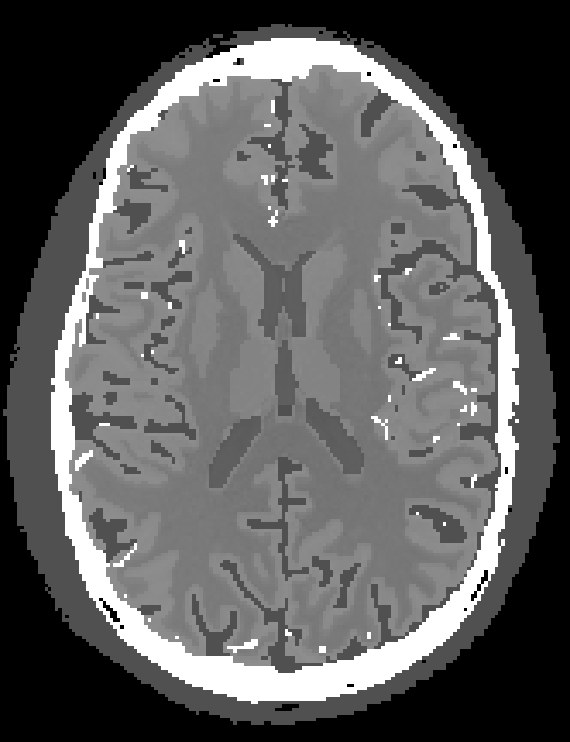}
  }
	\subfloat[Noisy input]{	
     \includegraphics[width = 0.228\textwidth]{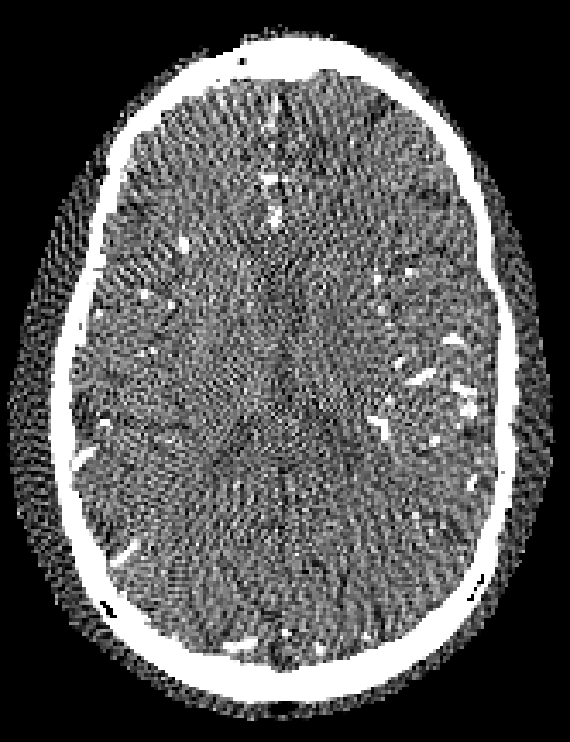}
  }\\
	\subfloat[SISO with QuaSI]{
     \includegraphics[width = 0.228\textwidth]{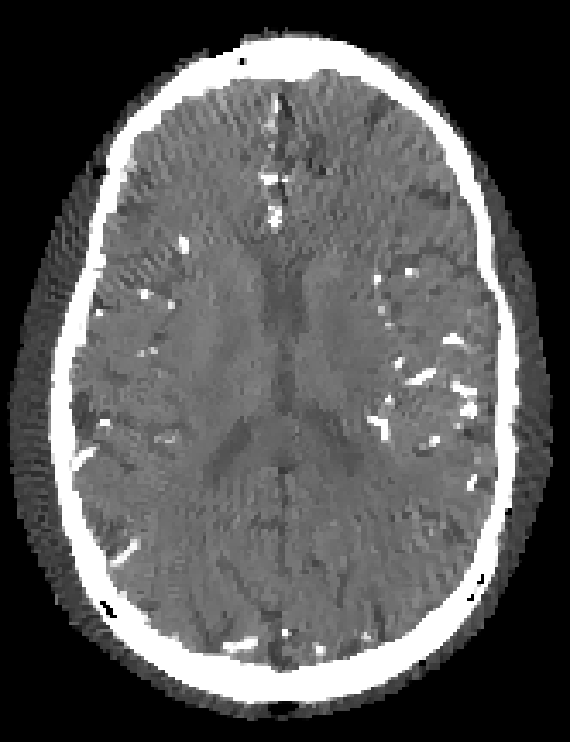}
  }
	\subfloat[SISO without QuaSI]{
     \includegraphics[width = 0.228\textwidth]{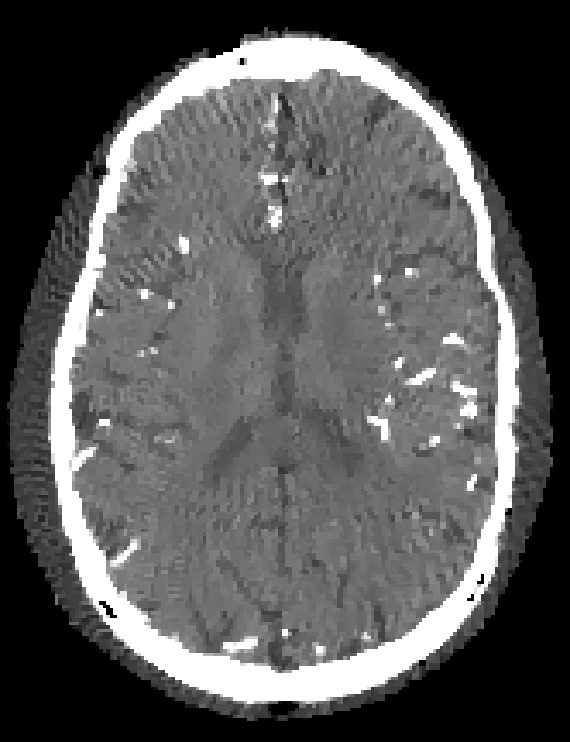}
  }\\
	\subfloat[MIMO with QuaSI]{
     \includegraphics[width = 0.228\textwidth]{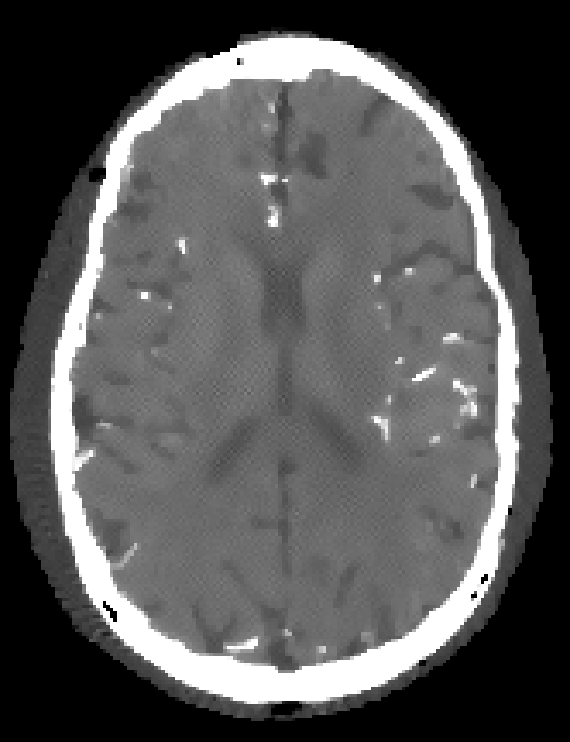}
  }
	\subfloat[MIMO without QuaSI]{
     \includegraphics[width = 0.228\textwidth]{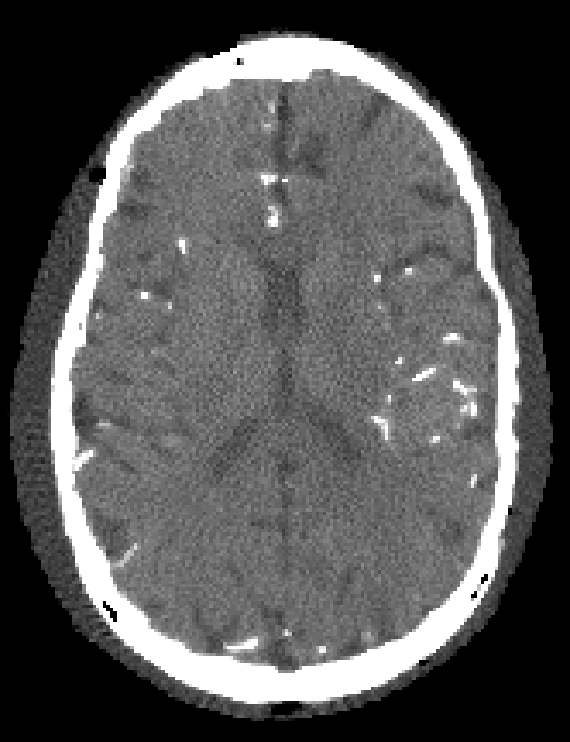}
  }

	\caption{Denoising on simulated C-arm CT data. (a) and (b) denote the ground truth data and the noisy input to the algorithm, respectively. (c) and (d) denote the denoised result with and without the QuaSI prior when using only a single volume (SISO) of the sequence. (e) and (f) denote the denoised result with and without the QuaSI prior, when using 1 volume (MISO). Note that for MISO, the input to the algorithm is not just the single volume as shown in the figure, but consists of a sequence of volumetric data.}
	\label{figure:ct_syn}
\end{figure}

\begin{figure*}[!tb]
	\subfloat[Noisy input]{
     \includegraphics[width = 0.32\textwidth]{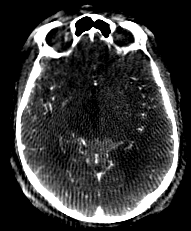}
  }
	\subfloat[SISO with QuaSI]{
     \includegraphics[width = 0.32\textwidth]{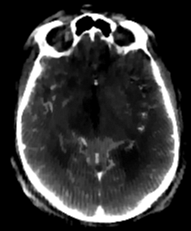}
  }
	\subfloat[MIMO with QuaSI]{
     \includegraphics[width = 0.32\textwidth]{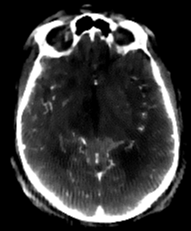}
  }
	\caption{Denoising on real clinical C-arm CT data. (a) denotes the noisy input, (b) denotes the denoised result when using only a single volume (SISO) of the sequence. (c) denotes the denoised result when using a volume sequence (MIMO). Note that for MIMO, the input to the algorithm is not just the single volume as shown in the figure, but consists of a sequence of volumetric data.}
	\label{figure:ct_real}
\end{figure*}

We compared the results from the proposed methods to the results from BM4D \citep{maggioni2013nonlocal}, which processes volumetric data and is an extension to the well-known BM3D \citep{Dabov2007}. We set the parameters of our method to $\alpha = 0.1 $, $ \lambda = 0.0005 $, $\beta = 0.1 $, $ \mu = 0.005$, $\gamma = 90 $ and $ \omega = 0.8$. These parameters have been optimized by investigating grid search on a small patch of the phantom data. The median filter regularization is computed on a $3\times3\times3$ kernel.

The algorithms are applied to and evaluated on a subset of the brain volume consisting of 30 consecutive slices. The slices, see Fig.~\ref{figure:ct_syn} for synthetic and Fig.~\ref{figure:ct_real} for the real data, show the complete head and contain all structures of interest such as bones, white matter, gray matter and (contrast-enhanced) vessels. The results from the evaluation of the realistic brain phantom show that the proposed denoising algorithm outperforms BM4D with regards to PSNR and SSIM, see Table \ref{table:measures}. Vessel structures are well-preserved within both volumes and boundaries between gray and white matter are perceivable. Further, a qualitative comparison between processed data with and without the use of the QuaSI prior (Fig.~\ref{figure:ct_syn} c,d and e,f) shows that the QuaSI prior is able to further lower the amount of noise in the volumetric image data.

\section{Conclusion}
\label{sec:conclusion}

In this paper, we have presented the quantile sparse image (QuaSI) prior and a corresponding spatio-temporal denoising algorithm suitable for volumetric OCT or CT data. For OCT denoising, we proposed two pipelines to either process B-scans or volumetric OCT data. The numerical optimization is derived using a linearization of the quantile filter and an alternating direction method of multipliers scheme for efficient minimization. We can show that a combination of QuaSI and Total Variation regularization outperforms state-of-the-art methods in terms of quantitative measures. Interestingly, our method can be applied to both CT and OCT data through minor modifications of the denoising pipeline. This suggests that it may be worthwhile to evaluate the potential of the QuaSI prior for inverse problems of other imaging modalities in future work.

\section*{References}

\bibliography{bibliography}

\end{document}

%% file: images/QuaSIInitialHisto.tikz
%
%
\definecolor{mycolor1}{rgb}{0.00000,0.44700,0.74100}%
\begin{tikzpicture}

\begin{axis}[%
width=0.951\figurewidth,
height=\figureheight,
at={(0\figurewidth,0\figureheight)},
scale only axis,
unbounded coords=jump,
xmin=-0.01,
xmax=0.2,
ymin=0,
ymax=200000,
axis background/.style={fill=white},
ylabel near ticks,
xlabel near ticks
]
\addplot[ybar interval, fill=red, fill opacity=0.6, draw=black, area legend] table[row sep=crcr] {%
x	y\\
0	96937\\
0.0252	40501\\
0.0504	32716\\
0.0756	17947\\
0.1008	12000\\
0.126	5355\\
0.1512	2713\\
0.1764	1499\\
0.2016	513\\
0.2268	238\\
0.252	89\\
0.2772	35\\
0.3024	20\\
0.3276	6\\
0.3528	4\\
0.378	1\\
0.4032	1\\
0.4284	0\\
0.4536	0\\
0.4788	0\\
0.504	0\\
0.5292	0\\
0.5544	0\\
0.5796	0\\
0.6048	0\\
0.63	0\\
0.6552	0\\
0.6804	0\\
0.7056	0\\
0.7308	0\\
0.756	0\\
0.7812	1\\
0.8064	1\\
};
\end{axis}
\end{tikzpicture}%

%% file: images/QuaSIGoldHisto.tikz
%
%
\definecolor{mycolor1}{rgb}{0.00000,0.44700,0.74100}%
\begin{tikzpicture}

\begin{axis}[%
width=0.951\figurewidth,
height=\figureheight,
at={(0\figurewidth,0\figureheight)},
scale only axis,
unbounded coords=jump,
xmin=-0.01,
xmax=0.2,
ymin=0,
ymax=200000,
axis background/.style={fill=white},
ylabel near ticks,
xlabel near ticks
]
\addplot[ybar interval, fill=red, fill opacity=0.6, draw=black, area legend] table[row sep=crcr] {%
x	y\\
0	197620\\
0.0112	12196\\
0.0224	680\\
0.0336	67\\
0.0448	7\\
0.056	2\\
0.0672	0\\
0.0784	0\\
0.0896	0\\
0.1008	0\\
0.112	0\\
0.1232	0\\
0.1344	0\\
0.1456	2\\
0.1568	0\\
0.168	0\\
0.1792	0\\
0.1904	0\\
0.2016	0\\
0.2128	0\\
0.224	0\\
0.2352	0\\
0.2464	0\\
0.2576	0\\
0.2688	0\\
0.28	0\\
0.2912	0\\
0.3024	0\\
0.3136	1\\
0.3248	0\\
0.336	0\\
0.3472	1\\
0.3584	1\\
};
\end{axis}
\end{tikzpicture}%

%% file: images/pigEyePSNR_color.tikz
%
%
\definecolor{mycolor1}{rgb}{1.00000,0.00000,1.00000}%
\definecolor{mycolor2}{rgb}{0.00000,1.00000,1.00000}%
\begin{tikzpicture}

\begin{axis}[%
width=0.951\figurewidth,
height=\figureheight,
at={(0\figurewidth,0\figureheight)},
scale only axis,
unbounded coords=jump,
xmin=1,
xmax=13,
xlabel style={font=\color{white!15!black}},
xlabel={Number of input images},
ymin=18,
ymax=27.2,
ylabel style={font=\color{white!15!black}},
ylabel={PSNR value},
axis background/.style={fill=white},
xmajorgrids,
ymajorgrids,
ylabel near ticks,
xlabel near ticks
]

\addplot [color=black, line width=1.0pt, 
    	mark=asterisk, mark options={solid, black}]
  table[row sep=crcr]{%
1	16.5386277837934\\
2	19.1615762895684\\
3	20.482039584299\\
4	21.3989032479102\\
5	22.0953951113771\\
6	22.6467361999872\\
7	23.0884851205616\\
8	23.4283059319348\\
9	23.7046554521625\\
10	23.9364926639876\\
11	24.1398788269585\\
12	24.313449842281\\
13	24.4396866945325\\
};

\addplot [color=blue, line width=1.0pt, 
    	mark=o, mark options={solid, blue}]
  table[row sep=crcr]{%
1	21.0828895398098\\
2	22.9514984723884\\
3	23.7838814122546\\
4	24.4035366571139\\
5	24.9421137309823\\
6	25.3150482291713\\
7	25.6293952924131\\
8	25.8538037915464\\
9	26.0154362023432\\
10	26.1497607704653\\
11	26.2642144348862\\
12	26.3687888986331\\
13	26.4190804923461\\
};

\addplot [color=mycolor1, line width=1.0pt, 
    	mark=diamond, mark options={solid, mycolor1}]
  table[row sep=crcr]{%
1	18.2540649439624\\
2	21.5533063711978\\
3	23.2667215651121\\
4	24.3244618555511\\
5	25.1344335441491\\
6	25.5162619812555\\
7	25.8302270969251\\
8	26.0598863205715\\
9	26.2571031088698\\
10	26.281622593084\\
11	26.3530097535537\\
12	26.4228841132997\\
13	26.4318238698356\\
};

\addplot [color=purple, line width=1pt,
    	mark = x, mark options = {solid,purple}]
  table[row sep=crcr]{%
1	nan\\
2	22.7434793490771\\
3	23.9291971023222\\
4	24.5575251579328\\
5	25.2062127248227\\
6	25.6215288766855\\
7	26.001948971497\\
8	26.2376103831663\\
9	26.4292287014248\\
10	26.5678613477309\\
11	26.6981267003153\\
12	26.8000603282638\\
13	26.8532531092616\\
};


\addplot [dashed,color=cyan, line width=1pt,
    	mark = x, mark options = {solid, cyan}]
  table[row sep=crcr]{%
1	16.7679111800535\\
2	19.3600695383406\\
3	20.652224680218\\
4	21.5545751638624\\
5	22.2426760464667\\
6	22.4652603000783\\
7	22.9179738203079\\
8	23.2652002241451\\
9	23.5515056726797\\
10	23.7872720722677\\
11	23.9988174050703\\
12	24.1845111114903\\
13	24.3223512371345\\
};

\addplot [dashed,color=green, line width=1pt,
    	mark = asterisk, mark options = {solid,green}]
  table[row sep=crcr]{%
1	18.1993638121143\\
2	21.6279560248122\\
3	23.2640140235146\\
4	24.3070105772002\\
5	25.0980829728877\\
6	25.6484312321899\\
7	26.0747298616096\\
8	26.3683527513876\\
9	26.5728248420968\\
10	26.7325269001144\\
11	26.860636069028\\
12	26.9707480462695\\
13	27.0190642276234\\
};

%

\addplot [color=red, line width=1pt,
    	mark=+, mark options = {solid, red}]
  table[row sep=crcr]{%
1	20.4023093591782\\
2	22.9045652780995\\
3	24.0567143022646\\
4	24.8485992085579\\
5	25.5073161359782\\
6	25.965476314597\\
7	26.3402128664471\\
8	26.5872122166017\\
9	26.7580395196329\\
10	26.9032740953736\\
11	27.011126088124\\
12	27.1121749105158\\
13	27.1425913959633\\
};
\end{axis}
\end{tikzpicture}%

%% file: images/pigEyeSSIM_color.tikz
%
%
\definecolor{mycolor1}{rgb}{1.00000,0.00000,1.00000}%
\definecolor{mycolor2}{rgb}{0.00000,1.00000,1.00000}%
\begin{tikzpicture}

\begin{axis}[%
width=0.951\figurewidth,
height=\figureheight,
at={(0\figurewidth,0\figureheight)},
scale only axis,
unbounded coords=jump,
xmin=1,
xmax=13,
xlabel style={font=\color{white!15!black}},
xlabel={Number of input images},
ymin=0.2,
ymax=0.87,
ylabel style={font=\color{white!15!black}},
ylabel={SSIM value},
axis background/.style={fill=white},
xmajorgrids,
ymajorgrids,
ylabel near ticks,
xlabel near ticks
]
\addplot [color=black, line width=1.0pt, 
    	mark=asterisk, mark options={solid, black}]
  table[row sep=crcr]{%
1	0.102486408885925\\
2	0.178325101393926\\
3	0.234889010574171\\
4	0.28160283817807\\
5	0.319428658177536\\
6	0.353535672646312\\
7	0.382232617800143\\
8	0.406852841781782\\
9	0.429053562941165\\
10	0.448508241169582\\
11	0.466462672081466\\
12	0.482479422644846\\
13	0.496828047370727\\
};

\addplot [color=blue, line width=1.0pt, 
    	mark=o, mark options={solid, blue}]
  table[row sep=crcr]{%
1	0.322289801065929\\
2	0.435554478207862\\
3	0.499934303387252\\
4	0.545379919219738\\
5	0.579896365729543\\
6	0.607003739260447\\
7	0.628578614777951\\
8	0.645984825770079\\
9	0.660765483172825\\
10	0.673188874662125\\
11	0.684230137848117\\
12	0.694095716755749\\
13	0.702501788167218\\
};

\addplot [color=mycolor1, line width=1.0pt, 
    	mark=diamond, mark options={solid, mycolor1}]
  table[row sep=crcr]{%
1	0.217933797216625\\
2	0.42126217428868\\
3	0.559925912302961\\
4	0.653317033597561\\
5	0.71648676298798\\
6	0.760560358545084\\
7	0.791173829410309\\
8	0.813444741449791\\
9	0.829747598201151\\
10	0.841927353022601\\
11	0.851402150890024\\
12	0.858160357678715\\
13	0.863183636320885\\
};

\addplot [color=purple, line width=1pt,
    	mark = x, mark options = {solid,purple}]
  table[row sep=crcr]{%
1	nan\\
2	0.429790108508575\\
3	0.515232010609527\\
4	0.565332339209876\\
5	0.611925662537554\\
6	0.645182166478127\\
7	0.674988980907526\\
8	0.696204517154726\\
9	0.716378612171376\\
10	0.73177763168603\\
11	0.74639276664247\\
12	0.757778180900044\\
13	0.768288819611982\\
};


\addplot [dashed,color=cyan, line width=1pt,
    	mark = x, mark options = {solid, cyan}]
  table[row sep=crcr]{%
1	0.107983525525051\\
2	0.187358737544676\\
3	0.245987214452808\\
4	0.294008722448088\\
5	0.333055131138383\\
6	0.364634008169541\\
7	0.395518628129191\\
8	0.421161778220443\\
9	0.444046735228563\\
10	0.463712921774257\\
11	0.481615427698713\\
12	0.497793172610035\\
13	0.512021330509773\\
};

\addplot [dashed,color=green, line width=1pt,
    	mark = asterisk, mark options = {solid,green}]
  table[row sep=crcr]{%
1	0.218272054106962\\
2	0.401053963850595\\
3	0.519113299422269\\
4	0.60062070673486\\
5	0.658848131686709\\
6	0.702334745556914\\
7	0.734259664789017\\
8	0.757986800965255\\
9	0.77667063385112\\
10	0.79145369240135\\
11	0.803413156303397\\
12	0.81307481193485\\
13	0.820894502210874\\
};
%

\addplot [color=red, line width=1pt,
    	mark=+, mark options = {solid, red}]
  table[row sep=crcr]{%
1	0.338865398032246\\
2	0.511566480621375\\
3	0.611349419762898\\
4	0.673926500971098\\
5	0.7176909317105\\
6	0.749081760158491\\
7	0.772408944097152\\
8	0.789471071148647\\
9	0.803863223861848\\
10	0.814634488620381\\
11	0.824027745402467\\
12	0.831156829035723\\
13	0.837460524722121\\
};
\end{axis}
\end{tikzpicture}%

%% file: images/clinicalMSR_color.tikz
%
%
\definecolor{mycolor1}{rgb}{1.00000,0.00000,1.00000}%
\definecolor{mycolor2}{rgb}{0.00000,1.00000,1.00000}%
\begin{tikzpicture}

\begin{axis}[%
width=0.951\figurewidth,
height=\figureheight,
at={(0\figurewidth,0\figureheight)},
scale only axis,
unbounded coords=jump,
xmin=1,
xmax=5,
xlabel style={font=\color{white!15!black}},
xlabel={Number of input images},
ymin=2.0,
ymax=6.7,
ylabel style={font=\color{white!15!black}},
ylabel={MSR value},
axis background/.style={fill=white},
xmajorgrids,
ymajorgrids,
ylabel near ticks,
xlabel near ticks
]

\addplot [color=black, line width=1.0pt, 
    	mark=asterisk, mark options={solid, black}]
  table[row sep=crcr]{%
1	2.69135319071544\\
2	3.02856779038749\\
3	3.2369253957581\\
4	3.35523948335106\\
5	3.42550788242545\\
};

\addplot [color=blue, line width=1.0pt, 
    	mark=o, mark options={solid, blue}]
  table[row sep=crcr]{%
1	5.04405015779332\\
2	5.28495214876942\\
3	5.38444626864553\\
4	5.42395901809518\\
5	5.44460822191596\\
};

\addplot [color=mycolor1, line width=1.0pt, 
    	mark=diamond, mark options={solid, mycolor1}]
  table[row sep=crcr]{%
1	4.92048830855562\\
2	5.89194788453228\\
3	6.38067326553359\\
4	6.59163000064814\\
5	6.65247496604004\\
};

\addplot [color=purple, line width=1pt,
    	mark = x, mark options = {solid,purple}]
  table[row sep=crcr]{%
1	nan\\
2	3.73617753813608\\
3	4.0258216115159\\
4	4.12151088274264\\
5	4.17830645158325\\
};


\addplot [dashed,color=cyan, line width=1pt,
    	mark = x, mark options = {solid, cyan}]
  table[row sep=crcr]{%
1	3.68337871999979\\
2	3.83132017773994\\
3	3.92473866128079\\
4	3.98267989222854\\
5	4.00672481019335\\
};

\addplot [dashed,color=green, line width=1pt,
    	mark = asterisk, mark options = {solid,green}]
  table[row sep=crcr]{%
1	4.89137811695109\\
2	5.60853586133739\\
3	5.93460919677672\\
4	6.04645740399313\\
5	6.02595350432043\\
};
%

\addplot [color=red, line width=1pt,
    	mark=+, mark options = {solid, red}]
  table[row sep=crcr]{%
1	5.33673212144268\\
2	5.83210938824553\\
3	6.11804622535642\\
4	6.2695224743868\\
5	6.31201509363651\\
};
\end{axis}
\end{tikzpicture}%


%% file: images/clinicalCNR_color.tikz
%
%
\definecolor{mycolor1}{rgb}{1.00000,0.00000,1.00000}%
\definecolor{mycolor2}{rgb}{0.00000,1.00000,1.00000}%
\begin{tikzpicture}

\begin{axis}[%
width=0.951\figurewidth,
height=\figureheight,
at={(0\figurewidth,0\figureheight)},
scale only axis,
unbounded coords=jump,
xmin=1,
xmax=5,
xlabel style={font=\color{white!15!black}},
xlabel={Number of input images},
ymin=2.0,
ymax=6.35,
ylabel style={font=\color{white!15!black}},
ylabel={CNR value},
axis background/.style={fill=white},
xmajorgrids,
ymajorgrids,
ylabel near ticks,
xlabel near ticks
]

\addplot [color=black, line width=1.0pt, 
    	mark=asterisk, mark options={solid, black}]
  table[row sep=crcr]{%
1	2.39633828189691\\
2	2.80104402105129\\
3	3.03605871701743\\
4	3.17785782018122\\
5	3.26693561030501\\
};

\addplot [color=blue, line width=1.0pt, 
    	mark=o, mark options={solid, blue}]
  table[row sep=crcr]{%
1	4.80355992065022\\
2	5.09852674975406\\
3	5.21619776817721\\
4	5.27794246949331\\
5	5.31133492806594\\
};

\addplot [color=mycolor1, line width=1.0pt, 
    	mark=diamond, mark options={solid, mycolor1}]
  table[row sep=crcr]{%
1	4.83640747296113\\
2	5.71724400617784\\
3	6.12033209088628\\
4	6.2878601057812\\
5	6.34121694628343\\
};

\addplot [color=purple, line width=1pt,
    	mark = x, mark options = {solid,purple}]
  table[row sep=crcr]{%
1	nan\\
2	3.61508676137382\\
3	3.90047118845621\\
4	4.01473071201931\\
5	4.07609488378815\\
};


\addplot [dashed,color=cyan, line width=1pt,
    	mark = x, mark options = {solid, cyan}]
  table[row sep=crcr]{%
1	3.6012663876337\\
2	3.77863022169093\\
3	3.87834191242554\\
4	3.93854144364467\\
5	3.9628945826\\
};

\addplot [dashed,color=green, line width=1pt,
    	mark = asterisk, mark options = {solid,green}]
  table[row sep=crcr]{%
1	4.81399924147969\\
2	5.43521253137376\\
3	5.69988977246005\\
4	5.79071042495071\\
5	5.78580462114144\\
};


\addplot [color=red, line width=1pt,
    	mark=+, mark options = {solid, red}]
  table[row sep=crcr]{%
1	5.24445172529597\\
2	5.75622427790219\\
3	6.01989377038856\\
4	6.14143484282207\\
5	6.18220947302432\\
};

\end{axis}
\end{tikzpicture}%

%% file: images/conv_PSNR_iter.tikz
%
%
\begin{tikzpicture}

\definecolor{gray4}{rgb}{0.2431,0.2431,0.2431}
\definecolor{gray3}{rgb}{0.3098,0.3098,0.3098}
\definecolor{gray2}{rgb}{0.4863,0.4863,0.4863}
\definecolor{gray1}{rgb}{0.7725,0.7725,0.7725}

\begin{axis}[%
width=0.951\figurewidth,
height=\figureheight,
at={(0\figurewidth,0\figureheight)},
scale only axis,
unbounded coords=jump,
xmin=0,
xmax=300,
ymin=24.5,
ymax=29.5,
ylabel style={font=\color{white!15!black}},
ylabel={PSNR},
xlabel style={font=\color{white!15!black}},
xlabel={Number of inner and outer iterations},
axis background/.style={fill=white},
xmajorgrids,
ymajorgrids,
xlabel near ticks,
ylabel near ticks
]
\addplot [color=black, line width=0.75pt, forget plot]
  table[row sep=crcr]{%
1	24.661462740074\\
2	24.9321012952417\\
3	25.3357945031032\\
4	25.7331697773387\\
5	26.0091794206517\\
6	26.0432311918526\\
7	26.0432311918526\\
8	26.1189459100779\\
9	26.1809963127787\\
10	26.1809963127787\\
11	26.4065956756952\\
12	26.5241257475114\\
13	26.7024329776134\\
14	26.9092666521803\\
15	27.0884279107541\\
16	27.1454110929849\\
17	27.1675096114449\\
18	27.1675096114449\\
19	27.1675096114449\\
20	27.1675096114449\\
21	27.3270721169274\\
22	27.4589899156044\\
23	27.4769349943889\\
24	27.6031497597997\\
25	27.6652577707251\\
26	27.6652577707251\\
27	27.7589179727578\\
28	27.7663442015763\\
29	27.7770750278091\\
30	27.7770750278091\\
31	27.8746789684252\\
32	27.9358185590299\\
33	27.9426473812919\\
34	27.9426473812919\\
35	28.0005846220551\\
36	28.0197615052225\\
37	28.0515890809355\\
38	28.080260491069\\
39	28.0887816091496\\
40	28.0933930682979\\
41	28.1469793257639\\
42	28.1813984622714\\
43	28.1849675224504\\
44	28.1849675224504\\
45	28.2103697269095\\
46	28.2199829959291\\
47	28.2240897080484\\
48	28.2286496912577\\
49	28.2286496912577\\
50	28.2286496912577\\
51	28.2768903698878\\
52	28.2908508085315\\
53	28.2940025240988\\
54	28.2940025240988\\
55	28.3055058655239\\
56	28.3140167726574\\
57	28.3343823614469\\
58	28.3421667156594\\
59	28.3439039858068\\
60	28.3439039858068\\
61	28.3740962477303\\
62	28.3857971557645\\
63	28.3886731670152\\
64	28.3893371966065\\
65	28.3893371966065\\
66	28.3893371966065\\
67	28.3893371966065\\
68	28.3916835758049\\
69	28.392978968135\\
70	28.392978968135\\
71	28.4070372480684\\
72	28.4156441881651\\
73	28.4174119044277\\
74	28.4174119044277\\
75	28.4116161934599\\
76	28.4151623880064\\
77	28.4161221621797\\
78	28.4161221621797\\
79	28.4161221621797\\
80	28.4161221621797\\
81	28.4379217662407\\
82	28.4452381470354\\
83	28.4458643093591\\
84	28.4458643093591\\
85	28.4424296442057\\
86	28.44328352436\\
87	28.4438876981073\\
88	28.4441479535319\\
89	28.4441479535319\\
90	28.4465983099513\\
91	28.4616465326783\\
92	28.4648817965239\\
93	28.465058336535\\
94	28.465058336535\\
95	28.4496171786245\\
96	28.4487374554013\\
97	28.4473021061014\\
98	28.4448402580824\\
99	28.4448402580824\\
100	28.4448402580824\\
101	28.4637963073598\\
102	28.4628673806232\\
103	28.4609151648362\\
104	28.4603281790402\\
105	28.4601324060588\\
106	28.4601324060588\\
107	28.4601798794023\\
108	28.4601798794023\\
109	28.4601798794023\\
110	28.4603147543705\\
111	28.4658634083988\\
112	28.4652415972414\\
113	28.4649026769086\\
114	28.4649026769086\\
115	28.451189332241\\
116	28.449642456852\\
117	28.4486023226322\\
118	28.4462890704584\\
119	28.4445748910758\\
120	28.4442560005281\\
121	28.4572539428556\\
122	28.4566928575593\\
123	28.4550193376227\\
124	28.4544260023609\\
125	28.4540256209529\\
126	28.4540256209529\\
127	28.4540256209529\\
128	28.4540256209529\\
129	28.4540256209529\\
130	28.4540256209529\\
131	28.4590421570664\\
132	28.4587219629188\\
133	28.4584836470209\\
134	28.4584836470209\\
135	28.4479102595015\\
136	28.446180167657\\
137	28.446180167657\\
138	28.444214421114\\
139	28.4427573204529\\
140	28.4408827082938\\
141	28.4491892735283\\
142	28.4496640577661\\
143	28.4493640506329\\
144	28.4493640506329\\
145	28.4328165925794\\
146	28.4285245515492\\
147	28.4224543536884\\
148	28.4172757305859\\
149	28.4168529896008\\
150	28.4168529896008\\
151	28.4316008604439\\
152	28.4305997089892\\
153	28.4273961847431\\
154	28.4268378780291\\
155	28.4265832301787\\
156	28.4265832301787\\
157	28.4265832301787\\
158	28.4266187352579\\
159	28.4266187352579\\
160	28.42658497176\\
161	28.4327153715379\\
162	28.4332035168114\\
163	28.4329482770475\\
164	28.418893922508\\
165	28.4155424148013\\
166	28.4111677138874\\
167	28.4081154408482\\
168	28.4078342825801\\
169	28.4078342825801\\
170	28.4078342825801\\
171	28.4193860055315\\
172	28.4174191303163\\
173	28.4152044951845\\
174	28.4146030248282\\
175	28.4142782663104\\
176	28.4142782663104\\
177	28.4142782663104\\
178	28.4141948600786\\
179	28.4140257679285\\
180	28.4140257679285\\
181	28.4193953109161\\
182	28.4198570118008\\
183	28.4196746145212\\
184	28.4196746145212\\
185	28.4065455004148\\
186	28.4039941783403\\
187	28.4041014992216\\
188	28.4041014992216\\
189	28.4041014992216\\
190	28.4016476439046\\
191	28.4086445230235\\
192	28.4067106636054\\
193	28.4045453085952\\
194	28.4039666510797\\
195	28.4039666510797\\
196	28.4036510505211\\
197	28.403421785149\\
198	28.4030810820104\\
199	28.4027856649085\\
200	28.4027856649085\\
201	28.4078499006352\\
202	28.4060225423803\\
203	28.4021485834058\\
204	28.4017976881259\\
205	28.4017976881259\\
206	28.4017976881259\\
207	28.4020145750381\\
208	28.401903834992\\
209	28.4015185858676\\
210	28.4012274000956\\
211	28.4039448865431\\
212	28.4042470142986\\
213	28.4040111974711\\
214	28.3929182627293\\
215	28.3888202714229\\
216	28.3839995323089\\
217	28.3821014203336\\
218	28.3818677023448\\
219	28.381616608838\\
220	28.3814085167664\\
221	28.3912355195834\\
222	28.3903615261011\\
223	28.3888769246313\\
224	28.3884134459618\\
225	28.3881113619976\\
226	28.3881113619976\\
227	28.3881113619976\\
228	28.3766219596547\\
229	28.3754897713916\\
230	28.3741972790341\\
231	28.3846168908636\\
232	28.382801334828\\
233	28.3794866176603\\
234	28.3791148926632\\
235	28.3791148926632\\
236	28.3767221969342\\
237	28.3755417896334\\
238	28.3738794295375\\
239	28.3736412730622\\
240	28.3733750446014\\
241	28.3807254105159\\
242	28.3781004402127\\
243	28.3777181479369\\
244	28.3777181479369\\
245	28.3773902992663\\
246	28.3771144957122\\
247	28.3771144957122\\
248	28.3771144957122\\
249	28.3768244340429\\
250	28.3765452226514\\
251	28.3804320856303\\
252	28.3805348165759\\
253	28.3802883431126\\
254	28.3802883431126\\
255	28.3802883431126\\
256	28.3802883431126\\
257	28.3655045955205\\
258	28.3634193704615\\
259	28.3635119210757\\
260	28.3603923022777\\
261	28.3718206247713\\
262	28.3720961892261\\
263	28.3719783510115\\
264	28.3719783510115\\
265	28.3719783510115\\
266	28.3719783510115\\
267	28.3721504382805\\
268	28.3721029638899\\
269	28.3721029638899\\
270	28.3720000917789\\
271	28.371359657877\\
272	28.3708149187223\\
273	28.3705884593052\\
274	28.3620822107444\\
275	28.3606316537249\\
276	28.3594533833691\\
277	28.358119182289\\
278	28.3577190490896\\
279	28.3575836204223\\
280	28.3575836204223\\
281	28.3665490613888\\
282	28.3664323596016\\
283	28.3662755293831\\
284	28.3662755293831\\
285	28.3662755293831\\
286	28.3662755293831\\
287	28.3662755293831\\
288	28.3554897643825\\
289	28.3557836637057\\
290	28.3560772641361\\
291	28.3673976335766\\
292	28.3669945005484\\
293	28.3658070173121\\
294	28.3658070173121\\
295	28.3646630739636\\
296	28.3642772415251\\
297	28.3640939027895\\
298	28.3640939027895\\
299	28.3640939027895\\
300	28.3640487529391\\
};
\addplot [color=blue, line width=0.75pt, forget plot]
  table[row sep=crcr]{%
1	24.661462740074\\
2	25.0718915958892\\
3	25.3331751976926\\
4	25.5187694273509\\
5	25.6625378915663\\
6	25.7798373130763\\
7	25.8820100216882\\
8	25.9700410210012\\
9	26.0469427815898\\
10	26.1155889806409\\
11	26.1774221328637\\
12	26.2336308862589\\
13	26.2842281906837\\
14	26.3329196754995\\
15	26.3771628048517\\
16	26.4202381109034\\
17	26.4604890479358\\
18	26.4986644976816\\
19	26.5341158629081\\
20	26.5683888213717\\
21	26.6029420754011\\
22	26.6354719992278\\
23	26.6675864443909\\
24	26.696881733158\\
25	26.7282497538091\\
26	26.7575210824302\\
27	26.7854742687369\\
28	26.8127039984324\\
29	26.8396777544794\\
30	26.8655779731928\\
31	26.8907875801391\\
32	26.915839904648\\
33	26.9392954324168\\
34	26.9648654293844\\
35	26.9879282623812\\
36	27.0114471211367\\
37	27.0331221524081\\
38	27.0557892849406\\
39	27.0768143184626\\
40	27.0976366929525\\
41	27.1182162765658\\
42	27.1387776495697\\
43	27.159242809712\\
44	27.1786184087696\\
45	27.1977214987411\\
46	27.2160028651011\\
47	27.2345483745885\\
48	27.2518650653577\\
49	27.2708794553419\\
50	27.2879416007472\\
51	27.3048773167566\\
52	27.3219174322874\\
53	27.338045527936\\
54	27.3539826748444\\
55	27.3696558703743\\
56	27.3856795936448\\
57	27.4004101911893\\
58	27.4163877564056\\
59	27.4309885766847\\
60	27.4458361856383\\
61	27.4605102189619\\
62	27.4746297238238\\
63	27.488917981691\\
64	27.5021051375253\\
65	27.5163787300587\\
66	27.5291363113861\\
67	27.542121702966\\
68	27.554521936259\\
69	27.5669814928766\\
70	27.5790244253912\\
71	27.5922977730305\\
72	27.60453172561\\
73	27.6160844431128\\
74	27.6277065006971\\
75	27.639053562019\\
76	27.6498301049577\\
77	27.6601644260598\\
78	27.6718969239725\\
79	27.6823978594988\\
80	27.6931345325656\\
81	27.7030940658363\\
82	27.713120539429\\
83	27.7225082280851\\
84	27.732315338383\\
85	27.7414677478104\\
86	27.751404490554\\
87	27.7608822956515\\
88	27.7693334928795\\
89	27.7789263706163\\
90	27.7881112670018\\
91	27.7970887669601\\
92	27.8050624174025\\
93	27.8135079635134\\
94	27.8220053759888\\
95	27.8299737333199\\
96	27.8383076157757\\
97	27.8458463600772\\
98	27.8537642637122\\
99	27.8618091922317\\
100	27.8694946487212\\
101	27.8767672503044\\
102	27.8842102793559\\
103	27.891105763199\\
104	27.8983786662845\\
105	27.9055994659774\\
106	27.9120425824321\\
107	27.9189743352687\\
108	27.9256197859596\\
109	27.9321083814507\\
110	27.9382450143421\\
111	27.9444945022786\\
112	27.9508473622033\\
113	27.9570261037246\\
114	27.9631812523161\\
115	27.9691938627381\\
116	27.9746671799921\\
117	27.9803756624869\\
118	27.986227528182\\
119	27.9916893334565\\
120	27.997242980796\\
121	28.0025497120581\\
122	28.0077946727858\\
123	28.0131098620658\\
124	28.0182049252512\\
125	28.0228349646861\\
126	28.028197734842\\
127	28.0332340926729\\
128	28.0383624345863\\
129	28.042897372276\\
130	28.0479305560987\\
131	28.0530909689152\\
132	28.0576872209516\\
133	28.0619053490234\\
134	28.0662158989018\\
135	28.0704456515404\\
136	28.0748234759452\\
137	28.0790776130944\\
138	28.0829614358028\\
139	28.0872527645027\\
140	28.0913603388567\\
141	28.0954538973943\\
142	28.0993340194074\\
143	28.1032536276967\\
144	28.1069554025461\\
145	28.110700587931\\
146	28.1144742601979\\
147	28.1180043024829\\
148	28.1215855518507\\
149	28.1250152266487\\
150	28.1282776097149\\
151	28.1317116783042\\
152	28.1348216703222\\
153	28.1381859083213\\
154	28.1416782449696\\
155	28.1448631186828\\
156	28.147696925274\\
157	28.1511754468652\\
158	28.1544870995217\\
159	28.1573312125592\\
160	28.1603295376535\\
161	28.1631072541153\\
162	28.1659941297623\\
163	28.1691236459155\\
164	28.1717550475729\\
165	28.1747246458954\\
166	28.1774310665616\\
167	28.1797594632473\\
168	28.1823859875834\\
169	28.1850050976802\\
170	28.1876769090951\\
171	28.1902187600244\\
172	28.192748692802\\
173	28.1952339481161\\
174	28.1976099525556\\
175	28.2000092357437\\
176	28.202548434102\\
177	28.2047306962874\\
178	28.2069666347382\\
179	28.2092179304476\\
180	28.2112579617708\\
181	28.2136998539833\\
182	28.2160108552436\\
183	28.2182296896436\\
184	28.220382656255\\
185	28.2224214165543\\
186	28.2245804579225\\
187	28.2264267457795\\
188	28.2284134464775\\
189	28.2301717347513\\
190	28.2322010677062\\
191	28.2341179268848\\
192	28.2358978381024\\
193	28.2376619035817\\
194	28.2394274617823\\
195	28.2412725821591\\
196	28.2428806935628\\
197	28.2447175327643\\
198	28.2462102971547\\
199	28.2479050102725\\
200	28.2496167191098\\
201	28.2514149842073\\
202	28.2530433595834\\
203	28.2545067710502\\
204	28.2561342428942\\
205	28.2576292183043\\
206	28.2591274156114\\
207	28.2605847810137\\
208	28.2620221168001\\
209	28.2634726882599\\
210	28.2649895841984\\
211	28.2665061116671\\
212	28.2678530947241\\
213	28.2691798154897\\
214	28.270341544185\\
215	28.2718573381658\\
216	28.2731390603948\\
217	28.2744292235653\\
218	28.2756194864027\\
219	28.2768500920353\\
220	28.2781030590503\\
221	28.2792481409369\\
222	28.2805078597969\\
223	28.2816771280974\\
224	28.2827829822966\\
225	28.2839218792017\\
226	28.2852562738632\\
227	28.2863555413404\\
228	28.2873779030541\\
229	28.2883770956939\\
230	28.2894420821339\\
231	28.2904343840699\\
232	28.2913506837419\\
233	28.2925164957843\\
234	28.2933337669883\\
235	28.2943135779784\\
236	28.2952202691224\\
237	28.2960536762207\\
238	28.2969819744781\\
239	28.2979506796752\\
240	28.2988872432877\\
241	28.2999348499491\\
242	28.3007196034769\\
243	28.3016822227148\\
244	28.3025431252709\\
245	28.3032965295133\\
246	28.3042644421906\\
247	28.3049954715343\\
248	28.3057384536278\\
249	28.3065689041717\\
250	28.3071931495387\\
251	28.30791444635\\
252	28.3085077262854\\
253	28.309294636706\\
254	28.3099709430922\\
255	28.3107010060708\\
256	28.3117094726508\\
257	28.3122077409125\\
258	28.3130325109627\\
259	28.3136764510774\\
260	28.3143999156827\\
261	28.3150065486942\\
262	28.315703692717\\
263	28.3162160255513\\
264	28.3168942443747\\
265	28.3175492272833\\
266	28.3181001332982\\
267	28.3185598934645\\
268	28.319102792056\\
269	28.319754010691\\
270	28.3201499414473\\
271	28.3207692747372\\
272	28.3214644799849\\
273	28.3220212246377\\
274	28.3224528528173\\
275	28.323055022013\\
276	28.3235649041544\\
277	28.3241765647286\\
278	28.3245512680015\\
279	28.3249621016148\\
280	28.3254705201898\\
281	28.3260405818362\\
282	28.3264411315939\\
283	28.3267927637073\\
284	28.3274367076679\\
285	28.3277566306659\\
286	28.3281414899719\\
287	28.3285418163639\\
288	28.3290651285022\\
289	28.3295787802979\\
290	28.3301381913917\\
291	28.3305088336019\\
292	28.3309794304569\\
293	28.3313892662748\\
294	28.331879632949\\
295	28.3321472280642\\
296	28.3326279442517\\
297	28.3329851752477\\
298	28.3334398440292\\
299	28.3336888354862\\
300	28.3340163144245\\
};
\addplot [color=red, line width=0.75pt, forget plot]
  table[row sep=crcr]{%
1	27.9367617247019\\
2	29.1603867016982\\
3	28.9624872715291\\
4	28.7375099726237\\
5	28.6240507592432\\
6	28.5721005624032\\
7	28.5480887062899\\
8	28.5480887062899\\
9	28.5480887062899\\
10	28.5480887062899\\
11	28.3400687508267\\
12	28.5184695448467\\
13	28.5414282078826\\
14	28.5244512607411\\
15	28.5094901024184\\
16	28.5000126144908\\
17	28.4947489328328\\
18	28.4919967865178\\
19	28.4907129142837\\
20	28.4902355658369\\
21	28.332284194232\\
22	28.476403583535\\
23	28.4999610991002\\
24	28.4906883654775\\
25	28.4809318569665\\
26	28.4747060968339\\
27	28.471177188781\\
28	28.4693113519879\\
29	28.4684884809344\\
30	28.4681939796685\\
31	28.3498537747794\\
32	28.4386841071144\\
33	28.451083404957\\
34	28.4447195738379\\
35	28.4376437239689\\
36	28.432631708901\\
37	28.4294915957735\\
38	28.4277192237258\\
39	28.4267642275764\\
40	28.4263006560487\\
41	28.3619656464716\\
42	28.4227797134369\\
43	28.4320479341828\\
44	28.4272159828352\\
45	28.4220074301837\\
46	28.4181475853898\\
47	28.4155998896277\\
48	28.4141445285672\\
49	28.4134126741851\\
50	28.4131015153817\\
51	28.3647754500266\\
52	28.4138815987577\\
53	28.4165441359528\\
54	28.4113395343109\\
55	28.4065241607924\\
56	28.4031540583332\\
57	28.4009870913234\\
58	28.3995648191049\\
59	28.3986779371774\\
60	28.3980986700062\\
61	28.3600380556681\\
62	28.3917934613863\\
63	28.3937094057892\\
64	28.3901331302388\\
65	28.3865792836313\\
66	28.3839058193599\\
67	28.3820793235472\\
68	28.3809751156964\\
69	28.380381014991\\
70	28.3800202484678\\
71	28.353374606969\\
72	28.3802650203781\\
73	28.3774407242663\\
74	28.3715089326139\\
75	28.3671765716341\\
76	28.3637515730317\\
77	28.3614718338663\\
78	28.3598946347615\\
79	28.3588251644459\\
80	28.3581106121186\\
81	28.3519795247656\\
82	28.3733140183683\\
83	28.3710198468998\\
84	28.3678505655099\\
85	28.3644733840044\\
86	28.3620943725551\\
87	28.360321541761\\
88	28.3592282179688\\
89	28.358540658666\\
90	28.358087955853\\
91	28.3403351494638\\
92	28.3606363629904\\
93	28.3574860199169\\
94	28.3530676419349\\
95	28.3499616596418\\
96	28.3478549426888\\
97	28.3464939275636\\
98	28.3457027036132\\
99	28.3453091649577\\
100	28.3450865884366\\
101	28.3357903102253\\
102	28.3535755328574\\
103	28.3493257470544\\
104	28.3451612252463\\
105	28.3418096847896\\
106	28.3394457323542\\
107	28.3378273047\\
108	28.3368309518024\\
109	28.3361914363909\\
110	28.3357160496564\\
111	28.3465213423681\\
112	28.3634571776575\\
113	28.3577447440612\\
114	28.3513411261691\\
115	28.3471128139007\\
116	28.3439864118912\\
117	28.3419036414289\\
118	28.340624193782\\
119	28.3397724869548\\
120	28.3391890834196\\
121	28.3429157543109\\
122	28.3552931334977\\
123	28.3503057476468\\
124	28.3451625239105\\
125	28.3414478091424\\
126	28.338834582703\\
127	28.3371871245222\\
128	28.3361557178523\\
129	28.3354761936556\\
130	28.3350645586535\\
131	28.3360962448681\\
132	28.3521539557525\\
133	28.3476277321664\\
134	28.3437105885473\\
135	28.3409690092146\\
136	28.3389809783201\\
137	28.3376780793066\\
138	28.3369197386688\\
139	28.3365077081041\\
140	28.3362407156856\\
141	28.3324734860865\\
142	28.3457482795917\\
143	28.3402915510271\\
144	28.3356825813771\\
145	28.3329426081938\\
146	28.3312133674632\\
147	28.3301312221231\\
148	28.3294945731987\\
149	28.3290778616286\\
150	28.3287677581018\\
151	28.3328416799145\\
152	28.3460937510154\\
153	28.3391286141283\\
154	28.3333206462898\\
155	28.3294739020622\\
156	28.3267868798641\\
157	28.3250226048861\\
158	28.3239040202173\\
159	28.3232008124868\\
160	28.3227559014147\\
161	28.3260214771951\\
162	28.338094061735\\
163	28.3317203155075\\
164	28.3269128695156\\
165	28.3237026076789\\
166	28.3214112322117\\
167	28.3199682124879\\
168	28.318873580013\\
169	28.3181209916848\\
170	28.3176251200035\\
171	28.323522233109\\
172	28.3371765692841\\
173	28.3318983404128\\
174	28.3274961922341\\
175	28.3245818283953\\
176	28.3223891578368\\
177	28.3207770267304\\
178	28.3197637886686\\
179	28.3190725440756\\
180	28.3185908142704\\
181	28.3222159695369\\
182	28.3319894657718\\
183	28.3247107599629\\
184	28.3183862213311\\
185	28.3143648051057\\
186	28.3119615426287\\
187	28.3102968939201\\
188	28.3092823414254\\
189	28.3086934814093\\
190	28.3083255051434\\
191	28.3273146322813\\
192	28.3423887330366\\
193	28.3363795944673\\
194	28.3308015852907\\
195	28.3268936330822\\
196	28.3243971434295\\
197	28.3229635250563\\
198	28.3221168366602\\
199	28.3215676876675\\
200	28.3211555866409\\
201	28.3236831574202\\
202	28.3338891806513\\
203	28.327572365954\\
204	28.3220819226577\\
205	28.3181184702953\\
206	28.3156517867721\\
207	28.3138761857391\\
208	28.3126824860377\\
209	28.3118968341699\\
210	28.3113227918552\\
211	28.3235850549385\\
212	28.3345951392767\\
213	28.3274172527399\\
214	28.3215656923807\\
215	28.3179130276511\\
216	28.31548290231\\
217	28.3136957883864\\
218	28.3125780295922\\
219	28.31184272183\\
220	28.3112877629412\\
221	28.3189867360181\\
222	28.3306984219698\\
223	28.3244725247546\\
224	28.3193479220432\\
225	28.3160469355427\\
226	28.3140296357867\\
227	28.3126388181321\\
228	28.3117166151377\\
229	28.3111240453927\\
230	28.3107347060901\\
231	28.3274191365219\\
232	28.3329888695151\\
233	28.3267400660933\\
234	28.3217789214371\\
235	28.318536866358\\
236	28.3164981351233\\
237	28.3150764022752\\
238	28.3141444712765\\
239	28.3135496534405\\
240	28.3131383744829\\
241	28.3202732037878\\
242	28.329035959263\\
243	28.3229096532041\\
244	28.3174794865006\\
245	28.313856401664\\
246	28.3116455822835\\
247	28.3101493952589\\
248	28.3091590156069\\
249	28.3084828788108\\
250	28.3080156672678\\
251	28.3153364542139\\
252	28.324373672791\\
253	28.3171595205336\\
254	28.3109017071988\\
255	28.3068104976887\\
256	28.3042984244167\\
257	28.3025723451592\\
258	28.3013803178225\\
259	28.3006423864277\\
260	28.3001288914038\\
261	28.3170390644106\\
262	28.3289163679961\\
263	28.3210514736553\\
264	28.3143611919583\\
265	28.3098615227773\\
266	28.3070958296902\\
267	28.3053328834944\\
268	28.3042551483758\\
269	28.303571604198\\
270	28.3031049002532\\
271	28.3101963074108\\
272	28.3186095540159\\
273	28.3103989741665\\
274	28.3040205912085\\
275	28.2999230525226\\
276	28.2971948403958\\
277	28.2953678043343\\
278	28.2940113341294\\
279	28.293123008764\\
280	28.2925257374201\\
281	28.3153652797408\\
282	28.3227716425126\\
283	28.3157284290365\\
284	28.3104918262397\\
285	28.3071751988517\\
286	28.3050600599738\\
287	28.3036547500937\\
288	28.3028328946353\\
289	28.302354977988\\
290	28.3020360937181\\
291	28.3120360411963\\
292	28.3230629564417\\
293	28.3171716051666\\
294	28.3111257089119\\
295	28.3068413826988\\
296	28.3040445255956\\
297	28.3022016655636\\
298	28.3010775125168\\
299	28.3002628631743\\
300	28.2997611683709\\
};
\end{axis}
\end{tikzpicture}%

%% file: images/conv_SSIM_iter.tikz
%
%
\begin{tikzpicture}

\begin{axis}[%
width=0.951\figurewidth,
height=\figureheight,
at={(0\figurewidth,0\figureheight)},
scale only axis,
unbounded coords=jump,
xmin=0,
xmax=300,
xlabel style={font=\color{white!15!black}},
xlabel={Number of inner and outer iterations},
ymin=0.4,
ymax=0.8,
ylabel style={font=\color{white!15!black}},
ylabel={SSIM},
axis background/.style={fill=white},
xmajorgrids,
ymajorgrids,
xlabel near ticks,
ylabel near ticks
]
\addplot [color=black, line width=0.8pt, forget plot]
  table[row sep=crcr]{%
1	0.447132792013231\\
2	0.470467194213733\\
3	0.490248031609507\\
4	0.535027981419754\\
5	0.566336266612457\\
6	0.570217827611425\\
7	0.570217827611425\\
8	0.579449268183018\\
9	0.587044608658933\\
10	0.587044608658933\\
11	0.610205495867125\\
12	0.621871728540799\\
13	0.630029587856091\\
14	0.652470300538481\\
15	0.67119704611202\\
16	0.677109027309769\\
17	0.679412838884696\\
18	0.679412838884696\\
19	0.679412838884696\\
20	0.679412838884696\\
21	0.695190440975232\\
22	0.707886748925659\\
23	0.709532172568961\\
24	0.718822792992596\\
25	0.724244099915571\\
26	0.724244099915571\\
27	0.733530213638752\\
28	0.734254116315106\\
29	0.735295343292766\\
30	0.735295343292766\\
31	0.744499819483657\\
32	0.750068173308777\\
33	0.750606704841783\\
34	0.750606704841783\\
35	0.753270707093871\\
36	0.755045220689387\\
37	0.758318634642998\\
38	0.761078297524092\\
39	0.761890147205961\\
40	0.762336065569206\\
41	0.767979039843398\\
42	0.770797538547217\\
43	0.771061883050198\\
44	0.771061883050198\\
45	0.771813175976829\\
46	0.772618267266076\\
47	0.773047311737532\\
48	0.773548545062137\\
49	0.773548545062137\\
50	0.773548545062137\\
51	0.778745053914404\\
52	0.779961957373213\\
53	0.780155912620025\\
54	0.780155912620025\\
55	0.779119491245776\\
56	0.779893562981844\\
57	0.781828011917944\\
58	0.782503084530719\\
59	0.782667379285663\\
60	0.782667379285663\\
61	0.786209982608074\\
62	0.787084616150734\\
63	0.787124805524764\\
64	0.78714930305371\\
65	0.78714930305371\\
66	0.78714930305371\\
67	0.78714930305371\\
68	0.787404545661127\\
69	0.787536038366196\\
70	0.787536038366196\\
71	0.789122534331431\\
72	0.78996323285451\\
73	0.790074572622523\\
74	0.790074572622523\\
75	0.788302156356146\\
76	0.78858553949244\\
77	0.788709339657087\\
78	0.788709339657087\\
79	0.788709339657087\\
80	0.788709339657087\\
81	0.791477689874253\\
82	0.792054517362402\\
83	0.792091767837102\\
84	0.792091767837102\\
85	0.791207576068801\\
86	0.791199097359645\\
87	0.791265760902294\\
88	0.79130962052825\\
89	0.79130962052825\\
90	0.791753698345446\\
91	0.793702654481453\\
92	0.794056475481657\\
93	0.794052198589591\\
94	0.794052198589591\\
95	0.791895754415146\\
96	0.79183289067391\\
97	0.791908662793226\\
98	0.79184748518783\\
99	0.79184748518783\\
100	0.79184748518783\\
101	0.794353318671433\\
102	0.7942679172279\\
103	0.794054708362647\\
104	0.794004713772065\\
105	0.793997835715867\\
106	0.793997835715867\\
107	0.794039122276083\\
108	0.794039122276083\\
109	0.794039122276083\\
110	0.794120532740398\\
111	0.794984920623837\\
112	0.794935201030131\\
113	0.794894645159267\\
114	0.794894645159267\\
115	0.793258798278337\\
116	0.793187611865869\\
117	0.793249291279803\\
118	0.793194766805497\\
119	0.793144465491107\\
120	0.793143417506321\\
121	0.794836094162868\\
122	0.794779440272492\\
123	0.794589480181835\\
124	0.794529284792092\\
125	0.794496246304627\\
126	0.794496246304627\\
127	0.794496246304627\\
128	0.794496246304627\\
129	0.794496246304627\\
130	0.794496246304627\\
131	0.79525551178117\\
132	0.795310231142414\\
133	0.795284847799119\\
134	0.795284847799119\\
135	0.794112924887263\\
136	0.794003456890253\\
137	0.794003456890253\\
138	0.793995203053402\\
139	0.793942677599529\\
140	0.793891182406847\\
141	0.79499948310758\\
142	0.795101243023207\\
143	0.795067891761545\\
144	0.795067891761545\\
145	0.793160931796938\\
146	0.792832229407174\\
147	0.792481648933633\\
148	0.79215360998837\\
149	0.792133376255702\\
150	0.792133376255702\\
151	0.793933637253725\\
152	0.793770002770642\\
153	0.793429509347751\\
154	0.793383450689017\\
155	0.793372887741453\\
156	0.793372887741453\\
157	0.793372887741453\\
158	0.793429134417145\\
159	0.793429134417145\\
160	0.793486231272121\\
161	0.794324349254487\\
162	0.794387345237614\\
163	0.794354569841142\\
164	0.792672579891897\\
165	0.792357806213331\\
166	0.792073559012295\\
167	0.791884968039795\\
168	0.791875968585127\\
169	0.791875968585127\\
170	0.791875968585127\\
171	0.79351650702484\\
172	0.793349768359026\\
173	0.793113533086732\\
174	0.793054990364656\\
175	0.79302956621935\\
176	0.79302956621935\\
177	0.79302956621935\\
178	0.793049652020435\\
179	0.793047932783218\\
180	0.793047932783218\\
181	0.793780491644019\\
182	0.793866778001666\\
183	0.793846134940115\\
184	0.793846134940115\\
185	0.792250247603716\\
186	0.792011333875157\\
187	0.792039201825789\\
188	0.792039201825789\\
189	0.792039201825789\\
190	0.792089491095986\\
191	0.793056347649183\\
192	0.792909012093651\\
193	0.792678790273154\\
194	0.792616165331443\\
195	0.792616165331443\\
196	0.792595465660655\\
197	0.792577273252588\\
198	0.792557956743862\\
199	0.792542141615398\\
200	0.792542141615398\\
201	0.793198239437432\\
202	0.792975048791918\\
203	0.792535196568275\\
204	0.79249356365333\\
205	0.79249356365333\\
206	0.79249356365333\\
207	0.792556937680042\\
208	0.792563355791159\\
209	0.792542112612998\\
210	0.79252023354776\\
211	0.792960926919698\\
212	0.793019543063134\\
213	0.792993239645123\\
214	0.791717885109042\\
215	0.791325378951222\\
216	0.790937699039247\\
217	0.790801497560456\\
218	0.790791139706209\\
219	0.790792026155445\\
220	0.790804817011009\\
221	0.792092772716621\\
222	0.791968926368928\\
223	0.791802149550463\\
224	0.791757237817072\\
225	0.791733470491548\\
226	0.791733470491548\\
227	0.791733470491548\\
228	0.790657568985387\\
229	0.790597224724551\\
230	0.790547329883946\\
231	0.791777576424568\\
232	0.791536765571781\\
233	0.791159550389917\\
234	0.791124239009408\\
235	0.791124239009408\\
236	0.790956846291393\\
237	0.790871471183391\\
238	0.790747369440493\\
239	0.790731160320907\\
240	0.7907161348744\\
241	0.791655525916124\\
242	0.79135221689913\\
243	0.791309093118666\\
244	0.791309093118666\\
245	0.79127693699061\\
246	0.791252397026468\\
247	0.791252397026468\\
248	0.791252397026468\\
249	0.791238626416618\\
250	0.791220012828491\\
251	0.791730011392013\\
252	0.791742829174857\\
253	0.791706854135392\\
254	0.791706854135392\\
255	0.791706854135392\\
256	0.791706854135392\\
257	0.789947370850091\\
258	0.789868411383993\\
259	0.789893065039324\\
260	0.789701901844259\\
261	0.79106005492991\\
262	0.79107553472209\\
263	0.791059356303777\\
264	0.791059356303777\\
265	0.791059356303777\\
266	0.791059356303777\\
267	0.791071165861906\\
268	0.791060523667248\\
269	0.791060523667248\\
270	0.791049568501909\\
271	0.791049182101187\\
272	0.791017535442623\\
273	0.790988540431967\\
274	0.790004661856935\\
275	0.789856118114032\\
276	0.789769060990227\\
277	0.789675669295487\\
278	0.7896530947487\\
279	0.789651418404465\\
280	0.789651418404465\\
281	0.790773107832284\\
282	0.790713566200282\\
283	0.790689446731163\\
284	0.790689446731163\\
285	0.790689446731163\\
286	0.790689446731163\\
287	0.790689446731163\\
288	0.789311856035371\\
289	0.789384114971487\\
290	0.789499407393605\\
291	0.790881039388471\\
292	0.790858405546271\\
293	0.790713470501543\\
294	0.790713470501543\\
295	0.790580567848491\\
296	0.790539987859783\\
297	0.790526877732289\\
298	0.790526877732289\\
299	0.790526877732289\\
300	0.790541475997178\\
};
\addplot [color=blue, line width=0.8pt, forget plot]
  table[row sep=crcr]{%
1	0.447132792013231\\
2	0.48227772842459\\
3	0.505533293925771\\
4	0.527238051646483\\
5	0.553626913057592\\
6	0.575093844575299\\
7	0.592384038202969\\
8	0.606175770619356\\
9	0.617705617609442\\
10	0.63297947277787\\
11	0.64677320550579\\
12	0.65924312856704\\
13	0.666135848256152\\
14	0.672562701922046\\
15	0.678077328475884\\
16	0.68363955496557\\
17	0.692127149753392\\
18	0.696774320808184\\
19	0.704325523883564\\
20	0.708269572236618\\
21	0.711962134365868\\
22	0.718052275196461\\
23	0.72135370655366\\
24	0.724470552519959\\
25	0.727234603905001\\
26	0.732088267372817\\
27	0.734644828756761\\
28	0.736941389324783\\
29	0.739303223567442\\
30	0.74135821076993\\
31	0.743404321699346\\
32	0.7452077027389\\
33	0.747112977301011\\
34	0.74888404851009\\
35	0.750456492574359\\
36	0.752133655657052\\
37	0.754655457041526\\
38	0.756209086840265\\
39	0.757552816413207\\
40	0.758828327867903\\
41	0.76001108994983\\
42	0.761151668419877\\
43	0.762301994344298\\
44	0.763392471944071\\
45	0.765074871981523\\
46	0.766134826565551\\
47	0.767677304090818\\
48	0.768611418359715\\
49	0.769468036572479\\
50	0.769876929907705\\
51	0.770573541756519\\
52	0.771304513929085\\
53	0.771956262494816\\
54	0.772632468090858\\
55	0.773272401739926\\
56	0.773871538878613\\
57	0.77476278609374\\
58	0.775419882540512\\
59	0.775966824777319\\
60	0.776468847395021\\
61	0.776967807133299\\
62	0.777649759333605\\
63	0.778194848412849\\
64	0.778624111851846\\
65	0.779026473712218\\
66	0.779436570191718\\
67	0.779765388903663\\
68	0.780129873455748\\
69	0.780557987588389\\
70	0.780981699727462\\
71	0.781327469744847\\
72	0.781614546914023\\
73	0.781910100089815\\
74	0.782173275577268\\
75	0.782444261731808\\
76	0.782685490933362\\
77	0.782929901226592\\
78	0.783179209478286\\
79	0.783377273282393\\
80	0.783605145050571\\
81	0.783792765119582\\
82	0.784000601754092\\
83	0.784199642201264\\
84	0.784384540520688\\
85	0.78455764154012\\
86	0.784741195890478\\
87	0.784900988898672\\
88	0.785159576360878\\
89	0.78533876612162\\
90	0.785469212563169\\
91	0.785592969408339\\
92	0.785703119729604\\
93	0.785856984257836\\
94	0.785994135152833\\
95	0.786176464146429\\
96	0.786278820130634\\
97	0.786463324314142\\
98	0.786585822628678\\
99	0.786677205111955\\
100	0.786767121005458\\
101	0.786884808868718\\
102	0.78698537709192\\
103	0.787049091457941\\
104	0.787150467038581\\
105	0.787219586104227\\
106	0.787303458102509\\
107	0.787375143003232\\
108	0.787443415112752\\
109	0.787524759663018\\
110	0.7875752595078\\
111	0.787654569843555\\
112	0.787687122712758\\
113	0.787717234676872\\
114	0.787805046960477\\
115	0.787862511172787\\
116	0.787903270216078\\
117	0.787980481424038\\
118	0.788015843940647\\
119	0.788083581264143\\
120	0.788105719202217\\
121	0.788150991026381\\
122	0.788185715341571\\
123	0.788255856782155\\
124	0.788296635202204\\
125	0.788331299708152\\
126	0.788387763064169\\
127	0.78845117271164\\
128	0.788444565579596\\
129	0.788490884461201\\
130	0.788495709845813\\
131	0.788521483848759\\
132	0.788488086605048\\
133	0.788614357764927\\
134	0.788568275379877\\
135	0.788692787059034\\
136	0.788666031530362\\
137	0.788718199236851\\
138	0.788747874161902\\
139	0.788776805240454\\
140	0.788823310362659\\
141	0.788824793573242\\
142	0.788847631793526\\
143	0.788848641714678\\
144	0.788861964436487\\
145	0.788903404670944\\
146	0.788900383443973\\
147	0.788938574826136\\
148	0.788963634301138\\
149	0.7888335638243\\
150	0.788978317182411\\
151	0.788919529305218\\
152	0.789018694988782\\
153	0.789080767472484\\
154	0.789007145912006\\
155	0.789054164925974\\
156	0.789069533400926\\
157	0.789094326403101\\
158	0.789073013334998\\
159	0.789128839596008\\
160	0.789117424970591\\
161	0.789173627635065\\
162	0.789118695952156\\
163	0.789162553109018\\
164	0.789155183663699\\
165	0.789173745467813\\
166	0.789170104876837\\
167	0.789187397888259\\
168	0.789186280138441\\
169	0.789184318768826\\
170	0.789206492252694\\
171	0.789217097821913\\
172	0.789144994453594\\
173	0.789235695217945\\
174	0.789163237455171\\
175	0.789244500773303\\
176	0.789263185480671\\
177	0.789257692057156\\
178	0.789204494275235\\
179	0.789289408419942\\
180	0.789298393369158\\
181	0.789193481121779\\
182	0.789288066070494\\
183	0.789308674030464\\
184	0.789314980407758\\
185	0.789215182276708\\
186	0.789255652747423\\
187	0.789361909057981\\
188	0.789342448115805\\
189	0.789368180289591\\
190	0.789347929577691\\
191	0.789383596136235\\
192	0.789386814106993\\
193	0.789402875100571\\
194	0.78939305407657\\
195	0.789268719502933\\
196	0.789314833068987\\
197	0.78931615129829\\
198	0.789324393441511\\
199	0.789414716542289\\
200	0.789347549877256\\
201	0.789424689980891\\
202	0.789352803586509\\
203	0.789450473866629\\
204	0.789426131557203\\
205	0.789442428785204\\
206	0.789454420652228\\
207	0.789459110711962\\
208	0.789449714416072\\
209	0.789464230828336\\
210	0.789345726677917\\
211	0.78938103245522\\
212	0.78944931753757\\
213	0.789471235022891\\
214	0.789515141749009\\
215	0.789483495843902\\
216	0.789356415512883\\
217	0.789364208136724\\
218	0.789452381165408\\
219	0.789488028322884\\
220	0.78951248408101\\
221	0.789383671640852\\
222	0.789483405785691\\
223	0.789497344121746\\
224	0.789495056843174\\
225	0.789517883044597\\
226	0.789362888955529\\
227	0.789480023991884\\
228	0.789486374238866\\
229	0.789513508774265\\
230	0.789509265397578\\
231	0.789521587505945\\
232	0.789539960384121\\
233	0.789534073953828\\
234	0.789528579702705\\
235	0.789529727878609\\
236	0.78941827120957\\
237	0.789428014957244\\
238	0.789520093265567\\
239	0.789501197267915\\
240	0.789532752897763\\
241	0.789518575973417\\
242	0.789488493776885\\
243	0.78937944622298\\
244	0.789405154530696\\
245	0.789420873156715\\
246	0.789489895941738\\
247	0.789511266195091\\
248	0.789414978425002\\
249	0.789504095179184\\
250	0.789507232472111\\
251	0.789492064354149\\
252	0.789394148070191\\
253	0.789378326360769\\
254	0.789471968181119\\
255	0.789488140209185\\
256	0.789487358338498\\
257	0.789518670839861\\
258	0.789503045513405\\
259	0.789501524750912\\
260	0.789505024057053\\
261	0.789508392752045\\
262	0.789512783387114\\
263	0.789528103742964\\
264	0.789539499945093\\
265	0.789556602928401\\
266	0.789549904618483\\
267	0.789432430585423\\
268	0.789517426094172\\
269	0.789452255868386\\
270	0.789466303210976\\
271	0.7895333905956\\
272	0.789534813747563\\
273	0.789551232405311\\
274	0.789587212442886\\
275	0.789474076235885\\
276	0.789558463535789\\
277	0.789552234986783\\
278	0.78955335472018\\
279	0.78943057345418\\
280	0.789516400394565\\
281	0.789523982111991\\
282	0.789440143122875\\
283	0.789536923419422\\
284	0.789553580874618\\
285	0.78956574926468\\
286	0.789563798826771\\
287	0.789562348911359\\
288	0.789544526661273\\
289	0.789551221335554\\
290	0.789532913414086\\
291	0.789464691087552\\
292	0.789445893851012\\
293	0.789536057027616\\
294	0.789540751935401\\
295	0.789548454615793\\
296	0.789532997924199\\
297	0.789437794190793\\
298	0.789428202426282\\
299	0.789426011503207\\
300	0.789417592386485\\
};
\addplot [color=red, line width=0.8pt, forget plot]
  table[row sep=crcr]{%
1	0.447132792013231\\
2	0.470467194213733\\
3	0.490248031609507\\
4	0.535027981419754\\
5	0.566336266612457\\
6	0.570217827611425\\
7	0.570217827611425\\
8	0.579449268183018\\
9	0.587044608658933\\
10	0.587044608658933\\
11	0.610205495867125\\
12	0.621871728540799\\
13	0.630029587856091\\
14	0.652470300538481\\
15	0.67119704611202\\
16	0.677109027309769\\
17	0.679412838884696\\
18	0.679412838884696\\
19	0.679412838884696\\
20	0.679412838884696\\
21	0.695190440975232\\
22	0.707886748925659\\
23	0.709532172568961\\
24	0.718822792992596\\
25	0.724244099915571\\
26	0.724244099915571\\
27	0.733530213638752\\
28	0.734254116315106\\
29	0.735295343292766\\
30	0.735295343292766\\
31	0.744499819483657\\
32	0.750068173308777\\
33	0.750606704841783\\
34	0.750606704841783\\
35	0.753270707093871\\
36	0.755045220689387\\
37	0.758318634642998\\
38	0.761078297524092\\
39	0.761890147205961\\
40	0.762336065569206\\
41	0.767979039843398\\
42	0.770797538547217\\
43	0.771061883050198\\
44	0.771061883050198\\
45	0.771813175976829\\
46	0.772618267266076\\
47	0.773047311737532\\
48	0.773548545062137\\
49	0.773548545062137\\
50	0.773548545062137\\
51	0.778745053914404\\
52	0.779961957373213\\
53	0.780155912620025\\
54	0.780155912620025\\
55	0.779119491245776\\
56	0.779893562981844\\
57	0.781828011917944\\
58	0.782503084530719\\
59	0.782667379285663\\
60	0.782667379285663\\
61	0.786209982608074\\
62	0.787084616150734\\
63	0.787124805524764\\
64	0.78714930305371\\
65	0.78714930305371\\
66	0.78714930305371\\
67	0.78714930305371\\
68	0.787404545661127\\
69	0.787536038366196\\
70	0.787536038366196\\
71	0.789122534331431\\
72	0.78996323285451\\
73	0.790074572622523\\
74	0.790074572622523\\
75	0.788302156356146\\
76	0.78858553949244\\
77	0.788709339657087\\
78	0.788709339657087\\
79	0.788709339657087\\
80	0.788709339657087\\
81	0.791477689874253\\
82	0.792054517362402\\
83	0.792091767837102\\
84	0.792091767837102\\
85	0.791207576068801\\
86	0.791199097359645\\
87	0.791265760902294\\
88	0.79130962052825\\
89	0.79130962052825\\
90	0.791753698345446\\
91	0.793702654481453\\
92	0.794056475481657\\
93	0.794052198589591\\
94	0.794052198589591\\
95	0.791895754415146\\
96	0.79183289067391\\
97	0.791908662793226\\
98	0.79184748518783\\
99	0.79184748518783\\
100	0.79184748518783\\
101	0.794353318671433\\
102	0.7942679172279\\
103	0.794054708362647\\
104	0.794004713772065\\
105	0.793997835715867\\
106	0.793997835715867\\
107	0.794039122276083\\
108	0.794039122276083\\
109	0.794039122276083\\
110	0.794120532740398\\
111	0.794984920623837\\
112	0.794935201030131\\
113	0.794894645159267\\
114	0.794894645159267\\
115	0.793258798278337\\
116	0.793187611865869\\
117	0.793249291279803\\
118	0.793194766805497\\
119	0.793144465491107\\
120	0.793143417506321\\
121	0.794836094162868\\
122	0.794779440272492\\
123	0.794589480181835\\
124	0.794529284792092\\
125	0.794496246304627\\
126	0.794496246304627\\
127	0.794496246304627\\
128	0.794496246304627\\
129	0.794496246304627\\
130	0.794496246304627\\
131	0.79525551178117\\
132	0.795310231142414\\
133	0.795284847799119\\
134	0.795284847799119\\
135	0.794112924887263\\
136	0.794003456890253\\
137	0.794003456890253\\
138	0.793995203053402\\
139	0.793942677599529\\
140	0.793891182406847\\
141	0.79499948310758\\
142	0.795101243023207\\
143	0.795067891761545\\
144	0.795067891761545\\
145	0.793160931796938\\
146	0.792832229407174\\
147	0.792481648933633\\
148	0.79215360998837\\
149	0.792133376255702\\
150	0.792133376255702\\
151	0.793933637253725\\
152	0.793770002770642\\
153	0.793429509347751\\
154	0.793383450689017\\
155	0.793372887741453\\
156	0.793372887741453\\
157	0.793372887741453\\
158	0.793429134417145\\
159	0.793429134417145\\
160	0.793486231272121\\
161	0.794324349254487\\
162	0.794387345237614\\
163	0.794354569841142\\
164	0.792672579891897\\
165	0.792357806213331\\
166	0.792073559012295\\
167	0.791884968039795\\
168	0.791875968585127\\
169	0.791875968585127\\
170	0.791875968585127\\
171	0.79351650702484\\
172	0.793349768359026\\
173	0.793113533086732\\
174	0.793054990364656\\
175	0.79302956621935\\
176	0.79302956621935\\
177	0.79302956621935\\
178	0.793049652020435\\
179	0.793047932783218\\
180	0.793047932783218\\
181	0.793780491644019\\
182	0.793866778001666\\
183	0.793846134940115\\
184	0.793846134940115\\
185	0.792250247603716\\
186	0.792011333875157\\
187	0.792039201825789\\
188	0.792039201825789\\
189	0.792039201825789\\
190	0.792089491095986\\
191	0.793056347649183\\
192	0.792909012093651\\
193	0.792678790273154\\
194	0.792616165331443\\
195	0.792616165331443\\
196	0.792595465660655\\
197	0.792577273252588\\
198	0.792557956743862\\
199	0.792542141615398\\
200	0.792542141615398\\
201	0.793198239437432\\
202	0.792975048791918\\
203	0.792535196568275\\
204	0.79249356365333\\
205	0.79249356365333\\
206	0.79249356365333\\
207	0.792556937680042\\
208	0.792563355791159\\
209	0.792542112612998\\
210	0.79252023354776\\
211	0.792960926919698\\
212	0.793019543063134\\
213	0.792993239645123\\
214	0.791717885109042\\
215	0.791325378951222\\
216	0.790937699039247\\
217	0.790801497560456\\
218	0.790791139706209\\
219	0.790792026155445\\
220	0.790804817011009\\
221	0.792092772716621\\
222	0.791968926368928\\
223	0.791802149550463\\
224	0.791757237817072\\
225	0.791733470491548\\
226	0.791733470491548\\
227	0.791733470491548\\
228	0.790657568985387\\
229	0.790597224724551\\
230	0.790547329883946\\
231	0.791777576424568\\
232	0.791536765571781\\
233	0.791159550389917\\
234	0.791124239009408\\
235	0.791124239009408\\
236	0.790956846291393\\
237	0.790871471183391\\
238	0.790747369440493\\
239	0.790731160320907\\
240	0.7907161348744\\
241	0.791655525916124\\
242	0.79135221689913\\
243	0.791309093118666\\
244	0.791309093118666\\
245	0.79127693699061\\
246	0.791252397026468\\
247	0.791252397026468\\
248	0.791252397026468\\
249	0.791238626416618\\
250	0.791220012828491\\
251	0.791730011392013\\
252	0.791742829174857\\
253	0.791706854135392\\
254	0.791706854135392\\
255	0.791706854135392\\
256	0.791706854135392\\
257	0.789947370850091\\
258	0.789868411383993\\
259	0.789893065039324\\
260	0.789701901844259\\
261	0.79106005492991\\
262	0.79107553472209\\
263	0.791059356303777\\
264	0.791059356303777\\
265	0.791059356303777\\
266	0.791059356303777\\
267	0.791071165861906\\
268	0.791060523667248\\
269	0.791060523667248\\
270	0.791049568501909\\
271	0.791049182101187\\
272	0.791017535442623\\
273	0.790988540431967\\
274	0.790004661856935\\
275	0.789856118114032\\
276	0.789769060990227\\
277	0.789675669295487\\
278	0.7896530947487\\
279	0.789651418404465\\
280	0.789651418404465\\
281	0.790773107832284\\
282	0.790713566200282\\
283	0.790689446731163\\
284	0.790689446731163\\
285	0.790689446731163\\
286	0.790689446731163\\
287	0.790689446731163\\
288	0.789311856035371\\
289	0.789384114971487\\
290	0.789499407393605\\
291	0.790881039388471\\
292	0.790858405546271\\
293	0.790713470501543\\
294	0.790713470501543\\
295	0.790580567848491\\
296	0.790539987859783\\
297	0.790526877732289\\
298	0.790526877732289\\
299	0.790526877732289\\
300	0.790541475997178\\
};
\end{axis}
\end{tikzpicture}%

%% file: images/conv_energy_iter.tikz
%
%
\begin{tikzpicture}

\definecolor{gray4}{rgb}{0.2431,0.2431,0.2431}
\definecolor{gray3}{rgb}{0.3098,0.3098,0.3098}
\definecolor{gray2}{rgb}{0.4863,0.4863,0.4863}
\definecolor{gray1}{rgb}{0.7725,0.7725,0.7725}

\begin{axis}[%
width=0.951\figurewidth,
height=\figureheight,
at={(0\figurewidth,0\figureheight)},
scale only axis,
unbounded coords=jump,
xmin=0,
xmax=300,
ymin=45000,
ymax=180000,
ylabel style={font=\color{white!15!black}},
ylabel={Energy},
xlabel style={font=\color{white!15!black}},
xlabel={Number of inner and outer iterations},
axis background/.style={fill=white},
xmajorgrids,
ymajorgrids,
xlabel near ticks,
ylabel near ticks
]
\addplot [color=black, line width=0.75pt, forget plot] 
  table[row sep=crcr]{%
1	141019.684277875\\
2	124941.799300844\\
3	174260.297429929\\
4	144881.022319945\\
5	120567.734671966\\
6	117563.566544812\\
7	117563.566544812\\
8	109985.873030147\\
9	104210.257945529\\
10	104210.257945529\\
11	92643.1761520379\\
12	85451.9621582979\\
13	109537.350779375\\
14	95724.387477513\\
15	86143.1134124373\\
16	82603.4967874539\\
17	80939.9933750634\\
18	80939.9933750634\\
19	80939.9933750634\\
20	80939.9933750634\\
21	77320.6894016022\\
22	72282.3160496865\\
23	71086.5550895666\\
24	74028.3755227238\\
25	72893.0317303515\\
26	72893.0317303515\\
27	68647.8810727926\\
28	68202.1558971088\\
29	67166.3090389325\\
30	67166.3090389325\\
31	67490.8901916915\\
32	63504.9934460199\\
33	63155.5598657763\\
34	63155.5598657763\\
35	69737.2301226651\\
36	68805.2555089617\\
37	64996.4015687943\\
38	63255.5406387053\\
39	62928.8149201276\\
40	62499.2220230913\\
41	60750.8734649002\\
42	60582.2104944488\\
43	60236.4976943626\\
44	60236.4976943626\\
45	63833.2877518609\\
46	63459.4198587845\\
47	62726.3851536145\\
48	61806.4945609029\\
49	61806.4945609029\\
50	61806.4945609029\\
51	60323.2088028737\\
52	57962.8895277926\\
53	57573.2171593604\\
54	57573.2171593604\\
55	63164.2768018648\\
56	62369.6568835762\\
57	60351.0163801026\\
58	59636.5546452318\\
59	59191.7174360312\\
60	59191.7174360312\\
61	57555.4493320977\\
62	57577.2327221739\\
63	57724.5366550051\\
64	57656.834007843\\
65	57656.834007843\\
66	57656.834007843\\
67	57656.834007843\\
68	57178.5848783206\\
69	56765.6397617813\\
70	56765.6397617813\\
71	57410.4264068608\\
72	55650.0236232847\\
73	55389.551651364\\
74	55389.551651364\\
75	60794.1917735701\\
76	60012.6286157874\\
77	59696.1024954574\\
78	59696.1024954574\\
79	59696.1024954574\\
80	59696.1024954574\\
81	57632.8349839238\\
82	56715.698799646\\
83	56396.3853016243\\
84	56396.3853016243\\
85	58569.8722699263\\
86	58066.9839996798\\
87	57542.965664439\\
88	57446.5737936766\\
89	57446.5737936766\\
90	56372.1457312259\\
91	55678.6263832295\\
92	54811.9532375195\\
93	54717.1220923352\\
94	54717.1220923352\\
95	59718.4669347562\\
96	59342.7113903647\\
97	58133.3099455638\\
98	57686.1268265501\\
99	57686.1268265501\\
100	57686.1268265501\\
101	55201.5218172912\\
102	56233.7615726548\\
103	56149.9515889741\\
104	55880.2813237698\\
105	55701.3126970172\\
106	55701.3126970172\\
107	55376.5704848165\\
108	55376.5704848165\\
109	55376.5704848165\\
110	54996.2174031223\\
111	55964.5758881427\\
112	55079.8278632438\\
113	54970.5271733284\\
114	54970.5271733284\\
115	57885.5635931163\\
116	57576.6671102643\\
117	56268.4473673612\\
118	55810.2812112762\\
119	55615.5234676934\\
120	55433.5696753841\\
121	54455.5035591517\\
122	54871.2519823951\\
123	54957.0259903983\\
124	54821.634461455\\
125	54674.4537049277\\
126	54674.4537049277\\
127	54674.4537049277\\
128	54674.4537049277\\
129	54674.4537049277\\
130	54674.4537049277\\
131	54726.5449228794\\
132	54232.1959080679\\
133	54058.8972058226\\
134	54058.8972058226\\
135	56410.807571065\\
136	56143.6946110623\\
137	56143.6946110623\\
138	55493.195367413\\
139	55281.307075006\\
140	54499.824979657\\
141	53868.1161506523\\
142	53265.2281607977\\
143	53204.1872843958\\
144	53204.1872843958\\
145	57167.4416428726\\
146	57012.1843537211\\
147	56238.4099307659\\
148	55994.3036302175\\
149	55914.4855638154\\
150	55914.4855638154\\
151	53926.3279878257\\
152	54899.8059724459\\
153	55089.1263013142\\
154	54918.0900358369\\
155	54749.1078954419\\
156	54749.1078954419\\
157	54749.1078954419\\
158	54365.9592090368\\
159	54365.9592090368\\
160	54044.9037306778\\
161	53564.3739350654\\
162	52900.7033656893\\
163	52801.5673989829\\
164	56420.9224285035\\
165	56356.6905858622\\
166	55487.6609978069\\
167	55231.5317529241\\
168	55152.1430306003\\
169	55152.1430306003\\
170	55152.1430306003\\
171	53618.6897629429\\
172	54360.7540936683\\
173	54381.1305765637\\
174	54255.5229493854\\
175	54093.7498611946\\
176	54093.7498611946\\
177	54093.7498611946\\
178	53745.0184876044\\
179	53646.0235855544\\
180	53646.0235855544\\
181	53316.5683124659\\
182	52634.7379756065\\
183	52532.647591292\\
184	52532.647591292\\
185	56163.9373120433\\
186	55930.2429844331\\
187	55687.9043900955\\
188	55687.9043900955\\
189	55687.9043900955\\
190	55662.416480797\\
191	53473.8821390887\\
192	53708.8302757587\\
193	53782.4897071563\\
194	53750.1991734183\\
195	53750.1991734183\\
196	53483.0218961957\\
197	53373.441260397\\
198	53283.2571392579\\
199	53257.4802335875\\
200	53257.4802335875\\
201	52390.8464609587\\
202	54249.19646582\\
203	54181.9500631366\\
204	54119.499810987\\
205	54119.499810987\\
206	54119.499810987\\
207	53503.4857803798\\
208	53297.3606659269\\
209	53109.4878779927\\
210	53061.4453848386\\
211	52831.2674291749\\
212	52209.4251281948\\
213	52130.656456337\\
214	55197.0835929464\\
215	55279.5058812551\\
216	54834.2888598477\\
217	54630.3974069922\\
218	54552.6867224954\\
219	54385.6919558718\\
220	54225.6844117408\\
221	52904.6391470496\\
222	53550.1370676236\\
223	53552.782654249\\
224	53432.9785034904\\
225	53287.5128422681\\
226	53287.5128422681\\
227	53287.5128422681\\
228	55070.0533086884\\
229	54870.3400219598\\
230	54166.0087528171\\
231	52776.7707623663\\
232	53690.1651595624\\
233	53985.8648359997\\
234	53934.5989037451\\
235	53934.5989037451\\
236	53850.8710680372\\
237	53954.3995128297\\
238	53609.457292001\\
239	53541.1680570015\\
240	53477.3698425433\\
241	53043.9918371667\\
242	53380.9273579997\\
243	53341.1150813552\\
244	53341.1150813552\\
245	53167.6422696674\\
246	53069.236561982\\
247	53069.236561982\\
248	53069.236561982\\
249	52943.3960438118\\
250	52924.6242679937\\
251	52788.4907714855\\
252	52311.5979473954\\
253	52254.114159788\\
254	52254.114159788\\
255	52254.114159788\\
256	52254.114159788\\
257	56378.9409621255\\
258	55412.4480059696\\
259	55244.7663653944\\
260	54775.3145607776\\
261	53346.5001232792\\
262	52983.0348120372\\
263	52913.7188185789\\
264	52913.7188185789\\
265	52913.7188185789\\
266	52913.7188185789\\
267	52631.9710994076\\
268	52557.0168298366\\
269	52557.0168298366\\
270	52519.3616908674\\
271	54123.4130212696\\
272	53220.723746378\\
273	53116.5184711061\\
274	54893.2060546391\\
275	54856.4895673842\\
276	54304.0397884463\\
277	54049.5158042198\\
278	53999.3998323247\\
279	53908.1965592452\\
280	53908.1965592452\\
281	53349.2527866675\\
282	52961.8010925205\\
283	52905.359709421\\
284	52905.359709421\\
285	52905.359709421\\
286	52905.359709421\\
287	52905.359709421\\
288	56336.4767028821\\
289	55767.2172753991\\
290	54532.2180798329\\
291	52653.6139092387\\
292	53226.0890412959\\
293	53232.9376047397\\
294	53232.9376047397\\
295	53010.9699656582\\
296	53057.7843328264\\
297	52943.1941697572\\
298	52943.1941697572\\
299	52943.1941697572\\
300	52782.7017805521\\
};
\addplot [color=blue, line width=0.75pt, forget plot] 
  table[row sep=crcr]{%
1	141019.684277875\\
2	115799.255197858\\
3	109675.704288699\\
4	104646.777959725\\
5	98834.435468418\\
6	93050.1679020519\\
7	87716.1953634486\\
8	83097.8502062313\\
9	78942.9101789921\\
10	75580.3747530717\\
11	73137.7882405489\\
12	70259.4893894024\\
13	68015.9297510687\\
14	66369.373953799\\
15	64587.8693341998\\
16	63301.5404294258\\
17	61870.728815362\\
18	61047.9740875648\\
19	59897.9815695287\\
20	59235.2471316978\\
21	58669.6739125732\\
22	57948.3080944349\\
23	57478.2824957427\\
24	56769.0774446962\\
25	56887.4628841777\\
26	56231.1449951807\\
27	55830.1037624016\\
28	55475.0694610284\\
29	55302.5485328856\\
30	55053.9761793369\\
31	54746.530280286\\
32	54585.842717083\\
33	54306.0473730621\\
34	54487.5484529533\\
35	54060.1406107504\\
36	54039.210379854\\
37	53712.3172498266\\
38	53725.8287585393\\
39	53469.2030386836\\
40	53435.41292469\\
41	53279.3540066762\\
42	53247.0349403288\\
43	53225.9743002728\\
44	53037.4744744016\\
45	53010.7116112037\\
46	52824.4509448727\\
47	52853.8119884016\\
48	52662.3721216907\\
49	52813.7894442766\\
50	52562.2461590305\\
51	52475.8737785478\\
52	52498.901977635\\
53	52391.7468013452\\
54	52344.0518407083\\
55	52247.0653176746\\
56	52265.9099059196\\
57	52119.5402935335\\
58	52255.7555264641\\
59	52098.2291480491\\
60	52099.2715747392\\
61	52052.2548222835\\
62	51977.4298364103\\
63	51993.6996151506\\
64	51828.4704757274\\
65	51973.4586990346\\
66	51798.6345184689\\
67	51797.0857645945\\
68	51706.2540434943\\
69	51682.5102153033\\
70	51614.7141102556\\
71	51800.6220280812\\
72	51675.8451520431\\
73	51563.4613363179\\
74	51561.8366525895\\
75	51513.6843754164\\
76	51455.1385611596\\
77	51342.1778694188\\
78	51589.1800172023\\
79	51375.7618818578\\
80	51432.3923910364\\
81	51331.5264956424\\
82	51301.3034235183\\
83	51242.1527236207\\
84	51262.3653005706\\
85	51177.0943282198\\
86	51318.1535392922\\
87	51258.7240858675\\
88	51109.2454204711\\
89	51235.3306717878\\
90	51241.7246074454\\
91	51165.1305589569\\
92	51041.0363617273\\
93	51085.6005406846\\
94	51064.1892796981\\
95	50999.3369739144\\
96	51028.004807957\\
97	50922.7587804308\\
98	51023.340102776\\
99	51038.9712960171\\
100	50956.7281282542\\
101	50906.6173348516\\
102	50952.8645205549\\
103	50818.0020657667\\
104	50899.3185507976\\
105	50906.5529962286\\
106	50773.9421097959\\
107	50852.4636136838\\
108	50833.1044226433\\
109	50739.5465738451\\
110	50752.594155135\\
111	50724.1134878136\\
112	50724.9373884142\\
113	50792.9969034761\\
114	50756.176136975\\
115	50688.1715565975\\
116	50634.7685121348\\
117	50700.0130984295\\
118	50694.0100586612\\
119	50642.9553205253\\
120	50600.5392071945\\
121	50607.9189808093\\
122	50605.9254436209\\
123	50580.126786128\\
124	50590.3704584167\\
125	50489.3065906672\\
126	50720.3794449242\\
127	50587.6911705863\\
128	50588.2144297768\\
129	50457.3882800395\\
130	50562.2896306024\\
131	50587.1020186584\\
132	50536.4433973375\\
133	50481.9451176078\\
134	50488.085604889\\
135	50451.5811742677\\
136	50476.4800626561\\
137	50438.3027347273\\
138	50378.89460018\\
139	50526.1735823242\\
140	50482.1264482948\\
141	50415.7281104629\\
142	50390.3648641777\\
143	50392.1575866886\\
144	50318.4516560498\\
145	50388.2497559733\\
146	50374.5838357445\\
147	50321.4099142831\\
148	50312.5218928198\\
149	50361.399659758\\
150	50268.4809696255\\
151	50353.7570407113\\
152	50204.9804966657\\
153	50419.4190592281\\
154	50357.7755454836\\
155	50260.8432003519\\
156	50201.7180569834\\
157	50399.6394156983\\
158	50294.2117637092\\
159	50213.1116935775\\
160	50263.5728117339\\
161	50200.9605019315\\
162	50301.8332683843\\
163	50283.8086229834\\
164	50205.47647503\\
165	50230.3583807726\\
166	50207.706319604\\
167	50175.2672122618\\
168	50192.0122310964\\
169	50126.2707747508\\
170	50262.7213003945\\
171	50199.40552361\\
172	50204.8775866789\\
173	50139.3948392181\\
174	50181.1006711047\\
175	50202.4011619123\\
176	50102.7044417826\\
177	50161.0980345869\\
178	50117.5255295968\\
179	50133.8267057099\\
180	50058.7774211705\\
181	50234.3454699012\\
182	50151.3334704521\\
183	50080.5140980247\\
184	50122.4825888498\\
185	50054.4389160565\\
186	50107.7922008648\\
187	50042.2957301014\\
188	50085.990829218\\
189	50058.8206662685\\
190	50050.2245876077\\
191	50018.1645104529\\
192	50083.1353762099\\
193	50011.7153272873\\
194	50008.4649910533\\
195	50003.0364225559\\
196	50001.2957651663\\
197	49956.6936197811\\
198	50046.3373528175\\
199	49955.5444144445\\
200	50097.141705157\\
201	50013.1338882922\\
202	50036.0771427841\\
203	49925.4501005933\\
204	50026.6802781193\\
205	49970.1400741866\\
206	49920.1624868202\\
207	49987.4360768723\\
208	49978.5161224142\\
209	49878.6483992792\\
210	50044.7308515343\\
211	49965.8305043886\\
212	49963.4409771646\\
213	49880.4116694612\\
214	50031.3526575293\\
215	49900.3986018588\\
216	49896.1518296682\\
217	49960.8530238404\\
218	49841.7666167766\\
219	50018.7081206551\\
220	49950.3944720298\\
221	49908.9586177242\\
222	49859.7220728383\\
223	49882.340503914\\
224	49812.5001131297\\
225	49954.1799570291\\
226	49992.131175443\\
227	49901.2817923589\\
228	49851.8100069162\\
229	49927.127540404\\
230	49884.7235207881\\
231	49903.079567996\\
232	49849.7106183875\\
233	49819.713593816\\
234	49937.6795245598\\
235	49865.6161762218\\
236	49813.9382057388\\
237	49833.2932963618\\
238	49815.3973613197\\
239	49835.3901575899\\
240	49876.7631546031\\
241	49777.5550472482\\
242	49901.5231020181\\
243	49851.6510790041\\
244	49772.3336493544\\
245	49842.278694323\\
246	49855.688674971\\
247	49768.4190823556\\
248	49854.5188026767\\
249	49817.3399611147\\
250	49713.0277717778\\
251	49701.3482183436\\
252	49869.065789456\\
253	49823.333117682\\
254	49740.3074777797\\
255	49830.2201043253\\
256	49785.6682332512\\
257	49783.7250363392\\
258	49742.6283964084\\
259	49854.1686637612\\
260	49777.3349165118\\
261	49758.0500307977\\
262	49709.8953846654\\
263	49833.07578356\\
264	49816.7438432775\\
265	49765.1941393012\\
266	49760.1916075594\\
267	49765.3772107681\\
268	49765.5149199725\\
269	49675.4850552232\\
270	49833.2496598724\\
271	49781.6412739896\\
272	49766.4981682435\\
273	49686.5124347684\\
274	49761.5425403504\\
275	49696.9954346596\\
276	49779.181772781\\
277	49702.0384855704\\
278	49667.2838718311\\
279	49722.4738242226\\
280	49771.1094505375\\
281	49706.9335780467\\
282	49682.945765341\\
283	49777.0764667567\\
284	49666.9571393913\\
285	49737.748210069\\
286	49694.4880368602\\
287	49697.8842549428\\
288	49753.7185869008\\
289	49765.6489821806\\
290	49688.5428717957\\
291	49712.2889468471\\
292	49658.8704188369\\
293	49693.1961346143\\
294	49665.1073818975\\
295	49656.1821843707\\
296	49762.9291939706\\
297	49710.0357710432\\
298	49684.8045137358\\
299	49723.8619836224\\
300	49640.0894978862\\
};
\addplot [color=red, line width=0.75pt, forget plot] 
  table[row sep=crcr]{%
1	80782.1927641269\\
2	66782.5712423604\\
3	64520.6076206574\\
4	63966.3093803826\\
5	63371.3400522195\\
6	62794.5177284347\\
7	62306.0157890627\\
8	62306.0157890627\\
9	62306.0157890627\\
10	62306.0157890627\\
11	77588.907217733\\
12	66584.2260730554\\
13	62122.9147115157\\
14	60422.4391718117\\
15	59444.4838079697\\
16	58855.6577330259\\
17	58439.3772601898\\
18	58124.4839042029\\
19	57883.5555339535\\
20	57701.2258403201\\
21	76115.1025977459\\
22	65390.4480861321\\
23	61015.800236351\\
24	59097.7148733572\\
25	58038.9677624671\\
26	57373.4485571419\\
27	56914.0984163978\\
28	56573.4280100318\\
29	56322.7640047889\\
30	56140.4561283135\\
31	72802.9686195028\\
32	64185.6726761346\\
33	60437.4308147409\\
34	58567.0238072814\\
35	57533.1944032725\\
36	56876.1613784682\\
37	56420.265977833\\
38	56092.834400117\\
39	55857.9311006208\\
40	55687.7972990678\\
41	69754.1117297195\\
42	62934.0548492422\\
43	59679.0217130438\\
44	57912.2641393576\\
45	56875.9572616791\\
46	56217.3437155886\\
47	55761.8989603681\\
48	55434.0150337356\\
49	55197.010825975\\
50	55020.1083821883\\
51	67263.7107825435\\
52	61755.6307458181\\
53	59034.840692109\\
54	57469.4769311025\\
55	56506.9827494264\\
56	55886.848884884\\
57	55456.7478136009\\
58	55145.4039741114\\
59	54923.3026829884\\
60	54768.2441532613\\
61	65941.2395479822\\
62	61221.1283151159\\
63	58749.1541145921\\
64	57257.0796859512\\
65	56318.6405688759\\
66	55711.1214767684\\
67	55274.7388422922\\
68	54967.6574168498\\
69	54745.080935284\\
70	54587.7772791952\\
71	64614.5612798859\\
72	60409.1625089934\\
73	58123.9423661419\\
74	56701.4422937783\\
75	55824.3436408233\\
76	55235.3237673936\\
77	54816.3613009774\\
78	54528.1629695291\\
79	54323.8462058595\\
80	54185.1862656286\\
81	63703.4179963666\\
82	60144.906276149\\
83	58011.6723124811\\
84	56634.6644719693\\
85	55723.5338808056\\
86	55151.8631526723\\
87	54749.8030048525\\
88	54465.58145334\\
89	54281.4686869003\\
90	54127.3156002252\\
91	63021.3921068068\\
92	59717.6937900302\\
93	57821.820361784\\
94	56530.2173651765\\
95	55694.2994307986\\
96	55142.0148297409\\
97	54746.692984426\\
98	54470.7187151019\\
99	54272.3484384793\\
100	54130.922790957\\
101	62627.7236432967\\
102	59625.9710014122\\
103	57800.7926271971\\
104	56510.7297439357\\
105	55674.9338230596\\
106	55129.0297984509\\
107	54746.4512307941\\
108	54471.5015338039\\
109	54276.3290475034\\
110	54131.0507263748\\
111	62047.3556064079\\
112	59314.0977386317\\
113	57649.5872704991\\
114	56461.488433103\\
115	55665.1977533327\\
116	55142.5471747738\\
117	54762.958055754\\
118	54502.1936409371\\
119	54307.9923447578\\
120	54164.4492043374\\
121	61898.0229168368\\
122	59268.9858183802\\
123	57584.9928365603\\
124	56379.4682940879\\
125	55588.2563863053\\
126	55062.0313739843\\
127	54682.6113131988\\
128	54434.4014690456\\
129	54239.6225527147\\
130	54104.3813879219\\
131	61481.1528365701\\
132	58888.2835089294\\
133	57251.0206544257\\
134	56043.5274709655\\
135	55254.2666465953\\
136	54721.08689091\\
137	54332.4056940501\\
138	54060.1731183373\\
139	53872.8165416961\\
140	53743.5749739521\\
141	61070.8064862037\\
142	58708.4962870882\\
143	57189.9852649164\\
144	56070.4963066011\\
145	55320.4100669634\\
146	54811.8983471819\\
147	54439.6285059903\\
148	54188.5027919358\\
149	54009.8181354292\\
150	53884.8013602073\\
151	60819.4078300716\\
152	58403.5854592306\\
153	56873.9592954025\\
154	55742.2021127615\\
155	54993.8716165819\\
156	54493.724597194\\
157	54130.9011154401\\
158	53879.4114481572\\
159	53707.4835955333\\
160	53590.613688149\\
161	60594.0406318029\\
162	58310.9580738129\\
163	56905.5694649032\\
164	55846.517412769\\
165	55142.7877722367\\
166	54666.7503489896\\
167	54319.2988509329\\
168	54068.9250540412\\
169	53888.5517480704\\
170	53765.581907337\\
171	60554.6257000233\\
172	58386.7753926666\\
173	56967.7724927639\\
174	55902.7826367097\\
175	55181.9253512613\\
176	54693.0896224759\\
177	54342.3864478625\\
178	54096.821230901\\
179	53925.2693697331\\
180	53808.4644253286\\
181	60564.4299952567\\
182	58480.026621581\\
183	57094.0272871242\\
184	56027.346349632\\
185	55296.4975212194\\
186	54806.2966691693\\
187	54447.029262677\\
188	54200.7313409924\\
189	54026.5937862366\\
190	53908.303094218\\
191	60364.6233707025\\
192	58255.2143622676\\
193	56859.4349533125\\
194	55800.5745880283\\
195	55077.7392753082\\
196	54585.1959622667\\
197	54230.64147755\\
198	53986.5659264049\\
199	53818.3072905054\\
200	53693.5687502396\\
201	60156.808139385\\
202	58095.5246208562\\
203	56749.0848267038\\
204	55743.3253997418\\
205	55050.9301577243\\
206	54590.747363731\\
207	54252.8316993934\\
208	54014.0912209012\\
209	53854.216679887\\
210	53731.0884322391\\
211	60197.8209955778\\
212	58145.1976651805\\
213	56812.9381924126\\
214	55791.0472270914\\
215	55083.4246505159\\
216	54607.7170938486\\
217	54249.5419327734\\
218	54000.3494686938\\
219	53827.9771856986\\
220	53701.9306923443\\
221	59974.0596549448\\
222	57980.6572696098\\
223	56696.2282352356\\
224	55693.1388869223\\
225	55018.9018893871\\
226	54551.6830630134\\
227	54203.4609716599\\
228	53954.703353839\\
229	53781.7206240716\\
230	53666.6887427888\\
231	59930.6302941826\\
232	58080.6155741396\\
233	56752.065264598\\
234	55745.921697989\\
235	55069.7675179902\\
236	54603.9760157164\\
237	54264.2456804735\\
238	54025.9586901711\\
239	53847.0491865382\\
240	53729.0873966213\\
241	59950.816460469\\
242	58021.8116822002\\
243	56737.4743773587\\
244	55730.5003748356\\
245	55037.8153872883\\
246	54572.2987757849\\
247	54234.5461894177\\
248	54004.921495697\\
249	53837.4403757855\\
250	53716.5982704268\\
251	60019.9506734553\\
252	58084.2322729807\\
253	56854.4478687231\\
254	55889.2391477037\\
255	55217.0701433608\\
256	54765.6281176364\\
257	54441.2642079157\\
258	54208.7682499875\\
259	54054.6586323996\\
260	53939.709649167\\
261	59877.3675147446\\
262	57942.4265567665\\
263	56651.5968682481\\
264	55662.947055474\\
265	54979.9088770737\\
266	54531.5537675988\\
267	54208.3857510604\\
268	53981.4160776551\\
269	53823.1410590481\\
270	53695.372280802\\
271	59926.846752608\\
272	58034.0039984798\\
273	56794.8931547749\\
274	55833.5661068647\\
275	55167.6191302513\\
276	54718.2405467951\\
277	54384.7391095679\\
278	54143.3063535497\\
279	53975.0986859837\\
280	53862.6387503523\\
281	59789.2021757185\\
282	57985.8033464993\\
283	56751.9380784928\\
284	55790.0848813545\\
285	55112.6295264693\\
286	54655.2182815288\\
287	54315.1816168819\\
288	54070.3087836921\\
289	53911.4312918115\\
290	53791.8658615736\\
291	59880.8605038312\\
292	58004.9818925275\\
293	56717.9785184676\\
294	55705.740122916\\
295	55021.042339679\\
296	54558.822858815\\
297	54225.0415428262\\
298	53999.2031782135\\
299	53816.0967144036\\
300	53693.972234785\\
};
\end{axis}
\end{tikzpicture}%

%% file: images/conv_PSNR_lambda.tikz
%
%
\begin{tikzpicture}

\definecolor{gray4}{rgb}{0.2431,0.2431,0.2431}
\definecolor{gray3}{rgb}{0.3098,0.3098,0.3098}
\definecolor{gray2}{rgb}{0.4863,0.4863,0.4863}
\definecolor{gray1}{rgb}{0.7725,0.7725,0.7725}

\begin{axis}[%
width=0.951\figurewidth,
height=\figureheight,
at={(0\figurewidth,0\figureheight)},
scale only axis,
unbounded coords=jump,
xmin=0,
xmax=300,
ymin=24.5,
ymax=29.5,
axis background/.style={fill=white},
ylabel style={font=\color{white!15!black}},
ylabel={PSNR},
xlabel style={font=\color{white!15!black}},
xlabel={Number of inner and outer iterations},
xmajorgrids,
ymajorgrids,
xlabel near ticks,
ylabel near ticks
]
\addplot [color=black, line width=0.75pt, forget plot]
  table[row sep=crcr]{%
1	24.661462740074\\
2	24.9321012952417\\
3	25.3357945031032\\
4	25.7331697773387\\
5	26.0091794206517\\
6	26.0432311918526\\
7	26.0432311918526\\
8	26.1189459100779\\
9	26.1809963127787\\
10	26.1809963127787\\
11	26.4065956756952\\
12	26.5241257475114\\
13	26.7024329776134\\
14	26.9092666521803\\
15	27.0884279107541\\
16	27.1454110929849\\
17	27.1675096114449\\
18	27.1675096114449\\
19	27.1675096114449\\
20	27.1675096114449\\
21	27.3270721169274\\
22	27.4589899156044\\
23	27.4769349943889\\
24	27.6031497597997\\
25	27.6652577707251\\
26	27.6652577707251\\
27	27.7589179727578\\
28	27.7663442015763\\
29	27.7770750278091\\
30	27.7770750278091\\
31	27.8746789684252\\
32	27.9358185590299\\
33	27.9426473812919\\
34	27.9426473812919\\
35	28.0005846220551\\
36	28.0197615052225\\
37	28.0515890809355\\
38	28.080260491069\\
39	28.0887816091496\\
40	28.0933930682979\\
41	28.1469793257639\\
42	28.1813984622714\\
43	28.1849675224504\\
44	28.1849675224504\\
45	28.2103697269095\\
46	28.2199829959291\\
47	28.2240897080484\\
48	28.2286496912577\\
49	28.2286496912577\\
50	28.2286496912577\\
51	28.2768903698878\\
52	28.2908508085315\\
53	28.2940025240988\\
54	28.2940025240988\\
55	28.3055058655239\\
56	28.3140167726574\\
57	28.3343823614469\\
58	28.3421667156594\\
59	28.3439039858068\\
60	28.3439039858068\\
61	28.3740962477303\\
62	28.3857971557645\\
63	28.3886731670152\\
64	28.3893371966065\\
65	28.3893371966065\\
66	28.3893371966065\\
67	28.3893371966065\\
68	28.3916835758049\\
69	28.392978968135\\
70	28.392978968135\\
71	28.4070372480684\\
72	28.4156441881651\\
73	28.4174119044277\\
74	28.4174119044277\\
75	28.4116161934599\\
76	28.4151623880064\\
77	28.4161221621797\\
78	28.4161221621797\\
79	28.4161221621797\\
80	28.4161221621797\\
81	28.4379217662407\\
82	28.4452381470354\\
83	28.4458643093591\\
84	28.4458643093591\\
85	28.4424296442057\\
86	28.44328352436\\
87	28.4438876981073\\
88	28.4441479535319\\
89	28.4441479535319\\
90	28.4465983099513\\
91	28.4616465326783\\
92	28.4648817965239\\
93	28.465058336535\\
94	28.465058336535\\
95	28.4496171786245\\
96	28.4487374554013\\
97	28.4473021061014\\
98	28.4448402580824\\
99	28.4448402580824\\
100	28.4448402580824\\
101	28.4637963073598\\
102	28.4628673806232\\
103	28.4609151648362\\
104	28.4603281790402\\
105	28.4601324060588\\
106	28.4601324060588\\
107	28.4601798794023\\
108	28.4601798794023\\
109	28.4601798794023\\
110	28.4603147543705\\
111	28.4658634083988\\
112	28.4652415972414\\
113	28.4649026769086\\
114	28.4649026769086\\
115	28.451189332241\\
116	28.449642456852\\
117	28.4486023226322\\
118	28.4462890704584\\
119	28.4445748910758\\
120	28.4442560005281\\
121	28.4572539428556\\
122	28.4566928575593\\
123	28.4550193376227\\
124	28.4544260023609\\
125	28.4540256209529\\
126	28.4540256209529\\
127	28.4540256209529\\
128	28.4540256209529\\
129	28.4540256209529\\
130	28.4540256209529\\
131	28.4590421570664\\
132	28.4587219629188\\
133	28.4584836470209\\
134	28.4584836470209\\
135	28.4479102595015\\
136	28.446180167657\\
137	28.446180167657\\
138	28.444214421114\\
139	28.4427573204529\\
140	28.4408827082938\\
141	28.4491892735283\\
142	28.4496640577661\\
143	28.4493640506329\\
144	28.4493640506329\\
145	28.4328165925794\\
146	28.4285245515492\\
147	28.4224543536884\\
148	28.4172757305859\\
149	28.4168529896008\\
150	28.4168529896008\\
151	28.4316008604439\\
152	28.4305997089892\\
153	28.4273961847431\\
154	28.4268378780291\\
155	28.4265832301787\\
156	28.4265832301787\\
157	28.4265832301787\\
158	28.4266187352579\\
159	28.4266187352579\\
160	28.42658497176\\
161	28.4327153715379\\
162	28.4332035168114\\
163	28.4329482770475\\
164	28.418893922508\\
165	28.4155424148013\\
166	28.4111677138874\\
167	28.4081154408482\\
168	28.4078342825801\\
169	28.4078342825801\\
170	28.4078342825801\\
171	28.4193860055315\\
172	28.4174191303163\\
173	28.4152044951845\\
174	28.4146030248282\\
175	28.4142782663104\\
176	28.4142782663104\\
177	28.4142782663104\\
178	28.4141948600786\\
179	28.4140257679285\\
180	28.4140257679285\\
181	28.4193953109161\\
182	28.4198570118008\\
183	28.4196746145212\\
184	28.4196746145212\\
185	28.4065455004148\\
186	28.4039941783403\\
187	28.4041014992216\\
188	28.4041014992216\\
189	28.4041014992216\\
190	28.4016476439046\\
191	28.4086445230235\\
192	28.4067106636054\\
193	28.4045453085952\\
194	28.4039666510797\\
195	28.4039666510797\\
196	28.4036510505211\\
197	28.403421785149\\
198	28.4030810820104\\
199	28.4027856649085\\
200	28.4027856649085\\
201	28.4078499006352\\
202	28.4060225423803\\
203	28.4021485834058\\
204	28.4017976881259\\
205	28.4017976881259\\
206	28.4017976881259\\
207	28.4020145750381\\
208	28.401903834992\\
209	28.4015185858676\\
210	28.4012274000956\\
211	28.4039448865431\\
212	28.4042470142986\\
213	28.4040111974711\\
214	28.3929182627293\\
215	28.3888202714229\\
216	28.3839995323089\\
217	28.3821014203336\\
218	28.3818677023448\\
219	28.381616608838\\
220	28.3814085167664\\
221	28.3912355195834\\
222	28.3903615261011\\
223	28.3888769246313\\
224	28.3884134459618\\
225	28.3881113619976\\
226	28.3881113619976\\
227	28.3881113619976\\
228	28.3766219596547\\
229	28.3754897713916\\
230	28.3741972790341\\
231	28.3846168908636\\
232	28.382801334828\\
233	28.3794866176603\\
234	28.3791148926632\\
235	28.3791148926632\\
236	28.3767221969342\\
237	28.3755417896334\\
238	28.3738794295375\\
239	28.3736412730622\\
240	28.3733750446014\\
241	28.3807254105159\\
242	28.3781004402127\\
243	28.3777181479369\\
244	28.3777181479369\\
245	28.3773902992663\\
246	28.3771144957122\\
247	28.3771144957122\\
248	28.3771144957122\\
249	28.3768244340429\\
250	28.3765452226514\\
251	28.3804320856303\\
252	28.3805348165759\\
253	28.3802883431126\\
254	28.3802883431126\\
255	28.3802883431126\\
256	28.3802883431126\\
257	28.3655045955205\\
258	28.3634193704615\\
259	28.3635119210757\\
260	28.3603923022777\\
261	28.3718206247713\\
262	28.3720961892261\\
263	28.3719783510115\\
264	28.3719783510115\\
265	28.3719783510115\\
266	28.3719783510115\\
267	28.3721504382805\\
268	28.3721029638899\\
269	28.3721029638899\\
270	28.3720000917789\\
271	28.371359657877\\
272	28.3708149187223\\
273	28.3705884593052\\
274	28.3620822107444\\
275	28.3606316537249\\
276	28.3594533833691\\
277	28.358119182289\\
278	28.3577190490896\\
279	28.3575836204223\\
280	28.3575836204223\\
281	28.3665490613888\\
282	28.3664323596016\\
283	28.3662755293831\\
284	28.3662755293831\\
285	28.3662755293831\\
286	28.3662755293831\\
287	28.3662755293831\\
288	28.3554897643825\\
289	28.3557836637057\\
290	28.3560772641361\\
291	28.3673976335766\\
292	28.3669945005484\\
293	28.3658070173121\\
294	28.3658070173121\\
295	28.3646630739636\\
296	28.3642772415251\\
297	28.3640939027895\\
298	28.3640939027895\\
299	28.3640939027895\\
300	28.3640487529391\\
};
\addplot [color=blue, line width=0.75pt, forget plot]
  table[row sep=crcr]{%
1	24.661462740074\\
2	24.661462740074\\
3	24.661462740074\\
4	24.661462740074\\
5	24.661462740074\\
6	24.661462740074\\
7	24.661462740074\\
8	24.661462740074\\
9	24.661462740074\\
10	24.661462740074\\
11	24.7703401100779\\
12	24.7703401100779\\
13	24.7703401100779\\
14	24.7703401100779\\
15	24.7703401100779\\
16	24.7703401100779\\
17	24.7703401100779\\
18	24.7703401100779\\
19	24.7703401100779\\
20	24.7703401100779\\
21	24.8325795577627\\
22	24.8325795577627\\
23	24.8325795577627\\
24	24.8325795577627\\
25	24.8325795577627\\
26	24.8325795577627\\
27	24.8325795577627\\
28	24.8325795577627\\
29	24.8325795577627\\
30	24.8325795577627\\
31	24.8325795577627\\
32	24.8325795577627\\
33	24.8325795577627\\
34	24.8325795577627\\
35	24.8325795577627\\
36	24.8325795577627\\
37	24.8325795577627\\
38	24.8325795577627\\
39	24.8325795577627\\
40	24.8325795577627\\
41	24.8325795577627\\
42	24.8325795577627\\
43	24.8325795577627\\
44	24.8325795577627\\
45	24.8325795577627\\
46	24.8325795577627\\
47	24.8325795577627\\
48	24.8325795577627\\
49	24.8325795577627\\
50	24.8325795577627\\
51	24.8325795577627\\
52	24.8325795577627\\
53	24.8325795577627\\
54	24.8325795577627\\
55	24.8325795577627\\
56	24.8325795577627\\
57	24.8325795577627\\
58	24.8325795577627\\
59	24.8325795577627\\
60	24.8325795577627\\
61	24.8325795577627\\
62	24.8325795577627\\
63	24.8325795577627\\
64	24.8325795577627\\
65	24.8325795577627\\
66	24.8325795577627\\
67	24.8325795577627\\
68	24.8325795577627\\
69	24.8325795577627\\
70	24.8325795577627\\
71	24.8325795577627\\
72	24.8325795577627\\
73	24.8325795577627\\
74	24.8325795577627\\
75	24.8325795577627\\
76	24.8325795577627\\
77	24.8325795577627\\
78	24.8325795577627\\
79	24.8325795577627\\
80	24.8325795577627\\
81	24.8325795577627\\
82	24.8325795577627\\
83	24.8325795577627\\
84	24.8325795577627\\
85	24.8325795577627\\
86	24.8325795577627\\
87	24.8325795577627\\
88	24.8325795577627\\
89	24.8325795577627\\
90	24.8325795577627\\
91	24.8325795577627\\
92	24.8325795577627\\
93	24.8325795577627\\
94	24.8325795577627\\
95	24.8325795577627\\
96	24.8325795577627\\
97	24.8325795577627\\
98	24.8325795577627\\
99	24.8325795577627\\
100	24.8325795577627\\
101	24.8325795577627\\
102	24.8325795577627\\
103	24.8325795577627\\
104	24.8325795577627\\
105	24.8325795577627\\
106	24.8325795577627\\
107	24.8325795577627\\
108	24.8325795577627\\
109	24.8325795577627\\
110	24.8325795577627\\
111	24.8325795577627\\
112	24.8325795577627\\
113	24.8325795577627\\
114	24.8325795577627\\
115	24.8325795577627\\
116	24.8325795577627\\
117	24.8325795577627\\
118	24.8325795577627\\
119	24.8325795577627\\
120	24.8325795577627\\
121	24.8325795577627\\
122	24.8325795577627\\
123	24.8325795577627\\
124	24.8325795577627\\
125	24.8325795577627\\
126	24.8325795577627\\
127	24.8325795577627\\
128	24.8325795577627\\
129	24.8325795577627\\
130	24.8325795577627\\
131	24.8325795577627\\
132	24.8325795577627\\
133	24.8325795577627\\
134	24.8325795577627\\
135	24.8325795577627\\
136	24.8325795577627\\
137	24.8325795577627\\
138	24.8325795577627\\
139	24.8325795577627\\
140	24.8325795577627\\
141	24.8325795577627\\
142	24.8325795577627\\
143	24.8325795577627\\
144	24.8325795577627\\
145	24.8325795577627\\
146	24.8325795577627\\
147	24.8325795577627\\
148	24.8325795577627\\
149	24.8325795577627\\
150	24.8325795577627\\
151	24.8325795577627\\
152	24.8325795577627\\
153	24.8325795577627\\
154	24.8325795577627\\
155	24.8325795577627\\
156	24.8325795577627\\
157	24.8325795577627\\
158	24.8325795577627\\
159	24.8325795577627\\
160	24.8325795577627\\
161	24.8325795577627\\
162	24.8325795577627\\
163	24.8325795577627\\
164	24.8325795577627\\
165	24.8325795577627\\
166	24.8325795577627\\
167	24.8325795577627\\
168	24.8325795577627\\
169	24.8325795577627\\
170	24.8325795577627\\
171	24.8325795577627\\
172	24.8325795577627\\
173	24.8325795577627\\
174	24.8325795577627\\
175	24.8325795577627\\
176	24.8325795577627\\
177	24.8325795577627\\
178	24.8325795577627\\
179	24.8325795577627\\
180	24.8325795577627\\
181	24.8325795577627\\
182	24.8325795577627\\
183	24.8325795577627\\
184	24.8325795577627\\
185	24.8325795577627\\
186	24.8325795577627\\
187	24.8325795577627\\
188	24.8325795577627\\
189	24.8325795577627\\
190	24.8325795577627\\
191	24.8325795577627\\
192	24.8325795577627\\
193	24.8325795577627\\
194	24.8325795577627\\
195	24.8325795577627\\
196	24.8325795577627\\
197	24.8325795577627\\
198	24.8325795577627\\
199	24.8325795577627\\
200	24.8325795577627\\
201	24.8325795577627\\
202	24.8325795577627\\
203	24.8325795577627\\
204	24.8325795577627\\
205	24.8325795577627\\
206	24.8325795577627\\
207	24.8325795577627\\
208	24.8325795577627\\
209	24.8325795577627\\
210	24.8325795577627\\
211	24.8325795577627\\
212	24.8325795577627\\
213	24.8325795577627\\
214	24.8325795577627\\
215	24.8325795577627\\
216	24.8325795577627\\
217	24.8325795577627\\
218	24.8325795577627\\
219	24.8325795577627\\
220	24.8325795577627\\
221	24.8325795577627\\
222	24.8325795577627\\
223	24.8325795577627\\
224	24.8325795577627\\
225	24.8325795577627\\
226	24.8325795577627\\
227	24.8325795577627\\
228	24.8325795577627\\
229	24.8325795577627\\
230	24.8325795577627\\
231	24.8325795577627\\
232	24.8325795577627\\
233	24.8325795577627\\
234	24.8325795577627\\
235	24.8325795577627\\
236	24.8325795577627\\
237	24.8325795577627\\
238	24.8325795577627\\
239	24.8325795577627\\
240	24.8325795577627\\
241	24.8325795577627\\
242	24.8325795577627\\
243	24.8325795577627\\
244	24.8325795577627\\
245	24.8325795577627\\
246	24.8325795577627\\
247	24.8325795577627\\
248	24.8325795577627\\
249	24.8325795577627\\
250	24.8325795577627\\
251	24.8325795577627\\
252	24.8325795577627\\
253	24.8325795577627\\
254	24.8325795577627\\
255	24.8325795577627\\
256	24.8325795577627\\
257	24.8325795577627\\
258	24.8325795577627\\
259	24.8325795577627\\
260	24.8325795577627\\
261	24.8325795577627\\
262	24.8325795577627\\
263	24.8325795577627\\
264	24.8325795577627\\
265	24.8325795577627\\
266	24.8325795577627\\
267	24.8325795577627\\
268	24.8325795577627\\
269	24.8325795577627\\
270	24.8325795577627\\
271	24.8325795577627\\
272	24.8325795577627\\
273	24.8325795577627\\
274	24.8325795577627\\
275	24.8325795577627\\
276	24.8325795577627\\
277	24.8325795577627\\
278	24.8325795577627\\
279	24.8325795577627\\
280	24.8325795577627\\
281	24.8325795577627\\
282	24.8325795577627\\
283	24.8325795577627\\
284	24.8325795577627\\
285	24.8325795577627\\
286	24.8325795577627\\
287	24.8325795577627\\
288	24.8325795577627\\
289	24.8325795577627\\
290	24.8325795577627\\
291	24.8325795577627\\
292	24.8325795577627\\
293	24.8325795577627\\
294	24.8325795577627\\
295	24.8325795577627\\
296	24.8325795577627\\
297	24.8325795577627\\
298	24.8325795577627\\
299	24.8325795577627\\
300	24.8325795577627\\
};
\addplot [color=green, line width=0.75pt, forget plot]
  table[row sep=crcr]{%
1	24.661462740074\\
2	24.8159341545167\\
3	24.9042355041291\\
4	24.9769261592412\\
5	25.0449073235787\\
6	25.1089030892928\\
7	25.1694709072762\\
8	25.2270635911034\\
9	25.2821016876877\\
10	25.3349094577921\\
11	25.7113094511681\\
12	25.8594718583102\\
13	25.9094061647414\\
14	25.953611842022\\
15	25.9951576032342\\
16	26.0353793469059\\
17	26.0741799535183\\
18	26.1114566828662\\
19	26.1470815892428\\
20	26.1816688199127\\
21	26.4219378382649\\
22	26.5188641456492\\
23	26.5504021432003\\
24	26.5775001362634\\
25	26.5775001362634\\
26	26.6080957991605\\
27	26.6327123088279\\
28	26.6327123088279\\
29	26.661364340458\\
30	26.6848823225676\\
31	26.7448031781751\\
32	26.7790776337336\\
33	26.8075276482403\\
34	26.8318647437185\\
35	26.8574460680328\\
36	26.879346038273\\
37	26.9020709620135\\
38	26.9230384138636\\
39	26.9438560670738\\
40	26.9638104921057\\
41	27.0441244491333\\
42	27.1067444976677\\
43	27.1251313806386\\
44	27.1417791530184\\
45	27.1417791530184\\
46	27.1417791530184\\
47	27.1417791530184\\
48	27.2110128052567\\
49	27.2518174584186\\
50	27.3096789709868\\
51	27.3719413857305\\
52	27.3848626428356\\
53	27.3848626428356\\
54	27.4352020162858\\
55	27.4751875876519\\
56	27.5095009094259\\
57	27.5455403667515\\
58	27.5728605212261\\
59	27.5827137576502\\
60	27.5827137576502\\
61	27.6320775270871\\
62	27.6415559135822\\
63	27.6415559135822\\
64	27.6787795637269\\
65	27.7075053399491\\
66	27.7331920363209\\
67	27.7587424855568\\
68	27.7808615409439\\
69	27.7884507385125\\
70	27.7884507385125\\
71	27.8107537347166\\
72	27.8214178269743\\
73	27.8692994622473\\
74	27.9015338770372\\
75	27.9060793662209\\
76	27.945134897214\\
77	27.9660491114444\\
78	27.9695960114515\\
79	27.9695960114515\\
80	27.9695960114515\\
81	27.9902230611538\\
82	28.0224224136325\\
83	28.0458513816631\\
84	28.0670476933202\\
85	28.076242160059\\
86	28.0871277150888\\
87	28.0963312820402\\
88	28.0994780337021\\
89	28.1023249675317\\
90	28.1023249675317\\
91	28.1276680038992\\
92	28.1397347149981\\
93	28.1424245073763\\
94	28.1450080629698\\
95	28.1450080629698\\
96	28.1450080629698\\
97	28.1482372671096\\
98	28.1502450663318\\
99	28.1502450663318\\
100	28.1527974649019\\
101	28.1682076928398\\
102	28.1764023431442\\
103	28.1784385100379\\
104	28.1804841038266\\
105	28.1825593742\\
106	28.1842985286133\\
107	28.1842985286133\\
108	28.1842985286133\\
109	28.1865093528748\\
110	28.1883203922207\\
111	28.1956418148269\\
112	28.2076523938868\\
113	28.2123055615512\\
114	28.213749756071\\
115	28.2153590202697\\
116	28.2166037909177\\
117	28.2166037909177\\
118	28.2166037909177\\
119	28.2166037909177\\
120	28.2166037909177\\
121	28.2286433952715\\
122	28.2378689301919\\
123	28.2417276788529\\
124	28.2427812289858\\
125	28.2439824448895\\
126	28.2451000681678\\
127	28.2451000681678\\
128	28.2451000681678\\
129	28.2451000681678\\
130	28.2451000681678\\
131	28.2576772218732\\
132	28.2614181350004\\
133	28.2624189710237\\
134	28.2633005370862\\
135	28.2633005370862\\
136	28.2633005370862\\
137	28.2633005370862\\
138	28.2633005370862\\
139	28.2633005370862\\
140	28.2633005370862\\
141	28.267932330835\\
142	28.2703409629514\\
143	28.2744587224984\\
144	28.2773942913333\\
145	28.2779034918624\\
146	28.2779034918624\\
147	28.2819124105935\\
148	28.2835483182568\\
149	28.2856930803548\\
150	28.2867461746399\\
151	28.3020853893442\\
152	28.3039745875662\\
153	28.3039745875662\\
154	28.3064521064982\\
155	28.3076698985597\\
156	28.3085501973684\\
157	28.3088585930584\\
158	28.309322973022\\
159	28.3097690111497\\
160	28.3101709015255\\
161	28.3194354293998\\
162	28.3219785249776\\
163	28.3219970750405\\
164	28.3221463680808\\
165	28.3223506645313\\
166	28.3223506645313\\
167	28.3223506645313\\
168	28.3223506645313\\
169	28.3223506645313\\
170	28.3223506645313\\
171	28.3298861627277\\
172	28.3309992496071\\
173	28.3311035812377\\
174	28.3311591672287\\
175	28.3311591672287\\
176	28.3311591672287\\
177	28.3313530376772\\
178	28.3313899159562\\
179	28.3313899159562\\
180	28.3314656448966\\
181	28.3360155826298\\
182	28.3371123591269\\
183	28.3366440906071\\
184	28.3365460635301\\
185	28.3365298500835\\
186	28.3365556230386\\
187	28.3365556230386\\
188	28.3365556230386\\
189	28.3365556230386\\
190	28.3365556230386\\
191	28.3415188096573\\
192	28.3424786241906\\
193	28.3411011679937\\
194	28.3409213669992\\
195	28.3409213669992\\
196	28.3385190242998\\
197	28.3373897646358\\
198	28.3372209304772\\
199	28.3372209304772\\
200	28.3365962055346\\
201	28.3470629384029\\
202	28.3477366963295\\
203	28.3467518188464\\
204	28.3464838991463\\
205	28.3462877149228\\
206	28.3461675845335\\
207	28.3461675845335\\
208	28.3461675845335\\
209	28.3461722484447\\
210	28.3461722484447\\
211	28.3524082870557\\
212	28.3522439171043\\
213	28.3512001182852\\
214	28.3508981289747\\
215	28.3506764131134\\
216	28.3504766806026\\
217	28.3504766806026\\
218	28.3504766806026\\
219	28.3504766806026\\
220	28.3504766806026\\
221	28.3554956683333\\
222	28.3546644618468\\
223	28.3532826897436\\
224	28.3529323224698\\
225	28.3526926591344\\
226	28.3525072670115\\
227	28.3525072670115\\
228	28.3525072670115\\
229	28.3525072670115\\
230	28.3525072670115\\
231	28.3568152449587\\
232	28.3553775440993\\
233	28.3550357490611\\
234	28.3547664426401\\
235	28.3547664426401\\
236	28.3547664426401\\
237	28.3547664426401\\
238	28.3547664426401\\
239	28.3547664426401\\
240	28.3547664426401\\
241	28.358699409184\\
242	28.3591965864516\\
243	28.3590487705008\\
244	28.3588639872622\\
245	28.358651428612\\
246	28.358651428612\\
247	28.358651428612\\
248	28.358651428612\\
249	28.3586307686374\\
250	28.3584465694251\\
251	28.3603980820034\\
252	28.3595551930964\\
253	28.359332247833\\
254	28.354018525349\\
255	28.3525887354627\\
256	28.3508068016956\\
257	28.3498224715633\\
258	28.3498224715633\\
259	28.3498224715633\\
260	28.3490826833007\\
261	28.3578842114998\\
262	28.3578648554804\\
263	28.3576075027206\\
264	28.3573367684138\\
265	28.3573367684138\\
266	28.3573367684138\\
267	28.3573367684138\\
268	28.3573367684138\\
269	28.3573367684138\\
270	28.3573367684138\\
271	28.3608288519796\\
272	28.3612278349613\\
273	28.361035637932\\
274	28.3607680509012\\
275	28.3604398409359\\
276	28.3604398409359\\
277	28.3604398409359\\
278	28.3604398409359\\
279	28.3604398409359\\
280	28.3604398409359\\
281	28.3620368888739\\
282	28.3622859660155\\
283	28.3621024870692\\
284	28.3563609858918\\
285	28.3548985837066\\
286	28.3548985837066\\
287	28.351868256445\\
288	28.3515812725116\\
289	28.3512392856544\\
290	28.3509807797025\\
291	28.3583177811561\\
292	28.3589618677397\\
293	28.3587969037397\\
294	28.3585293317711\\
295	28.3582307589644\\
296	28.3582307589644\\
297	28.3582307589644\\
298	28.3582307589644\\
299	28.3582307589644\\
300	28.3582307589644\\
};
\addplot [color=red, line width=0.75pt, forget plot]
  table[row sep=crcr]{%
1	24.661462740074\\
2	24.9312993019459\\
3	25.208446858805\\
4	25.6800233934054\\
5	25.9980056310564\\
6	26.0305723927558\\
7	26.0305723927558\\
8	26.0994895285183\\
9	26.0994895285183\\
10	26.0994895285183\\
11	26.4491724169177\\
12	26.7069233383715\\
13	26.8877426956528\\
14	26.9514705693906\\
15	26.9734471612691\\
16	26.9734471612691\\
17	26.9734471612691\\
18	26.9734471612691\\
19	27.0678590003672\\
20	27.0949449715226\\
21	27.2293001489424\\
22	27.3717518236877\\
23	27.3885244023847\\
24	27.3885244023847\\
25	27.5005283076658\\
26	27.5428867168993\\
27	27.6281574787621\\
28	27.6811521060067\\
29	27.705488735668\\
30	27.7149359700582\\
31	27.8070252130672\\
32	27.8798517221186\\
33	27.888009737407\\
34	27.888009737407\\
35	27.9472522559347\\
36	27.9733165728949\\
37	28.034592960817\\
38	28.0556108304094\\
39	28.0618613325109\\
40	28.0618613325109\\
41	28.1354794982852\\
42	28.1693862744035\\
43	28.1727468303241\\
44	28.1727468303241\\
45	28.1788499647784\\
46	28.1825108727632\\
47	28.1825108727632\\
48	28.1866299522502\\
49	28.1866299522502\\
50	28.1866299522502\\
51	28.2149465485906\\
52	28.2388571463861\\
53	28.2418750217378\\
54	28.2418750217378\\
55	28.2581497109907\\
56	28.2684115548783\\
57	28.290191842667\\
58	28.2970283275514\\
59	28.3064082082238\\
60	28.3064082082238\\
61	28.3363153243517\\
62	28.3512788295776\\
63	28.3530210690469\\
64	28.3582658251391\\
65	28.3614445759296\\
66	28.3686607321113\\
67	28.3734312534668\\
68	28.3748356212736\\
69	28.3759066755658\\
70	28.3759066755658\\
71	28.4000117176189\\
72	28.4067801185604\\
73	28.4077547588799\\
74	28.4077547588799\\
75	28.3999301848514\\
76	28.4027884138706\\
77	28.4089322479287\\
78	28.4104994295712\\
79	28.4112189545649\\
80	28.4112189545649\\
81	28.4367120916903\\
82	28.4396520726664\\
83	28.4401028742581\\
84	28.4401028742581\\
85	28.4401028742581\\
86	28.4401028742581\\
87	28.4407941371139\\
88	28.4408510413583\\
89	28.4406148308356\\
90	28.4402572736614\\
91	28.4480180210617\\
92	28.4529697988495\\
93	28.4532868343949\\
94	28.4532868343949\\
95	28.4453126328977\\
96	28.444362616766\\
97	28.4445822099679\\
98	28.4445822099679\\
99	28.4445822099679\\
100	28.4468377900185\\
101	28.4586538733912\\
102	28.4592618786945\\
103	28.4590243212572\\
104	28.4590243212572\\
105	28.458924197215\\
106	28.4587549323348\\
107	28.4587549323348\\
108	28.4585980484737\\
109	28.4585980484737\\
110	28.4583176900586\\
111	28.4621174481385\\
112	28.4614094894135\\
113	28.4613071205061\\
114	28.4613071205061\\
115	28.4506643257201\\
116	28.4504801473336\\
117	28.4448469109181\\
118	28.4433599094914\\
119	28.4430931433128\\
120	28.4429660892673\\
121	28.4545621397078\\
122	28.4535071759476\\
123	28.4507034818785\\
124	28.4501497690633\\
125	28.4501497690633\\
126	28.4501497690633\\
127	28.4501497690633\\
128	28.4501497690633\\
129	28.4501497690633\\
130	28.4501497690633\\
131	28.4534068797397\\
132	28.4526937366436\\
133	28.4524329479494\\
134	28.4524329479494\\
135	28.4418492827395\\
136	28.4414827613903\\
137	28.433366096998\\
138	28.4303861861261\\
139	28.4300109089967\\
140	28.4300109089967\\
141	28.4405581256604\\
142	28.439751214414\\
143	28.4369369619757\\
144	28.4361024403231\\
145	28.4355442922056\\
146	28.4349554657224\\
147	28.4344984888893\\
148	28.4344984888893\\
149	28.4344984888893\\
150	28.4341290360015\\
151	28.4396213444455\\
152	28.436583863538\\
153	28.4326613224739\\
154	28.4323348700869\\
155	28.4323348700869\\
156	28.4323348700869\\
157	28.4323348700869\\
158	28.4323348700869\\
159	28.4323348700869\\
160	28.4323348700869\\
161	28.4339036726139\\
162	28.4361730806533\\
163	28.4293001611549\\
164	28.4274996561853\\
165	28.4274996561853\\
166	28.4265602278727\\
167	28.4265602278727\\
168	28.4265602278727\\
169	28.4247709976673\\
170	28.4233229905075\\
171	28.4280514369671\\
172	28.4272957664906\\
173	28.426918405833\\
174	28.426918405833\\
175	28.4156095668774\\
176	28.4132724219421\\
177	28.4068991561417\\
178	28.4038007668423\\
179	28.4010761109575\\
180	28.400490584946\\
181	28.4118012573585\\
182	28.4090937796948\\
183	28.4086111815377\\
184	28.4086111815377\\
185	28.408151311432\\
186	28.407545806056\\
187	28.4070145358007\\
188	28.4070145358007\\
189	28.4063575401038\\
190	28.4059191811877\\
191	28.4109979716819\\
192	28.4096480788902\\
193	28.4093827622528\\
194	28.4093827622528\\
195	28.4014481261009\\
196	28.3995740789849\\
197	28.3995740789849\\
198	28.3980924032465\\
199	28.3947564174677\\
200	28.3947564174677\\
201	28.4032890669213\\
202	28.4023607112797\\
203	28.3978337329368\\
204	28.3974675781562\\
205	28.3974675781562\\
206	28.3915169646119\\
207	28.3877945374973\\
208	28.3873906107139\\
209	28.3873906107139\\
210	28.3849827889614\\
211	28.3961673314784\\
212	28.3947800279941\\
213	28.3931439451454\\
214	28.3931439451454\\
215	28.3913738465177\\
216	28.3913738465177\\
217	28.3913738465177\\
218	28.3899128615874\\
219	28.3896208709946\\
220	28.3896208709946\\
221	28.3950471670023\\
222	28.3939513255807\\
223	28.3934252785582\\
224	28.3934252785582\\
225	28.3934252785582\\
226	28.3928384235289\\
227	28.3922699777498\\
228	28.3922699777498\\
229	28.3916803003136\\
230	28.3916803003136\\
231	28.3940517696187\\
232	28.3923560397687\\
233	28.3918503886272\\
234	28.3809362168471\\
235	28.3785593353763\\
236	28.3785593353763\\
237	28.3769187565528\\
238	28.3761579665489\\
239	28.3758623889733\\
240	28.3758623889733\\
241	28.3840929709735\\
242	28.3833176435598\\
243	28.3830774722334\\
244	28.3830774722334\\
245	28.3830774722334\\
246	28.3830787725197\\
247	28.3829794533887\\
248	28.3829794533887\\
249	28.3828350173269\\
250	28.3828350173269\\
251	28.3857699045065\\
252	28.3847918092158\\
253	28.3845729275768\\
254	28.3845729275768\\
255	28.3775087612895\\
256	28.3757318979838\\
257	28.3705668866732\\
258	28.3678786438207\\
259	28.3678786438207\\
260	28.3665831838204\\
261	28.3774062571419\\
262	28.3761344154043\\
263	28.3758375584671\\
264	28.3758375584671\\
265	28.3755545165894\\
266	28.3753023312182\\
267	28.3753023312182\\
268	28.3753023312182\\
269	28.375103988803\\
270	28.375103988803\\
271	28.37829460455\\
272	28.3783692747138\\
273	28.3781207335826\\
274	28.3781207335826\\
275	28.3683253377809\\
276	28.3652003489481\\
277	28.361527611815\\
278	28.3601256703604\\
279	28.3601256703604\\
280	28.3591645427933\\
281	28.3704297156836\\
282	28.3704732091076\\
283	28.3690959931681\\
284	28.3687228853807\\
285	28.3687228853807\\
286	28.3687228853807\\
287	28.3685539393464\\
288	28.3683567688132\\
289	28.3681116816801\\
290	28.3677700883067\\
291	28.3724632403154\\
292	28.3718077185109\\
293	28.3715722552343\\
294	28.3715722552343\\
295	28.364657650658\\
296	28.3631119054037\\
297	28.3588819958223\\
298	28.3565749554601\\
299	28.3562956600375\\
300	28.3562956600375\\
};
\end{axis}
\end{tikzpicture}%

%% file: images/conv_SSIM_lambda.tikz
%
%
\begin{tikzpicture}

\definecolor{gray4}{rgb}{0.2431,0.2431,0.2431}
\definecolor{gray3}{rgb}{0.3098,0.3098,0.3098}
\definecolor{gray2}{rgb}{0.4863,0.4863,0.4863}
\definecolor{gray1}{rgb}{0.7725,0.7725,0.7725}

\begin{axis}[%
width=0.951\figurewidth,
height=\figureheight,
at={(0\figurewidth,0\figureheight)},
scale only axis,
unbounded coords=jump,
xmin=0,
xmax=300,
ymin=0.4,
ymax=0.8,
ylabel style={font=\color{white!15!black}},
ylabel={SSIM},
xlabel style={font=\color{white!15!black}},
xlabel={Number of inner and outer iterations},
axis background/.style={fill=white},
xmajorgrids,
ymajorgrids,
xlabel near ticks,
ylabel near ticks
]
\addplot [color=black, line width=0.75pt, forget plot] 
  table[row sep=crcr]{%
1	0.447132792013231\\
2	0.470467194213733\\
3	0.490248031609507\\
4	0.535027981419754\\
5	0.566336266612457\\
6	0.570217827611425\\
7	0.570217827611425\\
8	0.579449268183018\\
9	0.587044608658933\\
10	0.587044608658933\\
11	0.610205495867125\\
12	0.621871728540799\\
13	0.630029587856091\\
14	0.652470300538481\\
15	0.67119704611202\\
16	0.677109027309769\\
17	0.679412838884696\\
18	0.679412838884696\\
19	0.679412838884696\\
20	0.679412838884696\\
21	0.695190440975232\\
22	0.707886748925659\\
23	0.709532172568961\\
24	0.718822792992596\\
25	0.724244099915571\\
26	0.724244099915571\\
27	0.733530213638752\\
28	0.734254116315106\\
29	0.735295343292766\\
30	0.735295343292766\\
31	0.744499819483657\\
32	0.750068173308777\\
33	0.750606704841783\\
34	0.750606704841783\\
35	0.753270707093871\\
36	0.755045220689387\\
37	0.758318634642998\\
38	0.761078297524092\\
39	0.761890147205961\\
40	0.762336065569206\\
41	0.767979039843398\\
42	0.770797538547217\\
43	0.771061883050198\\
44	0.771061883050198\\
45	0.771813175976829\\
46	0.772618267266076\\
47	0.773047311737532\\
48	0.773548545062137\\
49	0.773548545062137\\
50	0.773548545062137\\
51	0.778745053914404\\
52	0.779961957373213\\
53	0.780155912620025\\
54	0.780155912620025\\
55	0.779119491245776\\
56	0.779893562981844\\
57	0.781828011917944\\
58	0.782503084530719\\
59	0.782667379285663\\
60	0.782667379285663\\
61	0.786209982608074\\
62	0.787084616150734\\
63	0.787124805524764\\
64	0.78714930305371\\
65	0.78714930305371\\
66	0.78714930305371\\
67	0.78714930305371\\
68	0.787404545661127\\
69	0.787536038366196\\
70	0.787536038366196\\
71	0.789122534331431\\
72	0.78996323285451\\
73	0.790074572622523\\
74	0.790074572622523\\
75	0.788302156356146\\
76	0.78858553949244\\
77	0.788709339657087\\
78	0.788709339657087\\
79	0.788709339657087\\
80	0.788709339657087\\
81	0.791477689874253\\
82	0.792054517362402\\
83	0.792091767837102\\
84	0.792091767837102\\
85	0.791207576068801\\
86	0.791199097359645\\
87	0.791265760902294\\
88	0.79130962052825\\
89	0.79130962052825\\
90	0.791753698345446\\
91	0.793702654481453\\
92	0.794056475481657\\
93	0.794052198589591\\
94	0.794052198589591\\
95	0.791895754415146\\
96	0.79183289067391\\
97	0.791908662793226\\
98	0.79184748518783\\
99	0.79184748518783\\
100	0.79184748518783\\
101	0.794353318671433\\
102	0.7942679172279\\
103	0.794054708362647\\
104	0.794004713772065\\
105	0.793997835715867\\
106	0.793997835715867\\
107	0.794039122276083\\
108	0.794039122276083\\
109	0.794039122276083\\
110	0.794120532740398\\
111	0.794984920623837\\
112	0.794935201030131\\
113	0.794894645159267\\
114	0.794894645159267\\
115	0.793258798278337\\
116	0.793187611865869\\
117	0.793249291279803\\
118	0.793194766805497\\
119	0.793144465491107\\
120	0.793143417506321\\
121	0.794836094162868\\
122	0.794779440272492\\
123	0.794589480181835\\
124	0.794529284792092\\
125	0.794496246304627\\
126	0.794496246304627\\
127	0.794496246304627\\
128	0.794496246304627\\
129	0.794496246304627\\
130	0.794496246304627\\
131	0.79525551178117\\
132	0.795310231142414\\
133	0.795284847799119\\
134	0.795284847799119\\
135	0.794112924887263\\
136	0.794003456890253\\
137	0.794003456890253\\
138	0.793995203053402\\
139	0.793942677599529\\
140	0.793891182406847\\
141	0.79499948310758\\
142	0.795101243023207\\
143	0.795067891761545\\
144	0.795067891761545\\
145	0.793160931796938\\
146	0.792832229407174\\
147	0.792481648933633\\
148	0.79215360998837\\
149	0.792133376255702\\
150	0.792133376255702\\
151	0.793933637253725\\
152	0.793770002770642\\
153	0.793429509347751\\
154	0.793383450689017\\
155	0.793372887741453\\
156	0.793372887741453\\
157	0.793372887741453\\
158	0.793429134417145\\
159	0.793429134417145\\
160	0.793486231272121\\
161	0.794324349254487\\
162	0.794387345237614\\
163	0.794354569841142\\
164	0.792672579891897\\
165	0.792357806213331\\
166	0.792073559012295\\
167	0.791884968039795\\
168	0.791875968585127\\
169	0.791875968585127\\
170	0.791875968585127\\
171	0.79351650702484\\
172	0.793349768359026\\
173	0.793113533086732\\
174	0.793054990364656\\
175	0.79302956621935\\
176	0.79302956621935\\
177	0.79302956621935\\
178	0.793049652020435\\
179	0.793047932783218\\
180	0.793047932783218\\
181	0.793780491644019\\
182	0.793866778001666\\
183	0.793846134940115\\
184	0.793846134940115\\
185	0.792250247603716\\
186	0.792011333875157\\
187	0.792039201825789\\
188	0.792039201825789\\
189	0.792039201825789\\
190	0.792089491095986\\
191	0.793056347649183\\
192	0.792909012093651\\
193	0.792678790273154\\
194	0.792616165331443\\
195	0.792616165331443\\
196	0.792595465660655\\
197	0.792577273252588\\
198	0.792557956743862\\
199	0.792542141615398\\
200	0.792542141615398\\
201	0.793198239437432\\
202	0.792975048791918\\
203	0.792535196568275\\
204	0.79249356365333\\
205	0.79249356365333\\
206	0.79249356365333\\
207	0.792556937680042\\
208	0.792563355791159\\
209	0.792542112612998\\
210	0.79252023354776\\
211	0.792960926919698\\
212	0.793019543063134\\
213	0.792993239645123\\
214	0.791717885109042\\
215	0.791325378951222\\
216	0.790937699039247\\
217	0.790801497560456\\
218	0.790791139706209\\
219	0.790792026155445\\
220	0.790804817011009\\
221	0.792092772716621\\
222	0.791968926368928\\
223	0.791802149550463\\
224	0.791757237817072\\
225	0.791733470491548\\
226	0.791733470491548\\
227	0.791733470491548\\
228	0.790657568985387\\
229	0.790597224724551\\
230	0.790547329883946\\
231	0.791777576424568\\
232	0.791536765571781\\
233	0.791159550389917\\
234	0.791124239009408\\
235	0.791124239009408\\
236	0.790956846291393\\
237	0.790871471183391\\
238	0.790747369440493\\
239	0.790731160320907\\
240	0.7907161348744\\
241	0.791655525916124\\
242	0.79135221689913\\
243	0.791309093118666\\
244	0.791309093118666\\
245	0.79127693699061\\
246	0.791252397026468\\
247	0.791252397026468\\
248	0.791252397026468\\
249	0.791238626416618\\
250	0.791220012828491\\
251	0.791730011392013\\
252	0.791742829174857\\
253	0.791706854135392\\
254	0.791706854135392\\
255	0.791706854135392\\
256	0.791706854135392\\
257	0.789947370850091\\
258	0.789868411383993\\
259	0.789893065039324\\
260	0.789701901844259\\
261	0.79106005492991\\
262	0.79107553472209\\
263	0.791059356303777\\
264	0.791059356303777\\
265	0.791059356303777\\
266	0.791059356303777\\
267	0.791071165861906\\
268	0.791060523667248\\
269	0.791060523667248\\
270	0.791049568501909\\
271	0.791049182101187\\
272	0.791017535442623\\
273	0.790988540431967\\
274	0.790004661856935\\
275	0.789856118114032\\
276	0.789769060990227\\
277	0.789675669295487\\
278	0.7896530947487\\
279	0.789651418404465\\
280	0.789651418404465\\
281	0.790773107832284\\
282	0.790713566200282\\
283	0.790689446731163\\
284	0.790689446731163\\
285	0.790689446731163\\
286	0.790689446731163\\
287	0.790689446731163\\
288	0.789311856035371\\
289	0.789384114971487\\
290	0.789499407393605\\
291	0.790881039388471\\
292	0.790858405546271\\
293	0.790713470501543\\
294	0.790713470501543\\
295	0.790580567848491\\
296	0.790539987859783\\
297	0.790526877732289\\
298	0.790526877732289\\
299	0.790526877732289\\
300	0.790541475997178\\
};
\addplot [color=blue, line width=0.75pt, forget plot]
  table[row sep=crcr]{%
1	0.447132792013231\\
2	0.447132792013231\\
3	0.447132792013231\\
4	0.447132792013231\\
5	0.447132792013231\\
6	0.447132792013231\\
7	0.447132792013231\\
8	0.447132792013231\\
9	0.447132792013231\\
10	0.447132792013231\\
11	0.457356954459276\\
12	0.457356954459276\\
13	0.457356954459276\\
14	0.457356954459276\\
15	0.457356954459276\\
16	0.457356954459276\\
17	0.457356954459276\\
18	0.457356954459276\\
19	0.457356954459276\\
20	0.457356954459276\\
21	0.463788217420508\\
22	0.463788217420508\\
23	0.463788217420508\\
24	0.463788217420508\\
25	0.463788217420508\\
26	0.463788217420508\\
27	0.463788217420508\\
28	0.463788217420508\\
29	0.463788217420508\\
30	0.463788217420508\\
31	0.463788217420508\\
32	0.463788217420508\\
33	0.463788217420508\\
34	0.463788217420508\\
35	0.463788217420508\\
36	0.463788217420508\\
37	0.463788217420508\\
38	0.463788217420508\\
39	0.463788217420508\\
40	0.463788217420508\\
41	0.463788217420508\\
42	0.463788217420508\\
43	0.463788217420508\\
44	0.463788217420508\\
45	0.463788217420508\\
46	0.463788217420508\\
47	0.463788217420508\\
48	0.463788217420508\\
49	0.463788217420508\\
50	0.463788217420508\\
51	0.463788217420508\\
52	0.463788217420508\\
53	0.463788217420508\\
54	0.463788217420508\\
55	0.463788217420508\\
56	0.463788217420508\\
57	0.463788217420508\\
58	0.463788217420508\\
59	0.463788217420508\\
60	0.463788217420508\\
61	0.463788217420508\\
62	0.463788217420508\\
63	0.463788217420508\\
64	0.463788217420508\\
65	0.463788217420508\\
66	0.463788217420508\\
67	0.463788217420508\\
68	0.463788217420508\\
69	0.463788217420508\\
70	0.463788217420508\\
71	0.463788217420508\\
72	0.463788217420508\\
73	0.463788217420508\\
74	0.463788217420508\\
75	0.463788217420508\\
76	0.463788217420508\\
77	0.463788217420508\\
78	0.463788217420508\\
79	0.463788217420508\\
80	0.463788217420508\\
81	0.463788217420508\\
82	0.463788217420508\\
83	0.463788217420508\\
84	0.463788217420508\\
85	0.463788217420508\\
86	0.463788217420508\\
87	0.463788217420508\\
88	0.463788217420508\\
89	0.463788217420508\\
90	0.463788217420508\\
91	0.463788217420508\\
92	0.463788217420508\\
93	0.463788217420508\\
94	0.463788217420508\\
95	0.463788217420508\\
96	0.463788217420508\\
97	0.463788217420508\\
98	0.463788217420508\\
99	0.463788217420508\\
100	0.463788217420508\\
101	0.463788217420508\\
102	0.463788217420508\\
103	0.463788217420508\\
104	0.463788217420508\\
105	0.463788217420508\\
106	0.463788217420508\\
107	0.463788217420508\\
108	0.463788217420508\\
109	0.463788217420508\\
110	0.463788217420508\\
111	0.463788217420508\\
112	0.463788217420508\\
113	0.463788217420508\\
114	0.463788217420508\\
115	0.463788217420508\\
116	0.463788217420508\\
117	0.463788217420508\\
118	0.463788217420508\\
119	0.463788217420508\\
120	0.463788217420508\\
121	0.463788217420508\\
122	0.463788217420508\\
123	0.463788217420508\\
124	0.463788217420508\\
125	0.463788217420508\\
126	0.463788217420508\\
127	0.463788217420508\\
128	0.463788217420508\\
129	0.463788217420508\\
130	0.463788217420508\\
131	0.463788217420508\\
132	0.463788217420508\\
133	0.463788217420508\\
134	0.463788217420508\\
135	0.463788217420508\\
136	0.463788217420508\\
137	0.463788217420508\\
138	0.463788217420508\\
139	0.463788217420508\\
140	0.463788217420508\\
141	0.463788217420508\\
142	0.463788217420508\\
143	0.463788217420508\\
144	0.463788217420508\\
145	0.463788217420508\\
146	0.463788217420508\\
147	0.463788217420508\\
148	0.463788217420508\\
149	0.463788217420508\\
150	0.463788217420508\\
151	0.463788217420508\\
152	0.463788217420508\\
153	0.463788217420508\\
154	0.463788217420508\\
155	0.463788217420508\\
156	0.463788217420508\\
157	0.463788217420508\\
158	0.463788217420508\\
159	0.463788217420508\\
160	0.463788217420508\\
161	0.463788217420508\\
162	0.463788217420508\\
163	0.463788217420508\\
164	0.463788217420508\\
165	0.463788217420508\\
166	0.463788217420508\\
167	0.463788217420508\\
168	0.463788217420508\\
169	0.463788217420508\\
170	0.463788217420508\\
171	0.463788217420508\\
172	0.463788217420508\\
173	0.463788217420508\\
174	0.463788217420508\\
175	0.463788217420508\\
176	0.463788217420508\\
177	0.463788217420508\\
178	0.463788217420508\\
179	0.463788217420508\\
180	0.463788217420508\\
181	0.463788217420508\\
182	0.463788217420508\\
183	0.463788217420508\\
184	0.463788217420508\\
185	0.463788217420508\\
186	0.463788217420508\\
187	0.463788217420508\\
188	0.463788217420508\\
189	0.463788217420508\\
190	0.463788217420508\\
191	0.463788217420508\\
192	0.463788217420508\\
193	0.463788217420508\\
194	0.463788217420508\\
195	0.463788217420508\\
196	0.463788217420508\\
197	0.463788217420508\\
198	0.463788217420508\\
199	0.463788217420508\\
200	0.463788217420508\\
201	0.463788217420508\\
202	0.463788217420508\\
203	0.463788217420508\\
204	0.463788217420508\\
205	0.463788217420508\\
206	0.463788217420508\\
207	0.463788217420508\\
208	0.463788217420508\\
209	0.463788217420508\\
210	0.463788217420508\\
211	0.463788217420508\\
212	0.463788217420508\\
213	0.463788217420508\\
214	0.463788217420508\\
215	0.463788217420508\\
216	0.463788217420508\\
217	0.463788217420508\\
218	0.463788217420508\\
219	0.463788217420508\\
220	0.463788217420508\\
221	0.463788217420508\\
222	0.463788217420508\\
223	0.463788217420508\\
224	0.463788217420508\\
225	0.463788217420508\\
226	0.463788217420508\\
227	0.463788217420508\\
228	0.463788217420508\\
229	0.463788217420508\\
230	0.463788217420508\\
231	0.463788217420508\\
232	0.463788217420508\\
233	0.463788217420508\\
234	0.463788217420508\\
235	0.463788217420508\\
236	0.463788217420508\\
237	0.463788217420508\\
238	0.463788217420508\\
239	0.463788217420508\\
240	0.463788217420508\\
241	0.463788217420508\\
242	0.463788217420508\\
243	0.463788217420508\\
244	0.463788217420508\\
245	0.463788217420508\\
246	0.463788217420508\\
247	0.463788217420508\\
248	0.463788217420508\\
249	0.463788217420508\\
250	0.463788217420508\\
251	0.463788217420508\\
252	0.463788217420508\\
253	0.463788217420508\\
254	0.463788217420508\\
255	0.463788217420508\\
256	0.463788217420508\\
257	0.463788217420508\\
258	0.463788217420508\\
259	0.463788217420508\\
260	0.463788217420508\\
261	0.463788217420508\\
262	0.463788217420508\\
263	0.463788217420508\\
264	0.463788217420508\\
265	0.463788217420508\\
266	0.463788217420508\\
267	0.463788217420508\\
268	0.463788217420508\\
269	0.463788217420508\\
270	0.463788217420508\\
271	0.463788217420508\\
272	0.463788217420508\\
273	0.463788217420508\\
274	0.463788217420508\\
275	0.463788217420508\\
276	0.463788217420508\\
277	0.463788217420508\\
278	0.463788217420508\\
279	0.463788217420508\\
280	0.463788217420508\\
281	0.463788217420508\\
282	0.463788217420508\\
283	0.463788217420508\\
284	0.463788217420508\\
285	0.463788217420508\\
286	0.463788217420508\\
287	0.463788217420508\\
288	0.463788217420508\\
289	0.463788217420508\\
290	0.463788217420508\\
291	0.463788217420508\\
292	0.463788217420508\\
293	0.463788217420508\\
294	0.463788217420508\\
295	0.463788217420508\\
296	0.463788217420508\\
297	0.463788217420508\\
298	0.463788217420508\\
299	0.463788217420508\\
300	0.463788217420508\\
};
\addplot [color=green, line width=0.75pt, forget plot]
  table[row sep=crcr]{%
1	0.447132792013231\\
2	0.462060242810135\\
3	0.471132696860893\\
4	0.478419541476287\\
5	0.485178267255007\\
6	0.491518004021879\\
7	0.497514548433167\\
8	0.503228512601971\\
9	0.508704159614457\\
10	0.513963275499108\\
11	0.550485004107518\\
12	0.564972991503358\\
13	0.569902515594207\\
14	0.574274453848045\\
15	0.578355041026173\\
16	0.582294260059361\\
17	0.586092295894477\\
18	0.589741204004077\\
19	0.593230983460368\\
20	0.596617650910734\\
21	0.620840608753559\\
22	0.630080142446619\\
23	0.6330489334039\\
24	0.635609105048113\\
25	0.635609105048113\\
26	0.638501300385841\\
27	0.640812706403952\\
28	0.640812706403952\\
29	0.643517377220075\\
30	0.645738910673704\\
31	0.652016573603617\\
32	0.655468839148969\\
33	0.658156426935687\\
34	0.66040880869975\\
35	0.662749587213224\\
36	0.664749654083582\\
37	0.66681952006209\\
38	0.668734816814096\\
39	0.670637104062467\\
40	0.672458505020307\\
41	0.680310747018381\\
42	0.685870129070814\\
43	0.687480754195873\\
44	0.688937985678217\\
45	0.688937985678217\\
46	0.688937985678217\\
47	0.688937985678217\\
48	0.694800313652649\\
49	0.698325230522795\\
50	0.703319001151204\\
51	0.70944978198044\\
52	0.710669393452145\\
53	0.710669393452145\\
54	0.714561615749764\\
55	0.7178223352242\\
56	0.720662663276402\\
57	0.723668840443377\\
58	0.72596367291397\\
59	0.726791772801772\\
60	0.726791772801772\\
61	0.731659309538342\\
62	0.732533049751709\\
63	0.732533049751709\\
64	0.735210512439889\\
65	0.737403590867055\\
66	0.739420461951451\\
67	0.741460008755623\\
68	0.743245733401529\\
69	0.74386594181866\\
70	0.74386594181866\\
71	0.74635155210068\\
72	0.747388780332828\\
73	0.750620112537326\\
74	0.752988460230478\\
75	0.753352599339754\\
76	0.756448911300236\\
77	0.75808271586469\\
78	0.758369384661841\\
79	0.758369384661841\\
80	0.758369384661841\\
81	0.760746698416543\\
82	0.763169174081107\\
83	0.76464422172874\\
84	0.765976325937017\\
85	0.76663363917872\\
86	0.767491377355492\\
87	0.768249810028499\\
88	0.768515603146205\\
89	0.76876313868557\\
90	0.76876313868557\\
91	0.771474369688549\\
92	0.772350447659543\\
93	0.772511496846872\\
94	0.772672993876112\\
95	0.772672993876112\\
96	0.772672993876112\\
97	0.772898587656112\\
98	0.773027489328243\\
99	0.773027489328243\\
100	0.773206274793927\\
101	0.774677929561396\\
102	0.775277368908163\\
103	0.775397478280767\\
104	0.775516368808069\\
105	0.775631243021113\\
106	0.775729155344951\\
107	0.775729155344951\\
108	0.775729155344951\\
109	0.77589025603663\\
110	0.776016027707585\\
111	0.776830226464656\\
112	0.777505674295025\\
113	0.777743489871918\\
114	0.777815741618167\\
115	0.777919446046924\\
116	0.778002533294162\\
117	0.778002533294162\\
118	0.778002533294162\\
119	0.778002533294162\\
120	0.778002533294162\\
121	0.779249026979106\\
122	0.779759992018008\\
123	0.7799021558129\\
124	0.77994864741205\\
125	0.78002404565242\\
126	0.780100654122177\\
127	0.780100654122177\\
128	0.780100654122177\\
129	0.780100654122177\\
130	0.780100654122177\\
131	0.781260830198903\\
132	0.781542421488827\\
133	0.781594285961283\\
134	0.781636990172025\\
135	0.781636990172025\\
136	0.781636990172025\\
137	0.781636990172025\\
138	0.781636990172025\\
139	0.781636990172025\\
140	0.781636990172025\\
141	0.782151519490336\\
142	0.78239823798268\\
143	0.782162246761987\\
144	0.782163399683208\\
145	0.782186748867787\\
146	0.782186748867787\\
147	0.782478400338949\\
148	0.782551637332125\\
149	0.782639636780604\\
150	0.782687634825155\\
151	0.784282748848973\\
152	0.784511653029503\\
153	0.784511653029503\\
154	0.78462667407536\\
155	0.784680227191677\\
156	0.784706023291003\\
157	0.784725408710148\\
158	0.784763573803938\\
159	0.784804435667216\\
160	0.784843889582704\\
161	0.785865948928669\\
162	0.785914396047777\\
163	0.785829863601936\\
164	0.785825068796295\\
165	0.785835021351116\\
166	0.785835021351116\\
167	0.785835021351116\\
168	0.785835021351116\\
169	0.785835021351116\\
170	0.785835021351116\\
171	0.786639498979209\\
172	0.786733062567216\\
173	0.786723102086679\\
174	0.786720315990546\\
175	0.786720315990546\\
176	0.786720315990546\\
177	0.78674073544117\\
178	0.786740175199998\\
179	0.786740175199998\\
180	0.786748298206412\\
181	0.787247934569965\\
182	0.787209234263106\\
183	0.787059850632406\\
184	0.787027692805028\\
185	0.787014538995887\\
186	0.787012858459717\\
187	0.787012858459717\\
188	0.787012858459717\\
189	0.787012858459717\\
190	0.787012858459717\\
191	0.787656065044592\\
192	0.787591694969247\\
193	0.787312946623783\\
194	0.787280565262355\\
195	0.787280565262355\\
196	0.786988068030004\\
197	0.786861574011644\\
198	0.786845667970019\\
199	0.786845667970019\\
200	0.786829194386573\\
201	0.787992336036175\\
202	0.788009300092389\\
203	0.787884176575278\\
204	0.787853831434182\\
205	0.78783759327847\\
206	0.787830817928201\\
207	0.787830817928201\\
208	0.787830817928201\\
209	0.787850690848037\\
210	0.787850690848037\\
211	0.78858840364421\\
212	0.788415836136083\\
213	0.788240563810116\\
214	0.788195849028523\\
215	0.788167655097433\\
216	0.788145309778544\\
217	0.788145309778544\\
218	0.788145309778544\\
219	0.788145309778544\\
220	0.788145309778544\\
221	0.788809411644488\\
222	0.788664364040478\\
223	0.788493774815554\\
224	0.788455206583341\\
225	0.788435639584886\\
226	0.788425917470064\\
227	0.788425917470064\\
228	0.788425917470064\\
229	0.788425917470064\\
230	0.788425917470064\\
231	0.788940562627257\\
232	0.788748679711306\\
233	0.788703862440887\\
234	0.788672143722784\\
235	0.788672143722784\\
236	0.788672143722784\\
237	0.788672143722784\\
238	0.788672143722784\\
239	0.788672143722784\\
240	0.788672143722784\\
241	0.789102149054545\\
242	0.789168718411358\\
243	0.789143254953942\\
244	0.789115479531559\\
245	0.789084104477217\\
246	0.789084104477217\\
247	0.789084104477217\\
248	0.789084104477217\\
249	0.789084620228727\\
250	0.789058853888333\\
251	0.789278684469578\\
252	0.789187687869574\\
253	0.78916276194422\\
254	0.788503094178935\\
255	0.788357206170862\\
256	0.788202553901134\\
257	0.788130147396096\\
258	0.788130147396096\\
259	0.788130147396096\\
260	0.788157232577753\\
261	0.789139329909055\\
262	0.789102413836712\\
263	0.789060421441258\\
264	0.789019547047807\\
265	0.789019547047807\\
266	0.789019547047807\\
267	0.789019547047807\\
268	0.789019547047807\\
269	0.789019547047807\\
270	0.789019547047807\\
271	0.789453687229806\\
272	0.789529950418675\\
273	0.789508904409397\\
274	0.789481593244845\\
275	0.789446392732605\\
276	0.789446392732605\\
277	0.789446392732605\\
278	0.789446392732605\\
279	0.789446392732605\\
280	0.789446392732605\\
281	0.789621119601656\\
282	0.789665631526734\\
283	0.789636560101454\\
284	0.788895931738449\\
285	0.788741539692387\\
286	0.788741539692387\\
287	0.788535811163691\\
288	0.788515786136774\\
289	0.788494158621668\\
290	0.788480056512291\\
291	0.789364942886859\\
292	0.789449344786731\\
293	0.789428986862201\\
294	0.789396843916227\\
295	0.789361503943614\\
296	0.789361503943614\\
297	0.789361503943614\\
298	0.789361503943614\\
299	0.789361503943614\\
300	0.789361503943614\\
};
\addplot [color=red, line width=0.75pt, forget plot]
  table[row sep=crcr]{%
1	0.447132792013231\\
2	0.470290238807149\\
3	0.481777346117096\\
4	0.531777497452313\\
5	0.566970159580661\\
6	0.57054221224309\\
7	0.57054221224309\\
8	0.578492677374971\\
9	0.578492677374971\\
10	0.578492677374971\\
11	0.612672640324817\\
12	0.636929646450859\\
13	0.653350960351759\\
14	0.65953784985606\\
15	0.661872823204455\\
16	0.661872823204455\\
17	0.661872823204455\\
18	0.661872823204455\\
19	0.671416456098492\\
20	0.673744437641287\\
21	0.686634375888961\\
22	0.700435445746797\\
23	0.701979394175707\\
24	0.701979394175707\\
25	0.708958345036712\\
26	0.71320111305797\\
27	0.721872461695826\\
28	0.727058698006499\\
29	0.729429587546768\\
30	0.730343464188618\\
31	0.739297700470442\\
32	0.745651910326358\\
33	0.746335605452918\\
34	0.746335605452918\\
35	0.74947969292623\\
36	0.751827779340322\\
37	0.757286626943826\\
38	0.759146790848947\\
39	0.759712357213135\\
40	0.759712357213135\\
41	0.766988602134982\\
42	0.769576464286326\\
43	0.76979152492\\
44	0.76979152492\\
45	0.770235238949605\\
46	0.770489450183069\\
47	0.770489450183069\\
48	0.770796228131974\\
49	0.770796228131974\\
50	0.770796228131974\\
51	0.773453369788742\\
52	0.775383112510685\\
53	0.775612432287588\\
54	0.775612432287588\\
55	0.775879139981944\\
56	0.77673013328537\\
57	0.778491857180373\\
58	0.779088795774479\\
59	0.779930114370194\\
60	0.779930114370194\\
61	0.783371636277278\\
62	0.784641784364181\\
63	0.784749398902769\\
64	0.784143953871289\\
65	0.784369346782161\\
66	0.785143183299918\\
67	0.785697834600841\\
68	0.78587016453011\\
69	0.786007168975595\\
70	0.786007168975595\\
71	0.789025017438113\\
72	0.78961542923971\\
73	0.789664679604682\\
74	0.789664679604682\\
75	0.787632259519056\\
76	0.787823402249553\\
77	0.788607002010697\\
78	0.788847565887439\\
79	0.78895393180209\\
80	0.78895393180209\\
81	0.792193621016577\\
82	0.792445304367264\\
83	0.792465106841278\\
84	0.792465106841278\\
85	0.792465106841278\\
86	0.792465106841278\\
87	0.792463131226713\\
88	0.792431216467172\\
89	0.79236345867345\\
90	0.792300299857396\\
91	0.793284941021198\\
92	0.793683437022848\\
93	0.793684180670862\\
94	0.793684180670862\\
95	0.792439922565578\\
96	0.792276323468846\\
97	0.792306386304956\\
98	0.792306386304956\\
99	0.792306386304956\\
100	0.792760955525396\\
101	0.794232463009814\\
102	0.79431940286859\\
103	0.794287935739983\\
104	0.794287935739983\\
105	0.794287472116596\\
106	0.794276957353331\\
107	0.794276957353331\\
108	0.794276055422362\\
109	0.794276055422362\\
110	0.794273628287345\\
111	0.794826368179152\\
112	0.794759464591434\\
113	0.79474709573518\\
114	0.79474709573518\\
115	0.793507354233093\\
116	0.793503226379241\\
117	0.793098717572822\\
118	0.793013043877035\\
119	0.793014373251505\\
120	0.793034563650732\\
121	0.794624599413808\\
122	0.794520522797944\\
123	0.794212705243536\\
124	0.7941586183693\\
125	0.7941586183693\\
126	0.7941586183693\\
127	0.7941586183693\\
128	0.7941586183693\\
129	0.7941586183693\\
130	0.7941586183693\\
131	0.794750106188061\\
132	0.794800407292349\\
133	0.794779890617291\\
134	0.794779890617291\\
135	0.79369043592412\\
136	0.79367506549251\\
137	0.793030953250082\\
138	0.792812105868965\\
139	0.792790566632549\\
140	0.792790566632549\\
141	0.794201217075675\\
142	0.794161046308773\\
143	0.79386034734478\\
144	0.793775125599641\\
145	0.793731384230119\\
146	0.79369028820515\\
147	0.793662799705016\\
148	0.793662799705016\\
149	0.793662799705016\\
150	0.793666005949741\\
151	0.794416755381846\\
152	0.794147999630068\\
153	0.793772393793023\\
154	0.793743078254482\\
155	0.793743078254482\\
156	0.793743078254482\\
157	0.793743078254482\\
158	0.793743078254482\\
159	0.793743078254482\\
160	0.793743078254482\\
161	0.794058835066569\\
162	0.794363588075428\\
163	0.793474156146209\\
164	0.793284562620582\\
165	0.793284562620582\\
166	0.793286074728911\\
167	0.793286074728911\\
168	0.793286074728911\\
169	0.793233689206095\\
170	0.793141531511986\\
171	0.793860852096232\\
172	0.79386562454185\\
173	0.793828524581924\\
174	0.793828524581924\\
175	0.792604658024988\\
176	0.792403377715233\\
177	0.791894601174959\\
178	0.791662415587128\\
179	0.791485704710753\\
180	0.791455139183177\\
181	0.792860728791538\\
182	0.792596712459613\\
183	0.792552969645818\\
184	0.792552969645818\\
185	0.792525209379185\\
186	0.79248716653895\\
187	0.792452604706699\\
188	0.792452604706699\\
189	0.792420575170458\\
190	0.792399267406467\\
191	0.793052640990337\\
192	0.792874789801056\\
193	0.792843764196731\\
194	0.792843764196731\\
195	0.791867073349264\\
196	0.791712554484508\\
197	0.791712554484508\\
198	0.791695583060244\\
199	0.79149441115542\\
200	0.79149441115542\\
201	0.792567311979292\\
202	0.79255678569088\\
203	0.792095697554162\\
204	0.79205884978254\\
205	0.79205884978254\\
206	0.791577031192652\\
207	0.791283609438605\\
208	0.791256884979869\\
209	0.791256884979869\\
210	0.791131935662696\\
211	0.792428868720232\\
212	0.792193340852199\\
213	0.79199019332952\\
214	0.79199019332952\\
215	0.79181301168343\\
216	0.79181301168343\\
217	0.79181301168343\\
218	0.791703520400907\\
219	0.791683348445013\\
220	0.791683348445013\\
221	0.792423147622653\\
222	0.792303183009516\\
223	0.79224151533399\\
224	0.79224151533399\\
225	0.79224151533399\\
226	0.792173895026971\\
227	0.792108107905008\\
228	0.792108107905008\\
229	0.792047630755068\\
230	0.792047630755068\\
231	0.792396225615333\\
232	0.79223275201706\\
233	0.792178235724272\\
234	0.790965717602639\\
235	0.790720116788337\\
236	0.790720116788337\\
237	0.790633067164283\\
238	0.790587101162469\\
239	0.790572543738014\\
240	0.790572543738014\\
241	0.791710044930149\\
242	0.791623867134927\\
243	0.791596245658758\\
244	0.791596245658758\\
245	0.791596245658758\\
246	0.791597489464591\\
247	0.791588442573491\\
248	0.791588442573491\\
249	0.791578824124892\\
250	0.791578824124892\\
251	0.791980871462237\\
252	0.791841022437192\\
253	0.79180747525759\\
254	0.79180747525759\\
255	0.790842499094885\\
256	0.790690781280196\\
257	0.790253597644975\\
258	0.790031752781721\\
259	0.790031752781721\\
260	0.789972396793713\\
261	0.791345520275618\\
262	0.79123756851464\\
263	0.791205611502664\\
264	0.791205611502664\\
265	0.791178913992618\\
266	0.791155625794492\\
267	0.791155625794492\\
268	0.791155625794492\\
269	0.791146413191546\\
270	0.791146413191546\\
271	0.79162359360852\\
272	0.791655367111748\\
273	0.791622844176433\\
274	0.791622844176433\\
275	0.790441230401287\\
276	0.790148258160516\\
277	0.789828544792306\\
278	0.789714641865925\\
279	0.789714641865925\\
280	0.789694265303313\\
281	0.791078111467331\\
282	0.791117248138379\\
283	0.790963880845501\\
284	0.790925930772214\\
285	0.790925930772214\\
286	0.790925930772214\\
287	0.790923333327581\\
288	0.790907907259015\\
289	0.790891173801203\\
290	0.790866188658393\\
291	0.791411771832706\\
292	0.791315119129295\\
293	0.791285622682948\\
294	0.791285622682948\\
295	0.790436271584062\\
296	0.790292300906574\\
297	0.789909947842007\\
298	0.789713135326386\\
299	0.78969371129247\\
300	0.78969371129247\\
};
\end{axis}
\end{tikzpicture}%

%% file: images/conv_PSNR_alpha.tikz
%
%
\begin{tikzpicture}

\definecolor{gray4}{rgb}{0.2431,0.2431,0.2431}
\definecolor{gray3}{rgb}{0.3098,0.3098,0.3098}
\definecolor{gray2}{rgb}{0.4863,0.4863,0.4863}
\definecolor{gray1}{rgb}{0.7725,0.7725,0.7725}

\begin{axis}[%
width=0.951\figurewidth,
height=\figureheight,
at={(0\figurewidth,0\figureheight)},
scale only axis,
unbounded coords=jump,
xmin=0,
xmax=300,
ymin=24.5,
ymax=29.5,
ylabel style={font=\color{white!15!black}},
ylabel={PSNR},
xlabel style={font=\color{white!15!black}},
xlabel={Number of inner and outer iterations},
axis background/.style={fill=white},
xmajorgrids,
ymajorgrids,
xlabel near ticks,
ylabel near ticks
]
\addplot [color=black, line width=0.75pt, forget plot]
  table[row sep=crcr]{%
1	24.661462740074\\
2	24.9321012952417\\
3	25.3357945031032\\
4	25.7331697773387\\
5	26.0091794206517\\
6	26.0432311918526\\
7	26.0432311918526\\
8	26.1189459100779\\
9	26.1809963127787\\
10	26.1809963127787\\
11	26.4065956756952\\
12	26.5241257475114\\
13	26.7024329776134\\
14	26.9092666521803\\
15	27.0884279107541\\
16	27.1454110929849\\
17	27.1675096114449\\
18	27.1675096114449\\
19	27.1675096114449\\
20	27.1675096114449\\
21	27.3270721169274\\
22	27.4589899156044\\
23	27.4769349943889\\
24	27.6031497597997\\
25	27.6652577707251\\
26	27.6652577707251\\
27	27.7589179727578\\
28	27.7663442015763\\
29	27.7770750278091\\
30	27.7770750278091\\
31	27.8746789684252\\
32	27.9358185590299\\
33	27.9426473812919\\
34	27.9426473812919\\
35	28.0005846220551\\
36	28.0197615052225\\
37	28.0515890809355\\
38	28.080260491069\\
39	28.0887816091496\\
40	28.0933930682979\\
41	28.1469793257639\\
42	28.1813984622714\\
43	28.1849675224504\\
44	28.1849675224504\\
45	28.2103697269095\\
46	28.2199829959291\\
47	28.2240897080484\\
48	28.2286496912577\\
49	28.2286496912577\\
50	28.2286496912577\\
51	28.2768903698878\\
52	28.2908508085315\\
53	28.2940025240988\\
54	28.2940025240988\\
55	28.3055058655239\\
56	28.3140167726574\\
57	28.3343823614469\\
58	28.3421667156594\\
59	28.3439039858068\\
60	28.3439039858068\\
61	28.3740962477303\\
62	28.3857971557645\\
63	28.3886731670152\\
64	28.3893371966065\\
65	28.3893371966065\\
66	28.3893371966065\\
67	28.3893371966065\\
68	28.3916835758049\\
69	28.392978968135\\
70	28.392978968135\\
71	28.4070372480684\\
72	28.4156441881651\\
73	28.4174119044277\\
74	28.4174119044277\\
75	28.4116161934599\\
76	28.4151623880064\\
77	28.4161221621797\\
78	28.4161221621797\\
79	28.4161221621797\\
80	28.4161221621797\\
81	28.4379217662407\\
82	28.4452381470354\\
83	28.4458643093591\\
84	28.4458643093591\\
85	28.4424296442057\\
86	28.44328352436\\
87	28.4438876981073\\
88	28.4441479535319\\
89	28.4441479535319\\
90	28.4465983099513\\
91	28.4616465326783\\
92	28.4648817965239\\
93	28.465058336535\\
94	28.465058336535\\
95	28.4496171786245\\
96	28.4487374554013\\
97	28.4473021061014\\
98	28.4448402580824\\
99	28.4448402580824\\
100	28.4448402580824\\
101	28.4637963073598\\
102	28.4628673806232\\
103	28.4609151648362\\
104	28.4603281790402\\
105	28.4601324060588\\
106	28.4601324060588\\
107	28.4601798794023\\
108	28.4601798794023\\
109	28.4601798794023\\
110	28.4603147543705\\
111	28.4658634083988\\
112	28.4652415972414\\
113	28.4649026769086\\
114	28.4649026769086\\
115	28.451189332241\\
116	28.449642456852\\
117	28.4486023226322\\
118	28.4462890704584\\
119	28.4445748910758\\
120	28.4442560005281\\
121	28.4572539428556\\
122	28.4566928575593\\
123	28.4550193376227\\
124	28.4544260023609\\
125	28.4540256209529\\
126	28.4540256209529\\
127	28.4540256209529\\
128	28.4540256209529\\
129	28.4540256209529\\
130	28.4540256209529\\
131	28.4590421570664\\
132	28.4587219629188\\
133	28.4584836470209\\
134	28.4584836470209\\
135	28.4479102595015\\
136	28.446180167657\\
137	28.446180167657\\
138	28.444214421114\\
139	28.4427573204529\\
140	28.4408827082938\\
141	28.4491892735283\\
142	28.4496640577661\\
143	28.4493640506329\\
144	28.4493640506329\\
145	28.4328165925794\\
146	28.4285245515492\\
147	28.4224543536884\\
148	28.4172757305859\\
149	28.4168529896008\\
150	28.4168529896008\\
151	28.4316008604439\\
152	28.4305997089892\\
153	28.4273961847431\\
154	28.4268378780291\\
155	28.4265832301787\\
156	28.4265832301787\\
157	28.4265832301787\\
158	28.4266187352579\\
159	28.4266187352579\\
160	28.42658497176\\
161	28.4327153715379\\
162	28.4332035168114\\
163	28.4329482770475\\
164	28.418893922508\\
165	28.4155424148013\\
166	28.4111677138874\\
167	28.4081154408482\\
168	28.4078342825801\\
169	28.4078342825801\\
170	28.4078342825801\\
171	28.4193860055315\\
172	28.4174191303163\\
173	28.4152044951845\\
174	28.4146030248282\\
175	28.4142782663104\\
176	28.4142782663104\\
177	28.4142782663104\\
178	28.4141948600786\\
179	28.4140257679285\\
180	28.4140257679285\\
181	28.4193953109161\\
182	28.4198570118008\\
183	28.4196746145212\\
184	28.4196746145212\\
185	28.4065455004148\\
186	28.4039941783403\\
187	28.4041014992216\\
188	28.4041014992216\\
189	28.4041014992216\\
190	28.4016476439046\\
191	28.4086445230235\\
192	28.4067106636054\\
193	28.4045453085952\\
194	28.4039666510797\\
195	28.4039666510797\\
196	28.4036510505211\\
197	28.403421785149\\
198	28.4030810820104\\
199	28.4027856649085\\
200	28.4027856649085\\
201	28.4078499006352\\
202	28.4060225423803\\
203	28.4021485834058\\
204	28.4017976881259\\
205	28.4017976881259\\
206	28.4017976881259\\
207	28.4020145750381\\
208	28.401903834992\\
209	28.4015185858676\\
210	28.4012274000956\\
211	28.4039448865431\\
212	28.4042470142986\\
213	28.4040111974711\\
214	28.3929182627293\\
215	28.3888202714229\\
216	28.3839995323089\\
217	28.3821014203336\\
218	28.3818677023448\\
219	28.381616608838\\
220	28.3814085167664\\
221	28.3912355195834\\
222	28.3903615261011\\
223	28.3888769246313\\
224	28.3884134459618\\
225	28.3881113619976\\
226	28.3881113619976\\
227	28.3881113619976\\
228	28.3766219596547\\
229	28.3754897713916\\
230	28.3741972790341\\
231	28.3846168908636\\
232	28.382801334828\\
233	28.3794866176603\\
234	28.3791148926632\\
235	28.3791148926632\\
236	28.3767221969342\\
237	28.3755417896334\\
238	28.3738794295375\\
239	28.3736412730622\\
240	28.3733750446014\\
241	28.3807254105159\\
242	28.3781004402127\\
243	28.3777181479369\\
244	28.3777181479369\\
245	28.3773902992663\\
246	28.3771144957122\\
247	28.3771144957122\\
248	28.3771144957122\\
249	28.3768244340429\\
250	28.3765452226514\\
251	28.3804320856303\\
252	28.3805348165759\\
253	28.3802883431126\\
254	28.3802883431126\\
255	28.3802883431126\\
256	28.3802883431126\\
257	28.3655045955205\\
258	28.3634193704615\\
259	28.3635119210757\\
260	28.3603923022777\\
261	28.3718206247713\\
262	28.3720961892261\\
263	28.3719783510115\\
264	28.3719783510115\\
265	28.3719783510115\\
266	28.3719783510115\\
267	28.3721504382805\\
268	28.3721029638899\\
269	28.3721029638899\\
270	28.3720000917789\\
271	28.371359657877\\
272	28.3708149187223\\
273	28.3705884593052\\
274	28.3620822107444\\
275	28.3606316537249\\
276	28.3594533833691\\
277	28.358119182289\\
278	28.3577190490896\\
279	28.3575836204223\\
280	28.3575836204223\\
281	28.3665490613888\\
282	28.3664323596016\\
283	28.3662755293831\\
284	28.3662755293831\\
285	28.3662755293831\\
286	28.3662755293831\\
287	28.3662755293831\\
288	28.3554897643825\\
289	28.3557836637057\\
290	28.3560772641361\\
291	28.3673976335766\\
292	28.3669945005484\\
293	28.3658070173121\\
294	28.3658070173121\\
295	28.3646630739636\\
296	28.3642772415251\\
297	28.3640939027895\\
298	28.3640939027895\\
299	28.3640939027895\\
300	28.3640487529391\\
};
\addplot [color=blue, line width=0.75pt, forget plot]
  table[row sep=crcr]{%
1	27.0865593097197\\
2	28.4794036700731\\
3	28.5322270950029\\
4	28.419213434884\\
5	28.3615442867505\\
6	28.3468948133662\\
7	28.3523173756132\\
8	28.3657986854641\\
9	28.3815846865619\\
10	28.3970227504584\\
11	28.2127167696265\\
12	28.2379180440787\\
13	28.2580594864102\\
14	28.2776575629411\\
15	28.2949074725611\\
16	28.3088007275222\\
17	28.3196675264509\\
18	28.3282154962222\\
19	28.3352353267752\\
20	28.3379855890866\\
21	28.2454629536635\\
22	28.2673211810975\\
23	28.2805409024501\\
24	28.2932366606157\\
25	28.3048479624407\\
26	28.3136880434756\\
27	28.319712982976\\
28	28.3236753949401\\
29	28.324703883716\\
30	28.326284558929\\
31	28.2358563119143\\
32	28.2566983366036\\
33	28.2686019851405\\
34	28.2812589537366\\
35	28.2927969314211\\
36	28.3013352634004\\
37	28.3068433403077\\
38	28.3101878153343\\
39	28.310853026946\\
40	28.3118179967789\\
41	28.2276585235416\\
42	28.250362901015\\
43	28.261981483373\\
44	28.2741586641701\\
45	28.2853185859873\\
46	28.2938560343146\\
47	28.2994539192488\\
48	28.3029462698186\\
49	28.3038487315812\\
50	28.3049013032237\\
51	28.2184726195261\\
52	28.2422890728582\\
53	28.2541491885331\\
54	28.2665841295535\\
55	28.2781763359218\\
56	28.2869835333564\\
57	28.2927733279009\\
58	28.2962763379708\\
59	28.2971115764967\\
60	28.2980925685089\\
61	28.2142340327231\\
62	28.2401894939632\\
63	28.2522960639726\\
64	28.2649452539487\\
65	28.2767071576925\\
66	28.2855745977224\\
67	28.2913560915757\\
68	28.2947505001315\\
69	28.2954513402343\\
70	28.2961980384757\\
71	28.2123649461124\\
72	28.2380940591849\\
73	28.2496143254895\\
74	28.2611604457034\\
75	28.271767738015\\
76	28.2800518619907\\
77	28.2858282847211\\
78	28.289637579107\\
79	28.2909387128451\\
80	28.2923847313411\\
81	28.2129986631934\\
82	28.241369649719\\
83	28.2543843623285\\
84	28.2663284714399\\
85	28.2767329847817\\
86	28.2842780977463\\
87	28.2891608509856\\
88	28.2921783799863\\
89	28.292950498221\\
90	28.2938688446674\\
91	28.2129357343261\\
92	28.2385571673569\\
93	28.2509235345745\\
94	28.2629720416608\\
95	28.2739143750376\\
96	28.2823106826249\\
97	28.2879665272842\\
98	28.2897551079014\\
99	28.2930634081862\\
100	28.2937793589805\\
101	28.2157821998964\\
102	28.2413738917869\\
103	28.2539018338734\\
104	28.2667357487132\\
105	28.2783583256718\\
106	28.2871798223141\\
107	28.293157976939\\
108	28.2951192466907\\
109	28.2985841194207\\
110	28.2993970611277\\
111	28.2185769851129\\
112	28.240069430181\\
113	28.2502885306811\\
114	28.2611810288038\\
115	28.2716053233631\\
116	28.2797216216023\\
117	28.2851433516488\\
118	28.2869575356966\\
119	28.2903357127073\\
120	28.291162039408\\
121	28.2193152273177\\
122	28.2441395737618\\
123	28.256467343097\\
124	28.2686016542728\\
125	28.2794401504333\\
126	28.2875197878804\\
127	28.2928692176019\\
128	28.2945947494044\\
129	28.2979807411546\\
130	28.2988124039043\\
131	28.2187151096686\\
132	28.2424923352432\\
133	28.2529912868044\\
134	28.2633756719874\\
135	28.2731042865468\\
136	28.2806894505575\\
137	28.2858489223401\\
138	28.2875664962327\\
139	28.2907092644743\\
140	28.2914243895529\\
141	28.2108488528163\\
142	28.2344711771462\\
143	28.2444970252082\\
144	28.2545257444823\\
145	28.2641471479844\\
146	28.2718845567552\\
147	28.2773136228332\\
148	28.2791581814248\\
149	28.2809554166934\\
150	28.2825451342306\\
151	28.212617303692\\
152	28.2362034362911\\
153	28.2474778160221\\
154	28.2585618135593\\
155	28.2689528189312\\
156	28.277163958706\\
157	28.2829452140582\\
158	28.2850021694\\
159	28.2869856319794\\
160	28.288727081553\\
161	28.2184744536272\\
162	28.2425507508011\\
163	28.2535354382819\\
164	28.2642989560107\\
165	28.2741869919139\\
166	28.2819056816648\\
167	28.2872046024109\\
168	28.289014732169\\
169	28.2907476231507\\
170	28.2922926846454\\
171	28.2205557917935\\
172	28.2439663174698\\
173	28.2543422782581\\
174	28.2644167955386\\
175	28.2737083833123\\
176	28.2810346704772\\
177	28.2860990944223\\
178	28.2878119053731\\
179	28.2894612416321\\
180	28.2909150643042\\
181	28.2221731229794\\
182	28.2448525057478\\
183	28.2546190993109\\
184	28.2645329068823\\
185	28.2739245234617\\
186	28.2812643077121\\
187	28.2862521767964\\
188	28.2879167944311\\
189	28.2895478987624\\
190	28.2910608003212\\
191	28.2182371098667\\
192	28.2405883398138\\
193	28.2512701811454\\
194	28.2616799407632\\
195	28.2708967468478\\
196	28.2777992074362\\
197	28.282311469391\\
198	28.2836957254128\\
199	28.284917009106\\
200	28.2859575353957\\
201	28.2169671200058\\
202	28.2392958066664\\
203	28.2496839294073\\
204	28.2608424741288\\
205	28.2715467620368\\
206	28.2799761427851\\
207	28.2858818562544\\
208	28.2879864254672\\
209	28.2899658786213\\
210	28.2916900555275\\
211	28.2206952525415\\
212	28.2414520732848\\
213	28.2506103040226\\
214	28.2600851083787\\
215	28.2692235627109\\
216	28.2764764314249\\
217	28.2815180338146\\
218	28.2832198424085\\
219	28.2848503796099\\
220	28.2862871817042\\
221	28.219043123254\\
222	28.2431040011654\\
223	28.2531821179111\\
224	28.2632297008689\\
225	28.272056698462\\
226	28.2786197706639\\
227	28.2830197967092\\
228	28.2843611694973\\
229	28.2856343922389\\
230	28.2867854867886\\
231	28.2195834307637\\
232	28.242262490838\\
233	28.2523722241708\\
234	28.262338167044\\
235	28.2714635118989\\
236	28.2786954251385\\
237	28.2837689748455\\
238	28.2855467638522\\
239	28.2873586791582\\
240	28.2890133974295\\
241	28.2233512461507\\
242	28.2452771517827\\
243	28.2548460785516\\
244	28.2648206241089\\
245	28.2739866525622\\
246	28.2809793528257\\
247	28.2856911704073\\
248	28.2872014592448\\
249	28.288646516676\\
250	28.2898826241865\\
251	28.2211932458075\\
252	28.2413038474858\\
253	28.2507256361021\\
254	28.2604969321407\\
255	28.2695857094542\\
256	28.2766136830437\\
257	28.2814584077292\\
258	28.2832507793004\\
259	28.2850966918982\\
260	28.2868005728078\\
261	28.2228576346292\\
262	28.2451204266774\\
263	28.2563561829113\\
264	28.2674903565029\\
265	28.2774705680079\\
266	28.2850427157639\\
267	28.290146294446\\
268	28.2917914099816\\
269	28.2933797377884\\
270	28.2947833865497\\
271	28.2225247235164\\
272	28.2416952852913\\
273	28.2506509314832\\
274	28.2602689274394\\
275	28.2690406917372\\
276	28.2759179730995\\
277	28.2807508017037\\
278	28.2824783307271\\
279	28.284250582862\\
280	28.2858631285501\\
281	28.2202465574352\\
282	28.2424584876808\\
283	28.2534760100687\\
284	28.2647370947068\\
285	28.2749863703865\\
286	28.2827701641288\\
287	28.2878168719982\\
288	28.2892733575698\\
289	28.2906157724427\\
290	28.2917754170316\\
291	28.2196947496101\\
292	28.2408390589214\\
293	28.2513846047508\\
294	28.2614391660809\\
295	28.2704648500086\\
296	28.2773086667252\\
297	28.2819404365196\\
298	28.283350415624\\
299	28.2847202534776\\
300	28.2859300835399\\
};
\addplot [color=green, line width=0.75pt, forget plot]
  table[row sep=crcr]{%
1	25.1943157684673\\
2	25.8415353677514\\
3	27.4695496881603\\
4	27.9485925114509\\
5	28.1775812984333\\
6	28.3216852496641\\
7	28.3985395268925\\
8	28.4434363097709\\
9	28.4671980039588\\
10	28.5111332049424\\
11	28.4591482643217\\
12	28.5181043175114\\
13	28.5294170722328\\
14	28.5221181276267\\
15	28.5108364389918\\
16	28.5009586739287\\
17	28.493926483634\\
18	28.4900273053822\\
19	28.4900273053822\\
20	28.4905738616555\\
21	28.4331981343664\\
22	28.4639234791193\\
23	28.4640572321907\\
24	28.4535708152858\\
25	28.4438758624025\\
26	28.4341229837158\\
27	28.4264517902094\\
28	28.4213856381255\\
29	28.4188894775893\\
30	28.4172369185772\\
31	28.3443160696424\\
32	28.3939332041125\\
33	28.4077729453673\\
34	28.4080171648324\\
35	28.4039572687074\\
36	28.3993140524887\\
37	28.3951577777753\\
38	28.3921852678524\\
39	28.3902670544843\\
40	28.3891791040605\\
41	28.3429174528896\\
42	28.3776861383222\\
43	28.3846125922339\\
44	28.3824507016081\\
45	28.3782163278026\\
46	28.3739233089742\\
47	28.3707891106881\\
48	28.368603177887\\
49	28.3674303268089\\
50	28.3667194982729\\
51	28.3311859492786\\
52	28.3625248031875\\
53	28.3705887895075\\
54	28.3694675251141\\
55	28.3650928800053\\
56	28.3604868690772\\
57	28.3565112173943\\
58	28.353812637179\\
59	28.3521479437846\\
60	28.3511937466629\\
61	28.3270904059448\\
62	28.3560728253847\\
63	28.3648619407756\\
64	28.3654436559506\\
65	28.3631880082523\\
66	28.360680009331\\
67	28.3585474442459\\
68	28.3570939261463\\
69	28.3563245303324\\
70	28.3559294702953\\
71	28.330168126721\\
72	28.352242620143\\
73	28.357892038267\\
74	28.3570992805344\\
75	28.3541271221933\\
76	28.3509275720703\\
77	28.3484216761809\\
78	28.3464929054326\\
79	28.3453047393066\\
80	28.3446046347069\\
81	28.3229642323173\\
82	28.342808698651\\
83	28.3472780829627\\
84	28.3460293587223\\
85	28.3427911625023\\
86	28.3393464762309\\
87	28.3363019121078\\
88	28.3342037805264\\
89	28.3330099322378\\
90	28.332388613739\\
91	28.3176438546051\\
92	28.3381673830509\\
93	28.3430273756361\\
94	28.3417269253283\\
95	28.3387652603428\\
96	28.3364930633541\\
97	28.3329407382249\\
98	28.3310920854334\\
99	28.3292115861662\\
100	28.3280480056237\\
101	28.3311724599517\\
102	28.3423862734508\\
103	28.3421272127665\\
104	28.3384454548623\\
105	28.3343447788997\\
106	28.3309954587582\\
107	28.3282815778133\\
108	28.3265504580295\\
109	28.325462525611\\
110	28.3252846074704\\
111	28.3146690473935\\
112	28.3329569563233\\
113	28.336087541269\\
114	28.3333710094178\\
115	28.3293432988342\\
116	28.3261893144676\\
117	28.3230242142207\\
118	28.3207883919739\\
119	28.3192839024936\\
120	28.318370223136\\
121	28.3053780134182\\
122	28.3243941722894\\
123	28.3290970722634\\
124	28.3276960683317\\
125	28.3242376282295\\
126	28.3208269623663\\
127	28.3178419818517\\
128	28.3158843705887\\
129	28.3146276771113\\
130	28.3138998413759\\
131	28.3044749584392\\
132	28.3229389928108\\
133	28.3265362606561\\
134	28.3247218341765\\
135	28.3212331203501\\
136	28.3179773986746\\
137	28.3179773986746\\
138	28.3160668103857\\
139	28.3146348876501\\
140	28.3137871089119\\
141	28.3171827457034\\
142	28.3274911074217\\
143	28.3271964853533\\
144	28.3238275818925\\
145	28.3191232478711\\
146	28.3148566631354\\
147	28.311416305468\\
148	28.308854786293\\
149	28.3072621668604\\
150	28.3062291172844\\
151	28.3004152410709\\
152	28.3182341865265\\
153	28.322341493857\\
154	28.3213964797235\\
155	28.318597953734\\
156	28.3156184460789\\
157	28.3131769727693\\
158	28.3116766629973\\
159	28.3103091998494\\
160	28.3094649347945\\
161	28.3147185072798\\
162	28.3256356616224\\
163	28.3255538405451\\
164	28.322121892869\\
165	28.3181020224541\\
166	28.3146587454095\\
167	28.3119056908469\\
168	28.3102339949128\\
169	28.309166886004\\
170	28.3085082787766\\
171	28.3009786794503\\
172	28.3173053302161\\
173	28.3202199268743\\
174	28.318373421306\\
175	28.3152461738422\\
176	28.3123198503568\\
177	28.3098869985186\\
178	28.3081992017047\\
179	28.3070745717012\\
180	28.3064205464108\\
181	28.3022939089539\\
182	28.318775806927\\
183	28.3218883649717\\
184	28.3203474983478\\
185	28.3174349569054\\
186	28.3144851144554\\
187	28.3122696346007\\
188	28.3108332723014\\
189	28.3100951888473\\
190	28.3098451836694\\
191	28.3020251898706\\
192	28.3185626255376\\
193	28.3216306474655\\
194	28.319548452151\\
195	28.3159697422616\\
196	28.3123569187358\\
197	28.3095884443365\\
198	28.3076713194026\\
199	28.3064129607809\\
200	28.3055835141501\\
201	28.3131274637666\\
202	28.3245675442275\\
203	28.3253919641848\\
204	28.3227955582951\\
205	28.3193539521347\\
206	28.3161571556562\\
207	28.3137262235228\\
208	28.3121422732567\\
209	28.3112135879229\\
210	28.3107671700343\\
211	28.3007675641808\\
212	28.3163484157926\\
213	28.3193080194793\\
214	28.3175755627599\\
215	28.3146788145435\\
216	28.3116198179661\\
217	28.3092332630934\\
218	28.3075731547131\\
219	28.3064917136541\\
220	28.3064917136541\\
221	28.313544437356\\
222	28.3258338905464\\
223	28.3265378979291\\
224	28.322885592085\\
225	28.3179360259535\\
226	28.3137760367173\\
227	28.310388295332\\
228	28.3081642154617\\
229	28.3067734905935\\
230	28.3065211116549\\
231	28.2959744850755\\
232	28.3113657415947\\
233	28.314372151695\\
234	28.3129417652964\\
235	28.3100153153124\\
236	28.3072353736586\\
237	28.3049861421476\\
238	28.3032661179687\\
239	28.302189568413\\
240	28.3015581005815\\
241	28.2958553597156\\
242	28.3122980530266\\
243	28.3160213248556\\
244	28.3150970830307\\
245	28.3125118472556\\
246	28.3094673797812\\
247	28.3070807706347\\
248	28.3055976619553\\
249	28.304929988251\\
250	28.3040512825722\\
251	28.2973008740144\\
252	28.3119619945259\\
253	28.3137114667208\\
254	28.3111977842146\\
255	28.3075818813585\\
256	28.3043290551293\\
257	28.3018756011024\\
258	28.3003916164147\\
259	28.2994997100936\\
260	28.2990202285989\\
261	28.2948018858441\\
262	28.3131721468425\\
263	28.3184896581828\\
264	28.3186130960073\\
265	28.3168424926136\\
266	28.3146166146813\\
267	28.3127025715359\\
268	28.3112103674424\\
269	28.3101726938579\\
270	28.3094657204091\\
271	28.3009835697827\\
272	28.3144192809293\\
273	28.316404712388\\
274	28.3143754513811\\
275	28.3112417722538\\
276	28.3083366106668\\
277	28.3058887528207\\
278	28.3042670093122\\
279	28.3033643531612\\
280	28.3028976903757\\
281	28.2979174217211\\
282	28.3146068317151\\
283	28.3180489237759\\
284	28.3169680734111\\
285	28.314918091649\\
286	28.3127506907777\\
287	28.3107194898365\\
288	28.3093659109012\\
289	28.3084747605652\\
290	28.3078895973918\\
291	28.2987726930507\\
292	28.3159761737802\\
293	28.3191823457589\\
294	28.3171564519381\\
295	28.3138848346863\\
296	28.3108070694171\\
297	28.3085300901341\\
298	28.3068419937695\\
299	28.3058031755799\\
300	28.305141256306\\
};
\addplot [color=red, line width=0.75pt, forget plot]
  table[row sep=crcr]{%
1	24.5985882393961\\
2	24.6815049714483\\
3	24.7206668398813\\
4	24.7534380682505\\
5	24.782959721592\\
6	24.8098809737787\\
7	24.8344746460107\\
8	24.8572221808411\\
9	24.8785425472192\\
10	24.898652983094\\
11	25.1834759276299\\
12	25.2504495185084\\
13	25.2700265459434\\
14	25.2870360031897\\
15	25.3024686041865\\
16	25.3171060375549\\
17	25.3304799966161\\
18	25.3430635255427\\
19	25.3549752731955\\
20	25.3662830076316\\
21	25.5552051756541\\
22	25.6043267097152\\
23	25.6163604040472\\
24	25.6269661282505\\
25	25.6371199497923\\
26	25.6464206539435\\
27	25.6554335032973\\
28	25.6638871289796\\
29	25.6720984276293\\
30	25.6800035231294\\
31	25.8090354915333\\
32	25.845757489708\\
33	25.8544080705479\\
34	25.862094704371\\
35	25.8695544958212\\
36	25.8763225611527\\
37	25.8830885287592\\
38	25.8893118032091\\
39	25.8956026843887\\
40	25.9016120094602\\
41	25.9913267261676\\
42	26.0186179189669\\
43	26.0252499499984\\
44	26.0311823002891\\
45	26.0368629251072\\
46	26.0422335822184\\
47	26.0474917121499\\
48	26.0524684600298\\
49	26.0573796018428\\
50	26.0620537133729\\
51	26.1131500815039\\
52	26.1219269544299\\
53	26.1219269544299\\
54	26.1592701115029\\
55	26.188653636567\\
56	26.2007906984184\\
57	26.2056133148192\\
58	26.2104355220096\\
59	26.2150468841625\\
60	26.2150468841625\\
61	26.2554161045254\\
62	26.2766899812583\\
63	26.288154928355\\
64	26.292366409449\\
65	26.2962793709747\\
66	26.3002763132418\\
67	26.3039024004881\\
68	26.3076131062591\\
69	26.3110175704379\\
70	26.3143398981024\\
71	26.3542550604777\\
72	26.3690498947961\\
73	26.3733151881846\\
74	26.3767061034736\\
75	26.3767061034736\\
76	26.3767061034736\\
77	26.3767061034736\\
78	26.4044254885112\\
79	26.4206677010183\\
80	26.4316776060581\\
81	26.4677684784711\\
82	26.4803023476918\\
83	26.4835769237309\\
84	26.4863817063089\\
85	26.4863817063089\\
86	26.4863817063089\\
87	26.4897785740237\\
88	26.4920989348627\\
89	26.4920989348627\\
90	26.4951258713117\\
91	26.5149460961132\\
92	26.5286779933241\\
93	26.5365115443416\\
94	26.5398363281328\\
95	26.5423630979933\\
96	26.5423630979933\\
97	26.5423630979933\\
98	26.5423630979933\\
99	26.5423630979933\\
100	26.5423630979933\\
101	26.5647865604124\\
102	26.5739616401667\\
103	26.5772718463743\\
104	26.5795585872618\\
105	26.5795585872618\\
106	26.5795585872618\\
107	26.5795585872618\\
108	26.5826585208812\\
109	26.5850507937528\\
110	26.5850507937528\\
111	26.6046069677253\\
112	26.6133362628294\\
113	26.6165463672174\\
114	26.6185555829181\\
115	26.6185555829181\\
116	26.6185555829181\\
117	26.6212950607428\\
118	26.6234840506071\\
119	26.6234840506071\\
120	26.6234840506071\\
121	26.6404537303539\\
122	26.6558477171171\\
123	26.6723578270888\\
124	26.679249362936\\
125	26.6880140611223\\
126	26.6946856900062\\
127	26.6968386371351\\
128	26.6990673099611\\
129	26.701240726654\\
130	26.7034059632448\\
131	26.7211244851383\\
132	26.7287440147723\\
133	26.7311854912952\\
134	26.7331458810941\\
135	26.7331458810941\\
136	26.7331458810941\\
137	26.7331458810941\\
138	26.7331458810941\\
139	26.7331458810941\\
140	26.7331458810941\\
141	26.7430248024504\\
142	26.7503527573771\\
143	26.7527282058309\\
144	26.7548562508831\\
145	26.7548562508831\\
146	26.7548562508831\\
147	26.7548562508831\\
148	26.7548562508831\\
149	26.7548562508831\\
150	26.7548562508831\\
151	26.7676978171091\\
152	26.7807933899276\\
153	26.7981496351424\\
154	26.8100292876853\\
155	26.8191860566151\\
156	26.821475115149\\
157	26.821475115149\\
158	26.8240907538694\\
159	26.8260443602421\\
160	26.8260443602421\\
161	26.8399653715188\\
162	26.8463982518678\\
163	26.8488518540494\\
164	26.8504447687796\\
165	26.8504447687796\\
166	26.8504447687796\\
167	26.8526969136148\\
168	26.8543208673178\\
169	26.8543208673178\\
170	26.8564303189073\\
171	26.8690170152143\\
172	26.8746498330698\\
173	26.8767542898727\\
174	26.8782627465622\\
175	26.8782627465622\\
176	26.8782627465622\\
177	26.8804659914097\\
178	26.8818591772086\\
179	26.8818591772086\\
180	26.883325983929\\
181	26.8951254740021\\
182	26.9065907842288\\
183	26.908342902158\\
184	26.9101233348941\\
185	26.9101233348941\\
186	26.9101233348941\\
187	26.9101233348941\\
188	26.9101233348941\\
189	26.9101233348941\\
190	26.9101233348941\\
191	26.9169696072362\\
192	26.9226731783346\\
193	26.9247687216504\\
194	26.9264449224627\\
195	26.9490012545176\\
196	26.9598440060428\\
197	26.9771952382411\\
198	26.987647500298\\
199	26.9893334103127\\
200	26.9911189945361\\
201	26.9985880546088\\
202	27.0002398882036\\
203	27.0002398882036\\
204	27.0107229163031\\
205	27.0233826597091\\
206	27.024672521499\\
207	27.041288449085\\
208	27.0479050139026\\
209	27.0570356866966\\
210	27.058294264779\\
211	27.0688337397798\\
212	27.0749319518534\\
213	27.0795729292269\\
214	27.0795729292269\\
215	27.0814937493885\\
216	27.0829877037148\\
217	27.0829877037148\\
218	27.0829877037148\\
219	27.0829877037148\\
220	27.0829877037148\\
221	27.0888817670634\\
222	27.0994321448855\\
223	27.1048428828782\\
224	27.1064484728485\\
225	27.1081991033464\\
226	27.1096804103866\\
227	27.1096804103866\\
228	27.1096804103866\\
229	27.1096804103866\\
230	27.1096804103866\\
231	27.1154507244302\\
232	27.1203943956476\\
233	27.1219644249367\\
234	27.123356569995\\
235	27.123356569995\\
236	27.123356569995\\
237	27.123356569995\\
238	27.1396988028504\\
239	27.1484722713032\\
240	27.1642582036546\\
241	27.1735873105634\\
242	27.1835001337222\\
243	27.1882700201691\\
244	27.1892553193135\\
245	27.1892553193135\\
246	27.1892553193135\\
247	27.1908637001142\\
248	27.1920872363785\\
249	27.1920872363785\\
250	27.1920872363785\\
251	27.200799411814\\
252	27.2093483007613\\
253	27.2151563394523\\
254	27.2196890616191\\
255	27.2196890616191\\
256	27.2196890616191\\
257	27.2254922731568\\
258	27.226706727087\\
259	27.226706727087\\
260	27.2283303523542\\
261	27.236580305674\\
262	27.2409006327734\\
263	27.2424137637577\\
264	27.2434909417459\\
265	27.2434909417459\\
266	27.2434909417459\\
267	27.2448905911985\\
268	27.246249839776\\
269	27.246249839776\\
270	27.2474700861469\\
271	27.2555634912065\\
272	27.2637821638057\\
273	27.2654189472634\\
274	27.2664712686977\\
275	27.2664712686977\\
276	27.2664712686977\\
277	27.2679972693749\\
278	27.2692285630463\\
279	27.2692285630463\\
280	27.2706988015457\\
281	27.2782495703697\\
282	27.2860395143226\\
283	27.2872868442008\\
284	27.2872868442008\\
285	27.2872868442008\\
286	27.2872868442008\\
287	27.3024173092251\\
288	27.3062250717767\\
289	27.3123092615512\\
290	27.315909808548\\
291	27.3237626506864\\
292	27.3323196553446\\
293	27.3337446749833\\
294	27.3348250526975\\
295	27.3348250526975\\
296	27.3348250526975\\
297	27.3362070191823\\
298	27.3373336214757\\
299	27.3373336214757\\
300	27.3373336214757\\
};
\end{axis}
\end{tikzpicture}%

%% file: images/conv_SSIM_alpha.tikz
%
%
\begin{tikzpicture}

\begin{axis}[%
width=0.951\figurewidth,
height=\figureheight,
at={(0\figurewidth,0\figureheight)},
scale only axis,
unbounded coords=jump,
xmin=0,
xmax=300,
ymin=0.4,
ymax=0.8,
ylabel style={font=\color{white!15!black}},
ylabel={SSIM},
xlabel style={font=\color{white!15!black}},
xlabel={Number of inner and outer iterations},
axis background/.style={fill=white},
xmajorgrids,
ymajorgrids,
xlabel near ticks,
ylabel near ticks
]
\addplot [color=black, line width=0.8pt, forget plot]
  table[row sep=crcr]{%
1	0.447132792013231\\
2	0.470467194213733\\
3	0.490248031609507\\
4	0.535027981419754\\
5	0.566336266612457\\
6	0.570217827611425\\
7	0.570217827611425\\
8	0.579449268183018\\
9	0.587044608658933\\
10	0.587044608658933\\
11	0.610205495867125\\
12	0.621871728540799\\
13	0.630029587856091\\
14	0.652470300538481\\
15	0.67119704611202\\
16	0.677109027309769\\
17	0.679412838884696\\
18	0.679412838884696\\
19	0.679412838884696\\
20	0.679412838884696\\
21	0.695190440975232\\
22	0.707886748925659\\
23	0.709532172568961\\
24	0.718822792992596\\
25	0.724244099915571\\
26	0.724244099915571\\
27	0.733530213638752\\
28	0.734254116315106\\
29	0.735295343292766\\
30	0.735295343292766\\
31	0.744499819483657\\
32	0.750068173308777\\
33	0.750606704841783\\
34	0.750606704841783\\
35	0.753270707093871\\
36	0.755045220689387\\
37	0.758318634642998\\
38	0.761078297524092\\
39	0.761890147205961\\
40	0.762336065569206\\
41	0.767979039843398\\
42	0.770797538547217\\
43	0.771061883050198\\
44	0.771061883050198\\
45	0.771813175976829\\
46	0.772618267266076\\
47	0.773047311737532\\
48	0.773548545062137\\
49	0.773548545062137\\
50	0.773548545062137\\
51	0.778745053914404\\
52	0.779961957373213\\
53	0.780155912620025\\
54	0.780155912620025\\
55	0.779119491245776\\
56	0.779893562981844\\
57	0.781828011917944\\
58	0.782503084530719\\
59	0.782667379285663\\
60	0.782667379285663\\
61	0.786209982608074\\
62	0.787084616150734\\
63	0.787124805524764\\
64	0.78714930305371\\
65	0.78714930305371\\
66	0.78714930305371\\
67	0.78714930305371\\
68	0.787404545661127\\
69	0.787536038366196\\
70	0.787536038366196\\
71	0.789122534331431\\
72	0.78996323285451\\
73	0.790074572622523\\
74	0.790074572622523\\
75	0.788302156356146\\
76	0.78858553949244\\
77	0.788709339657087\\
78	0.788709339657087\\
79	0.788709339657087\\
80	0.788709339657087\\
81	0.791477689874253\\
82	0.792054517362402\\
83	0.792091767837102\\
84	0.792091767837102\\
85	0.791207576068801\\
86	0.791199097359645\\
87	0.791265760902294\\
88	0.79130962052825\\
89	0.79130962052825\\
90	0.791753698345446\\
91	0.793702654481453\\
92	0.794056475481657\\
93	0.794052198589591\\
94	0.794052198589591\\
95	0.791895754415146\\
96	0.79183289067391\\
97	0.791908662793226\\
98	0.79184748518783\\
99	0.79184748518783\\
100	0.79184748518783\\
101	0.794353318671433\\
102	0.7942679172279\\
103	0.794054708362647\\
104	0.794004713772065\\
105	0.793997835715867\\
106	0.793997835715867\\
107	0.794039122276083\\
108	0.794039122276083\\
109	0.794039122276083\\
110	0.794120532740398\\
111	0.794984920623837\\
112	0.794935201030131\\
113	0.794894645159267\\
114	0.794894645159267\\
115	0.793258798278337\\
116	0.793187611865869\\
117	0.793249291279803\\
118	0.793194766805497\\
119	0.793144465491107\\
120	0.793143417506321\\
121	0.794836094162868\\
122	0.794779440272492\\
123	0.794589480181835\\
124	0.794529284792092\\
125	0.794496246304627\\
126	0.794496246304627\\
127	0.794496246304627\\
128	0.794496246304627\\
129	0.794496246304627\\
130	0.794496246304627\\
131	0.79525551178117\\
132	0.795310231142414\\
133	0.795284847799119\\
134	0.795284847799119\\
135	0.794112924887263\\
136	0.794003456890253\\
137	0.794003456890253\\
138	0.793995203053402\\
139	0.793942677599529\\
140	0.793891182406847\\
141	0.79499948310758\\
142	0.795101243023207\\
143	0.795067891761545\\
144	0.795067891761545\\
145	0.793160931796938\\
146	0.792832229407174\\
147	0.792481648933633\\
148	0.79215360998837\\
149	0.792133376255702\\
150	0.792133376255702\\
151	0.793933637253725\\
152	0.793770002770642\\
153	0.793429509347751\\
154	0.793383450689017\\
155	0.793372887741453\\
156	0.793372887741453\\
157	0.793372887741453\\
158	0.793429134417145\\
159	0.793429134417145\\
160	0.793486231272121\\
161	0.794324349254487\\
162	0.794387345237614\\
163	0.794354569841142\\
164	0.792672579891897\\
165	0.792357806213331\\
166	0.792073559012295\\
167	0.791884968039795\\
168	0.791875968585127\\
169	0.791875968585127\\
170	0.791875968585127\\
171	0.79351650702484\\
172	0.793349768359026\\
173	0.793113533086732\\
174	0.793054990364656\\
175	0.79302956621935\\
176	0.79302956621935\\
177	0.79302956621935\\
178	0.793049652020435\\
179	0.793047932783218\\
180	0.793047932783218\\
181	0.793780491644019\\
182	0.793866778001666\\
183	0.793846134940115\\
184	0.793846134940115\\
185	0.792250247603716\\
186	0.792011333875157\\
187	0.792039201825789\\
188	0.792039201825789\\
189	0.792039201825789\\
190	0.792089491095986\\
191	0.793056347649183\\
192	0.792909012093651\\
193	0.792678790273154\\
194	0.792616165331443\\
195	0.792616165331443\\
196	0.792595465660655\\
197	0.792577273252588\\
198	0.792557956743862\\
199	0.792542141615398\\
200	0.792542141615398\\
201	0.793198239437432\\
202	0.792975048791918\\
203	0.792535196568275\\
204	0.79249356365333\\
205	0.79249356365333\\
206	0.79249356365333\\
207	0.792556937680042\\
208	0.792563355791159\\
209	0.792542112612998\\
210	0.79252023354776\\
211	0.792960926919698\\
212	0.793019543063134\\
213	0.792993239645123\\
214	0.791717885109042\\
215	0.791325378951222\\
216	0.790937699039247\\
217	0.790801497560456\\
218	0.790791139706209\\
219	0.790792026155445\\
220	0.790804817011009\\
221	0.792092772716621\\
222	0.791968926368928\\
223	0.791802149550463\\
224	0.791757237817072\\
225	0.791733470491548\\
226	0.791733470491548\\
227	0.791733470491548\\
228	0.790657568985387\\
229	0.790597224724551\\
230	0.790547329883946\\
231	0.791777576424568\\
232	0.791536765571781\\
233	0.791159550389917\\
234	0.791124239009408\\
235	0.791124239009408\\
236	0.790956846291393\\
237	0.790871471183391\\
238	0.790747369440493\\
239	0.790731160320907\\
240	0.7907161348744\\
241	0.791655525916124\\
242	0.79135221689913\\
243	0.791309093118666\\
244	0.791309093118666\\
245	0.79127693699061\\
246	0.791252397026468\\
247	0.791252397026468\\
248	0.791252397026468\\
249	0.791238626416618\\
250	0.791220012828491\\
251	0.791730011392013\\
252	0.791742829174857\\
253	0.791706854135392\\
254	0.791706854135392\\
255	0.791706854135392\\
256	0.791706854135392\\
257	0.789947370850091\\
258	0.789868411383993\\
259	0.789893065039324\\
260	0.789701901844259\\
261	0.79106005492991\\
262	0.79107553472209\\
263	0.791059356303777\\
264	0.791059356303777\\
265	0.791059356303777\\
266	0.791059356303777\\
267	0.791071165861906\\
268	0.791060523667248\\
269	0.791060523667248\\
270	0.791049568501909\\
271	0.791049182101187\\
272	0.791017535442623\\
273	0.790988540431967\\
274	0.790004661856935\\
275	0.789856118114032\\
276	0.789769060990227\\
277	0.789675669295487\\
278	0.7896530947487\\
279	0.789651418404465\\
280	0.789651418404465\\
281	0.790773107832284\\
282	0.790713566200282\\
283	0.790689446731163\\
284	0.790689446731163\\
285	0.790689446731163\\
286	0.790689446731163\\
287	0.790689446731163\\
288	0.789311856035371\\
289	0.789384114971487\\
290	0.789499407393605\\
291	0.790881039388471\\
292	0.790858405546271\\
293	0.790713470501543\\
294	0.790713470501543\\
295	0.790580567848491\\
296	0.790539987859783\\
297	0.790526877732289\\
298	0.790526877732289\\
299	0.790526877732289\\
300	0.790541475997178\\
};
\addplot [color=blue, line width=0.8pt, forget plot]
  table[row sep=crcr]{%
1	0.640324029531773\\
2	0.767459583578617\\
3	0.790246515979081\\
4	0.788208257554328\\
5	0.78555783535934\\
6	0.784598463064272\\
7	0.78496348204026\\
8	0.785807457254821\\
9	0.786762624862023\\
10	0.787675464263963\\
11	0.775861920027047\\
12	0.780457375479376\\
13	0.782311301270334\\
14	0.783621391303964\\
15	0.784609298474996\\
16	0.785347212151894\\
17	0.785902017950485\\
18	0.786337018991527\\
19	0.786697477110989\\
20	0.78682856486854\\
21	0.778981187575945\\
22	0.782057373639045\\
23	0.783591317898935\\
24	0.784681906422155\\
25	0.785504849670281\\
26	0.786069532871777\\
27	0.786426847561047\\
28	0.786650295540781\\
29	0.786696371065789\\
30	0.786777435471888\\
31	0.778709568114476\\
32	0.781430016562191\\
33	0.782850220328371\\
34	0.78397654094253\\
35	0.784848499024663\\
36	0.785441250978848\\
37	0.785798456369164\\
38	0.786000008615269\\
39	0.786028561854466\\
40	0.786077296564715\\
41	0.778387286677928\\
42	0.781149458491074\\
43	0.782576214636462\\
44	0.783706218948256\\
45	0.784580493940099\\
46	0.78519795334107\\
47	0.785585018182222\\
48	0.785818541960276\\
49	0.785875731976298\\
50	0.78594258854914\\
51	0.777711873823133\\
52	0.780613305401895\\
53	0.782130806235134\\
54	0.78334546798066\\
55	0.78429047758015\\
56	0.78493011125252\\
57	0.785321165801814\\
58	0.785542442140247\\
59	0.785583463805533\\
60	0.785630827628717\\
61	0.777226950908729\\
62	0.780224646861543\\
63	0.781740453796485\\
64	0.782982247562027\\
65	0.783962043314409\\
66	0.784641786609127\\
67	0.785069473906995\\
68	0.785314441180461\\
69	0.785365187517554\\
70	0.785419323866902\\
71	0.777227574765466\\
72	0.780175762794293\\
73	0.781600073309678\\
74	0.78273423858142\\
75	0.783602264833821\\
76	0.784216629316164\\
77	0.784618640278543\\
78	0.784873039211654\\
79	0.78495629483524\\
80	0.785046504030325\\
81	0.777122500625646\\
82	0.780404048701816\\
83	0.781977253653784\\
84	0.783172311080962\\
85	0.784062505545378\\
86	0.784664724722985\\
87	0.785039703757019\\
88	0.785261047812118\\
89	0.785313335055646\\
90	0.785370092566437\\
91	0.777534320226632\\
92	0.780669424263118\\
93	0.782139257869855\\
94	0.783266816033791\\
95	0.784130834706911\\
96	0.784743927330333\\
97	0.785140569304378\\
98	0.785266357941486\\
99	0.785491485488747\\
100	0.785540407114152\\
101	0.777785985753829\\
102	0.780601314201966\\
103	0.782043505688713\\
104	0.783222881371855\\
105	0.78415546213055\\
106	0.784819763568696\\
107	0.785262761411373\\
108	0.785416023842192\\
109	0.785678021927961\\
110	0.785743680123754\\
111	0.777719540946463\\
112	0.780492989217817\\
113	0.781887808449934\\
114	0.782997990986957\\
115	0.783874087888773\\
116	0.784502512308025\\
117	0.784907122854977\\
118	0.785048224585131\\
119	0.785296624172211\\
120	0.785358949936843\\
121	0.777994970690374\\
122	0.780966148901946\\
123	0.782469317742211\\
124	0.783638404507035\\
125	0.784520462427065\\
126	0.785118503653521\\
127	0.785494450018105\\
128	0.785610377257301\\
129	0.785824564185618\\
130	0.785871691129091\\
131	0.777675678306904\\
132	0.780566847859619\\
133	0.781946621976352\\
134	0.782988429300493\\
135	0.783805095096486\\
136	0.784391268990217\\
137	0.784773740377457\\
138	0.784899597633296\\
139	0.785125130046933\\
140	0.785175285317516\\
141	0.777245773502194\\
142	0.780100486687369\\
143	0.781471334532864\\
144	0.782544828714152\\
145	0.783409286903215\\
146	0.784045469411636\\
147	0.784471946148054\\
148	0.78462012987349\\
149	0.784756759603809\\
150	0.784875699035811\\
151	0.777313203430905\\
152	0.780111277585876\\
153	0.781510263180337\\
154	0.78262605421546\\
155	0.783518899438019\\
156	0.784166718623965\\
157	0.784600007547747\\
158	0.784754039252679\\
159	0.784894968291665\\
160	0.785012333762754\\
161	0.77783889124906\\
162	0.780749085285979\\
163	0.782131346349841\\
164	0.783202718770252\\
165	0.784032680823511\\
166	0.784630311303737\\
167	0.785022670830651\\
168	0.785157012256265\\
169	0.785276851747111\\
170	0.785378280505357\\
171	0.778026793268631\\
172	0.780828653936635\\
173	0.782166630593389\\
174	0.783230789907041\\
175	0.784072986587981\\
176	0.784683436841655\\
177	0.785078203705748\\
178	0.785213209962342\\
179	0.785333904430861\\
180	0.785437099764308\\
181	0.778413581070979\\
182	0.781058605275503\\
183	0.782333401157428\\
184	0.783352780713045\\
185	0.784169499953643\\
186	0.784756095496263\\
187	0.785135982979604\\
188	0.785264441331486\\
189	0.785383571538856\\
190	0.785490703578415\\
191	0.777945981844797\\
192	0.780657298543785\\
193	0.782032574161886\\
194	0.783089173091385\\
195	0.783885502333993\\
196	0.784436611149127\\
197	0.784786472499797\\
198	0.784897185289376\\
199	0.784994826094379\\
200	0.785076605508538\\
201	0.778035281991901\\
202	0.780675808892287\\
203	0.781987381438921\\
204	0.783074035455204\\
205	0.783945750160964\\
206	0.784575317510824\\
207	0.784991362439695\\
208	0.785139990247594\\
209	0.785274577478486\\
210	0.785387732567163\\
211	0.778029388791689\\
212	0.780646364830376\\
213	0.781929757669287\\
214	0.782939340636195\\
215	0.783738352027918\\
216	0.784313301988246\\
217	0.784693574352257\\
218	0.784820742444683\\
219	0.784937303666046\\
220	0.785034857272348\\
221	0.777846424016543\\
222	0.780683146747084\\
223	0.782012312562164\\
224	0.783048801230789\\
225	0.783817430608015\\
226	0.784348465745986\\
227	0.784698031904129\\
228	0.784811897050552\\
229	0.784917949850383\\
230	0.785014340570778\\
231	0.778135325036292\\
232	0.780778149197784\\
233	0.782069841902264\\
234	0.783076981286718\\
235	0.78386337982509\\
236	0.784437653675516\\
237	0.784819069976983\\
238	0.78495166991528\\
239	0.785080894660394\\
240	0.785195115144916\\
241	0.77834979758171\\
242	0.780950639494716\\
243	0.782210495991079\\
244	0.783231246316012\\
245	0.784030691914281\\
246	0.784602166028189\\
247	0.784983310355678\\
248	0.785122213290686\\
249	0.785254219043659\\
250	0.785365784083295\\
251	0.778501591820672\\
252	0.781047582909898\\
253	0.782274527415376\\
254	0.783257753422394\\
255	0.784023114958494\\
256	0.784556453923145\\
257	0.78491101201861\\
258	0.785043210690314\\
259	0.785172409816668\\
260	0.785287913579187\\
261	0.778741566661502\\
262	0.781333404288426\\
263	0.78266450868206\\
264	0.783716415588961\\
265	0.784540441536938\\
266	0.785115728199996\\
267	0.785486258212735\\
268	0.785601984608733\\
269	0.785712199355662\\
270	0.785807542723527\\
271	0.778570491734527\\
272	0.780971927926241\\
273	0.782172810286324\\
274	0.783139025429621\\
275	0.783876101036483\\
276	0.784415042064035\\
277	0.784788508885582\\
278	0.784932579148922\\
279	0.785077120632011\\
280	0.785204746179718\\
281	0.778399902086337\\
282	0.781010293465852\\
283	0.782339244136299\\
284	0.783415966729105\\
285	0.78426106112533\\
286	0.784857641297501\\
287	0.785236450562887\\
288	0.785355982576997\\
289	0.78546992686458\\
290	0.78557159313428\\
291	0.7782050261531\\
292	0.780899557235033\\
293	0.782242845298717\\
294	0.783247710904935\\
295	0.783992650856756\\
296	0.784494221593334\\
297	0.784813029304991\\
298	0.784912637089666\\
299	0.785009190133362\\
300	0.785096026032934\\
};
\addplot [color=green, line width=0.8pt, forget plot]
  table[row sep=crcr]{%
1	0.488767721431062\\
2	0.548614481460239\\
3	0.701637179636719\\
4	0.744197290726305\\
5	0.766586953128166\\
6	0.779731128170693\\
7	0.786217418067848\\
8	0.789578316565815\\
9	0.791106800195398\\
10	0.793303921068688\\
11	0.784165666518453\\
12	0.793333101928389\\
13	0.795600462792381\\
14	0.795543105386687\\
15	0.795072365803477\\
16	0.794610664029967\\
17	0.794314333309993\\
18	0.794153545718609\\
19	0.794153545718609\\
20	0.794236875197893\\
21	0.785917185890487\\
22	0.791549264160792\\
23	0.792675283697013\\
24	0.792439341893268\\
25	0.792087789920671\\
26	0.791846262156767\\
27	0.791665685941045\\
28	0.79158801932491\\
29	0.791567395761135\\
30	0.791608788530472\\
31	0.781767672502123\\
32	0.788369318016659\\
33	0.789919364370335\\
34	0.790133540322562\\
35	0.790079606842038\\
36	0.790028504494309\\
37	0.789998809966593\\
38	0.789997290959928\\
39	0.79002539281322\\
40	0.790062126309915\\
41	0.783590371167624\\
42	0.7880542554002\\
43	0.788940579005204\\
44	0.789009292777506\\
45	0.788976503976484\\
46	0.788942317421381\\
47	0.788927039187951\\
48	0.788937633725062\\
49	0.788948431368089\\
50	0.788981564683607\\
51	0.783664119690044\\
52	0.787275526152548\\
53	0.788081036064732\\
54	0.788149813598081\\
55	0.788065074227307\\
56	0.787976344604508\\
57	0.787921834902908\\
58	0.787903195993731\\
59	0.787920068443336\\
60	0.787945238975454\\
61	0.784148951499559\\
62	0.787173169789091\\
63	0.78783758179194\\
64	0.787923836284283\\
65	0.787932261839011\\
66	0.787959216768618\\
67	0.788011334289556\\
68	0.788066472336669\\
69	0.788114089026796\\
70	0.788158380995482\\
71	0.784787084971144\\
72	0.787271108791114\\
73	0.787880108328937\\
74	0.788002961885102\\
75	0.78801167243216\\
76	0.787984527246981\\
77	0.787967578015299\\
78	0.78796378927013\\
79	0.787962467280782\\
80	0.787972497603727\\
81	0.784774923061825\\
82	0.786846855469135\\
83	0.787171538432731\\
84	0.78711276841595\\
85	0.7870234073098\\
86	0.786949630660464\\
87	0.78690952159586\\
88	0.786900241489712\\
89	0.786912323783366\\
90	0.78693433056805\\
91	0.784698160347775\\
92	0.78678256726266\\
93	0.787200845547336\\
94	0.787233317604941\\
95	0.787178563897237\\
96	0.787153459870315\\
97	0.787134067547085\\
98	0.787088020116006\\
99	0.787062639722857\\
100	0.787041471052506\\
101	0.786354732522525\\
102	0.787258337370966\\
103	0.787193852715573\\
104	0.787038118313918\\
105	0.786928545330642\\
106	0.786860847254673\\
107	0.786825463125347\\
108	0.786813505402378\\
109	0.786824690943497\\
110	0.78682350908619\\
111	0.784951447248796\\
112	0.786710570040338\\
113	0.786950362768516\\
114	0.7868646997044\\
115	0.786758242244145\\
116	0.786680891357703\\
117	0.786637984033851\\
118	0.786607658977228\\
119	0.786599987967467\\
120	0.786599077806796\\
121	0.784432895769806\\
122	0.786163517589435\\
123	0.786422963126151\\
124	0.786364917061548\\
125	0.786283211980917\\
126	0.786221269403038\\
127	0.786188766191523\\
128	0.786180791947114\\
129	0.786197788795968\\
130	0.786221433442447\\
131	0.784495371824454\\
132	0.786222725998731\\
133	0.786445477775481\\
134	0.786334713109772\\
135	0.786206089289807\\
136	0.786113117976102\\
137	0.786113117976102\\
138	0.786210781412195\\
139	0.786229901174093\\
140	0.786233569868413\\
141	0.785957610403018\\
142	0.7867420774979\\
143	0.78656441275607\\
144	0.786324612720345\\
145	0.786129274781542\\
146	0.785990698109796\\
147	0.785916481527907\\
148	0.785876207741365\\
149	0.785862990642104\\
150	0.785863791398015\\
151	0.784475547295738\\
152	0.786092343332351\\
153	0.786347901939102\\
154	0.786308371588253\\
155	0.786218915802595\\
156	0.786141395612364\\
157	0.786096265259153\\
158	0.78607616378724\\
159	0.786083282504409\\
160	0.786093572749617\\
161	0.785930811518812\\
162	0.786676224751129\\
163	0.786529109062622\\
164	0.786331386386146\\
165	0.786199784325224\\
166	0.786121869123676\\
167	0.786085418036372\\
168	0.786078757145172\\
169	0.786087363090836\\
170	0.786101751444676\\
171	0.784559354985666\\
172	0.785979320403313\\
173	0.786182517875051\\
174	0.786160690916156\\
175	0.78611304461764\\
176	0.786065620988896\\
177	0.786036702105454\\
178	0.786017683059786\\
179	0.786011521723205\\
180	0.786014603352954\\
181	0.784778948910462\\
182	0.786185375125574\\
183	0.786357103242902\\
184	0.786325097048051\\
185	0.786295671707706\\
186	0.78627894464243\\
187	0.786283822940283\\
188	0.786302575042941\\
189	0.786328659068982\\
190	0.786368049456584\\
191	0.78490402878307\\
192	0.786343104036595\\
193	0.786488251526853\\
194	0.786375186823549\\
195	0.786248720033896\\
196	0.786150293502967\\
197	0.786093413638702\\
198	0.786072058600794\\
199	0.786070470671992\\
200	0.786072753507274\\
201	0.786035714950915\\
202	0.786681443107154\\
203	0.786504017851159\\
204	0.786288248195848\\
205	0.786144586955149\\
206	0.786073843452521\\
207	0.786035643618268\\
208	0.786037080284927\\
209	0.786042670137289\\
210	0.786059724801581\\
211	0.78459117313348\\
212	0.785996049684571\\
213	0.786174074277885\\
214	0.786087459825276\\
215	0.785976955327494\\
216	0.785882755947254\\
217	0.785819898711998\\
218	0.785786633929532\\
219	0.785775898394366\\
220	0.785775898394366\\
221	0.785849597792365\\
222	0.786724355314235\\
223	0.78666426210913\\
224	0.786460012852035\\
225	0.786298785568973\\
226	0.786191520986369\\
227	0.786126411929394\\
228	0.786092659124744\\
229	0.786091270430875\\
230	0.786089316244913\\
231	0.784524780956204\\
232	0.785890581476353\\
233	0.786058724418045\\
234	0.786016288578645\\
235	0.785966488388206\\
236	0.785929653761512\\
237	0.785908723699955\\
238	0.785899342106306\\
239	0.785896856587789\\
240	0.785900849068741\\
241	0.78451417762895\\
242	0.785890710906371\\
243	0.786033584843603\\
244	0.785967062612368\\
245	0.785891174920338\\
246	0.785816106847251\\
247	0.785777752496549\\
248	0.785756436727556\\
249	0.785759363104065\\
250	0.785780024078492\\
251	0.784416676816197\\
252	0.785702951540923\\
253	0.785798589518668\\
254	0.785692943598697\\
255	0.785580488384519\\
256	0.785497898069795\\
257	0.785451541969326\\
258	0.785438840246582\\
259	0.785446801480894\\
260	0.785458924066746\\
261	0.784226503687972\\
262	0.785803598209052\\
263	0.786129385986784\\
264	0.786169684624189\\
265	0.786143060406522\\
266	0.786108537790861\\
267	0.786081844231691\\
268	0.786070274181108\\
269	0.786072909906439\\
270	0.786088567530822\\
271	0.784546783000271\\
272	0.785766342070316\\
273	0.785831155453879\\
274	0.785715654378702\\
275	0.785608925274067\\
276	0.785541372918841\\
277	0.785510186807736\\
278	0.78550497109895\\
279	0.785523779041643\\
280	0.785549959258028\\
281	0.784367865672619\\
282	0.785886267787099\\
283	0.786136334078135\\
284	0.786116571180079\\
285	0.786075382444142\\
286	0.786074780893311\\
287	0.78607731322669\\
288	0.786087064130157\\
289	0.786117753052016\\
290	0.78615021224578\\
291	0.784382016005019\\
292	0.78581168019179\\
293	0.785940947418514\\
294	0.785829709292314\\
295	0.785742018730199\\
296	0.78569072908635\\
297	0.785673835272327\\
298	0.785678165034759\\
299	0.785688125008956\\
300	0.785703058054591\\
};
\addplot [color=red, line width=0.8pt, forget plot]
  table[row sep=crcr]{%
1	0.442358041588876\\
2	0.450776393427798\\
3	0.454605954816981\\
4	0.457691038692024\\
5	0.460423310342987\\
6	0.462887896939118\\
7	0.465126005604617\\
8	0.467195988001033\\
9	0.469135283594323\\
10	0.470954914970991\\
11	0.497499486282741\\
12	0.503711260161763\\
13	0.505477276115817\\
14	0.506993753786094\\
15	0.508337654886608\\
16	0.509586870254338\\
17	0.510719005087894\\
18	0.511781476816351\\
19	0.512785749770476\\
20	0.513737367421607\\
21	0.532698488721394\\
22	0.537131928039273\\
23	0.538151012360449\\
24	0.539047059698654\\
25	0.539882375802846\\
26	0.540639840746913\\
27	0.541375514181784\\
28	0.542074339399433\\
29	0.542760114718177\\
30	0.543425478387633\\
31	0.55635507382283\\
32	0.55960660284774\\
33	0.560341596534323\\
34	0.560993841995275\\
35	0.561612666642015\\
36	0.562168865458528\\
37	0.562725425554445\\
38	0.56324214731511\\
39	0.563768655302134\\
40	0.564276552966229\\
41	0.573314553794911\\
42	0.575675057430374\\
43	0.576231812572236\\
44	0.576745801071384\\
45	0.57722887597872\\
46	0.577682905447916\\
47	0.57813007830881\\
48	0.578557671063112\\
49	0.578984361761309\\
50	0.579395023069828\\
51	0.584750610814616\\
52	0.58560145044685\\
53	0.58560145044685\\
54	0.588350578897073\\
55	0.5909067687679\\
56	0.592104735990653\\
57	0.592606115380665\\
58	0.593126685789451\\
59	0.59362882501869\\
60	0.59362882501869\\
61	0.597898233340102\\
62	0.599800654168129\\
63	0.600799146424754\\
64	0.601178538435545\\
65	0.601541915810754\\
66	0.601914797255585\\
67	0.602251636592288\\
68	0.602592272173893\\
69	0.602903850330673\\
70	0.603207902192367\\
71	0.607134201546025\\
72	0.608443314927463\\
73	0.608814567797361\\
74	0.60910753513026\\
75	0.60910753513026\\
76	0.60910753513026\\
77	0.60910753513026\\
78	0.61117069814412\\
79	0.612797064176606\\
80	0.613958125795662\\
81	0.617836145710911\\
82	0.618978297365964\\
83	0.619270973960589\\
84	0.619518734398045\\
85	0.619518734398045\\
86	0.619518734398045\\
87	0.619825412189901\\
88	0.620029757753001\\
89	0.620029757753001\\
90	0.620300665359116\\
91	0.622349755845151\\
92	0.623590359342274\\
93	0.624295910232902\\
94	0.624599445986255\\
95	0.624836521957103\\
96	0.624836521957103\\
97	0.624836521957103\\
98	0.624836521957103\\
99	0.624836521957103\\
100	0.624836521957103\\
101	0.626987258415915\\
102	0.627872125781118\\
103	0.628177502621266\\
104	0.628391682601252\\
105	0.628391682601252\\
106	0.628391682601252\\
107	0.628391682601252\\
108	0.628695102355908\\
109	0.628915622698572\\
110	0.628915622698572\\
111	0.630785749070773\\
112	0.631641893261461\\
113	0.63194235949725\\
114	0.632127501459858\\
115	0.632127501459858\\
116	0.632127501459858\\
117	0.632378484896768\\
118	0.63257403119683\\
119	0.63257403119683\\
120	0.63257403119683\\
121	0.634158624252188\\
122	0.635616238572041\\
123	0.637199055578844\\
124	0.637888573965436\\
125	0.638785743878365\\
126	0.639472963087232\\
127	0.639695826602702\\
128	0.639927992345117\\
129	0.64015498600614\\
130	0.640381048002595\\
131	0.642057016985063\\
132	0.642761596328311\\
133	0.64298423511142\\
134	0.643161666169326\\
135	0.643161666169326\\
136	0.643161666169326\\
137	0.643161666169326\\
138	0.643161666169326\\
139	0.643161666169326\\
140	0.643161666169326\\
141	0.644053553324671\\
142	0.644739810680809\\
143	0.644959441286586\\
144	0.645160713692651\\
145	0.645160713692651\\
146	0.645160713692651\\
147	0.645160713692651\\
148	0.645160713692651\\
149	0.645160713692651\\
150	0.645160713692651\\
151	0.646310390169649\\
152	0.647640100629886\\
153	0.649351011760029\\
154	0.650551104901049\\
155	0.651486680598064\\
156	0.651720807026237\\
157	0.651720807026237\\
158	0.65199649693783\\
159	0.652201978402906\\
160	0.652201978402906\\
161	0.653561637876734\\
162	0.654211804865509\\
163	0.654449915529415\\
164	0.65460130314783\\
165	0.65460130314783\\
166	0.65460130314783\\
167	0.654811182725221\\
168	0.654960837586058\\
169	0.654960837586058\\
170	0.655158112582397\\
171	0.656417041360327\\
172	0.656979341414815\\
173	0.657183314720091\\
174	0.657326878093356\\
175	0.657326878093356\\
176	0.657326878093356\\
177	0.657534237882897\\
178	0.657665406459221\\
179	0.657665406459221\\
180	0.657805393235026\\
181	0.658978916210281\\
182	0.660102539111441\\
183	0.660274422066421\\
184	0.660449548378857\\
185	0.660449548378857\\
186	0.660449548378857\\
187	0.660449548378857\\
188	0.660449548378857\\
189	0.660449548378857\\
190	0.660449548378857\\
191	0.661082654267613\\
192	0.661631683991932\\
193	0.661828722134732\\
194	0.661986055591923\\
195	0.664065995271543\\
196	0.665165115310394\\
197	0.666925692976748\\
198	0.66797154059218\\
199	0.668141426038896\\
200	0.668322292999826\\
201	0.669092676325756\\
202	0.669256355380226\\
203	0.669256355380226\\
204	0.67010323104701\\
205	0.67133950559144\\
206	0.671472548998319\\
207	0.673200893574409\\
208	0.673863545688833\\
209	0.674763208333821\\
210	0.674886345064658\\
211	0.6759564655097\\
212	0.676526170687822\\
213	0.676954651090811\\
214	0.676954651090811\\
215	0.677131009212359\\
216	0.677268819887383\\
217	0.677268819887383\\
218	0.677268819887383\\
219	0.677268819887383\\
220	0.677268819887383\\
221	0.677837192732016\\
222	0.678823695118581\\
223	0.679357387152245\\
224	0.679516132024627\\
225	0.679692090962238\\
226	0.679841985443924\\
227	0.679841985443924\\
228	0.679841985443924\\
229	0.679841985443924\\
230	0.679841985443924\\
231	0.68041322986131\\
232	0.680882613153526\\
233	0.681028840200634\\
234	0.681158244484879\\
235	0.681158244484879\\
236	0.681158244484879\\
237	0.681158244484879\\
238	0.682594716750496\\
239	0.683450853262408\\
240	0.685002943766177\\
241	0.685995334450578\\
242	0.686926291160399\\
243	0.68737387797444\\
244	0.687465942388589\\
245	0.687465942388589\\
246	0.687465942388589\\
247	0.687618108817235\\
248	0.687733076623013\\
249	0.687733076623013\\
250	0.687733076623013\\
251	0.688580936794342\\
252	0.689383743971721\\
253	0.689930536205227\\
254	0.690361431397177\\
255	0.690361431397177\\
256	0.690361431397177\\
257	0.69092407982681\\
258	0.69104102825757\\
259	0.69104102825757\\
260	0.691198391362682\\
261	0.69202875560361\\
262	0.692435219185982\\
263	0.692574450037376\\
264	0.692671858893507\\
265	0.692671858893507\\
266	0.692671858893507\\
267	0.692797492818175\\
268	0.692919055159834\\
269	0.692919055159834\\
270	0.693029831558456\\
271	0.693801430449093\\
272	0.694574007082712\\
273	0.694729116291975\\
274	0.694829693507178\\
275	0.694829693507178\\
276	0.694829693507178\\
277	0.694978948643627\\
278	0.69509731402674\\
279	0.69509731402674\\
280	0.695236735776178\\
281	0.695963044138242\\
282	0.696696729490513\\
283	0.696812648995676\\
284	0.696812648995676\\
285	0.696812648995676\\
286	0.696812648995676\\
287	0.698161059021293\\
288	0.698524992858826\\
289	0.699109135287757\\
290	0.699452953715952\\
291	0.700243655214566\\
292	0.701030238064712\\
293	0.70116244210508\\
294	0.701262467449261\\
295	0.701262467449261\\
296	0.701262467449261\\
297	0.701391771324456\\
298	0.701496425046743\\
299	0.701496425046743\\
300	0.701496425046743\\
};
\end{axis}
\end{tikzpicture}%

%% file: images/sensitivityPSNR.tikz
%
%
\definecolor{mycolor1}{rgb}{0.00000,0.44700,0.74100}%
\definecolor{mycolor2}{rgb}{0.85000,0.32500,0.09800}%
\definecolor{mycolor3}{rgb}{0.92900,0.69400,0.12500}%
\definecolor{mycolor4}{rgb}{0.49400,0.18400,0.55600}%
\begin{tikzpicture}

\begin{axis}[%
width=0.953\figurewidth,
height=\figureheight,
at={(0\figurewidth,0\figureheight)},
scale only axis,
unbounded coords=jump,
xmode=log,
xmin=0.01,
xmax=10,
xminorticks=true,
xlabel style={font=\color{white!15!black}},
xlabel={$\lambda$},
ymin=24.1,
ymax=24.6,
ylabel style={font=\color{white!15!black}},
ylabel={PSNR},
axis background/.style={fill=white},
legend style={legend cell align=left, align=left, draw=white!15!black},
xmajorgrids,
ymajorgrids,
ylabel near ticks,
xlabel near ticks
]
\addplot [color=black, line width=0.8pt]
  table[row sep=crcr]{%
0.01	24.2159654409486\\
0.0109749876549306	24.2218200074333\\
0.0120450354025878	24.2275826099011\\
0.0132194114846603	24.2339516161031\\
0.0145082877849594	24.2407893655533\\
0.0159228279334109	24.2480740854975\\
0.0174752840000768	24.2554299565388\\
0.0191791026167249	24.2638012588702\\
0.0210490414451202	24.2721788605138\\
0.0231012970008316	24.280733434989\\
0.0253536449397011	24.2900015630831\\
0.0278255940220712	24.2995502738948\\
0.0305385550883342	24.309097620587\\
0.0335160265093884	24.3187754566239\\
0.0367837977182863	24.3281408317911\\
0.0403701725859655	24.339243888109\\
0.0443062145758388	24.34958157051\\
0.0486260158006535	24.3588233405306\\
0.0533669923120631	24.3686658066364\\
0.0585702081805667	24.3778382658043\\
0.0642807311728432	24.3869641439027\\
0.0705480231071865	24.3952885912987\\
0.0774263682681127	24.4023664026085\\
0.0849753435908644	24.4092957313587\\
0.093260334688322	24.4147756900043\\
};

\addplot [color=blue, line width=0.8pt]
  table[row sep=crcr]{%
0.01	24.2225776946514\\
0.0109749876549306	24.2279767884906\\
0.0120450354025878	24.234169680893\\
0.0132194114846603	24.2408943881386\\
0.0145082877849594	24.2470617238873\\
0.0159228279334109	24.2539935401784\\
0.0174752840000768	24.2624584716537\\
0.0191791026167249	24.2698382360992\\
0.0210490414451202	24.2791361168734\\
0.0231012970008316	24.2875894567163\\
0.0253536449397011	24.2969052406557\\
0.0278255940220712	24.3070051476818\\
0.0305385550883342	24.3178005763415\\
0.0335160265093884	24.3280420833098\\
0.0367837977182863	24.3389766402134\\
0.0403701725859655	24.3503693910355\\
0.0443062145758388	24.3605690132702\\
0.0486260158006535	24.3708895472784\\
0.0533669923120631	24.3834594355629\\
0.0585702081805667	24.3919725792419\\
0.0642807311728432	24.4028493918011\\
0.0705480231071865	24.4134198258495\\
0.0774263682681127	24.4229844298988\\
0.0849753435908644	24.4307084302559\\
0.093260334688322	24.4388865133674\\
0.102353102189903	24.4468375317484\\
0.112332403297803	24.451575640214\\
0.123284673944207	24.4588729669171\\
0.135304777457981	24.4615168286869\\
0.148496826225446	24.4635316299508\\
0.162975083462064	24.4673667578677\\
0.178864952905744	24.4663291058708\\
0.196304065004027	24.4693307508904\\
0.215443469003188	24.4705989643377\\
0.236448941264541	24.4704259945759\\
0.259502421139974	24.4708175949936\\
0.28480358684358	24.4678678272701\\
0.312571584968824	24.4685108240815\\
0.343046928631492	24.469420980152\\
0.376493580679247	24.4694234654512\\
0.413201240011534	24.4703863109039\\
0.453487850812858	24.470485296241\\
0.497702356433211	24.4700455186903\\
0.546227721768434	24.4704370832086\\
0.599484250318941	24.4707605273715\\
0.657933224657568	24.4707605273715\\
0.722080901838546	24.4707605273715\\
0.792482898353917	24.4707605273715\\
0.869749002617783	24.4707605273715\\
0.954548456661834	24.4707605273715\\
};

\addplot [color=red, line width=0.8pt]
  table[row sep=crcr]{%
0.01	24.254135795904\\
0.0109749876549306	24.2590294629431\\
0.0120450354025878	24.2650237468222\\
0.0132194114846603	24.2698663248768\\
0.0145082877849594	24.2761321184136\\
0.0159228279334109	24.2831088842563\\
0.0174752840000768	24.2899436963439\\
0.0191791026167249	24.2970703265946\\
0.0210490414451202	24.3050735828503\\
0.0231012970008316	24.3140571539823\\
0.0253536449397011	24.3219101628055\\
0.0278255940220712	24.331106529716\\
0.0305385550883342	24.3395960913471\\
0.0335160265093884	24.3499128004017\\
0.0367837977182863	24.359592770338\\
0.0403701725859655	24.3692440226871\\
0.0443062145758388	24.3799097776398\\
0.0486260158006535	24.3904872643763\\
0.0533669923120631	24.4016734378877\\
0.0585702081805667	24.4115040216646\\
0.0642807311728432	24.4214798629716\\
0.0705480231071865	24.433168649394\\
0.0774263682681127	24.4430559828197\\
0.0849753435908644	24.4520774786527\\
0.093260334688322	24.4595550371212\\
0.102353102189903	24.4681467678129\\
0.112332403297803	24.4759531481803\\
0.123284673944207	24.4828440790817\\
0.135304777457981	24.4887913311042\\
0.148496826225446	24.4926459273918\\
0.162975083462064	24.4978155627062\\
0.178864952905744	24.4983272478998\\
0.196304065004027	24.5001510093509\\
0.215443469003188	24.504099124945\\
0.236448941264541	24.5046051328345\\
0.259502421139974	24.5075442224765\\
0.28480358684358	24.5035811882853\\
0.312571584968824	24.506298580073\\
0.343046928631492	24.5074779894558\\
0.376493580679247	24.5062118126569\\
0.413201240011534	24.5073803637428\\
0.453487850812858	24.5066487092705\\
0.497702356433211	24.5081613744453\\
0.546227721768434	24.5082612250556\\
0.599484250318941	24.5074013257133\\
0.657933224657568	24.50836683431\\
0.722080901838546	24.5076128653048\\
0.792482898353917	24.5087815385661\\
0.869749002617783	24.5095318299443\\
0.954548456661834	24.5086517115904\\
1.04761575278967	24.5089794101246\\
1.14975699539774	24.5074797028562\\
1.26185688306602	24.5082211440461\\
1.38488637139387	24.507732868883\\
1.51991108295293	24.5080618290149\\
1.66810053720006	24.5080618290149\\
1.83073828029537	24.5080618290149\\
2.00923300256505	24.5080618290149\\
2.20513073990305	24.5080618290149\\
2.42012826479438	24.5080618290149\\
2.65608778294669	24.5080618290149\\
2.91505306282518	24.5080618290149\\
3.19926713779738	24.5080618290149\\
3.51119173421513	24.5080618290149\\
3.85352859371053	24.5080618290149\\
4.2292428743895	24.5080618290149\\
4.64158883361278	24.5080618290149\\
5.09413801481638	24.5080618290149\\
5.59081018251222	24.5080618290149\\
6.13590727341318	24.5080618290149\\
6.73415065775082	24.5080618290149\\
7.39072203352578	24.5080618290149\\
8.11130830789687	24.5080618290149\\
8.90215085445038	24.5080618290149\\
9.77009957299226	24.5080618290149\\
};

\end{axis}
\end{tikzpicture}%

%% file: images/sensitivitySSIM.tikz
%
%
\definecolor{mycolor1}{rgb}{0.00000,0.44700,0.74100}%
\definecolor{mycolor2}{rgb}{0.85000,0.32500,0.09800}%
\definecolor{mycolor3}{rgb}{0.92900,0.69400,0.12500}%
\definecolor{mycolor4}{rgb}{0.49400,0.18400,0.55600}%
\begin{tikzpicture}

\begin{axis}[%
width=0.953\figurewidth,
height=\figureheight,
at={(0\figurewidth,0\figureheight)},
scale only axis,
unbounded coords=jump,
xmode=log,
xmin=0.01,
xmax=10,
xminorticks=true,
xlabel style={font=\color{white!15!black}},
xlabel={$\lambda$},
ymin=0.58,
ymax=0.64,
ylabel style={font=\color{white!15!black}},
ylabel={SSIM},
axis background/.style={fill=white},
legend style={legend cell align=left, align=left, draw=white!15!black},
xmajorgrids,
ymajorgrids,
ylabel near ticks,
xlabel near ticks
]
\addplot [color=black, line width=0.8pt]
  table[row sep=crcr]{%
0.01	0.59042318971894\\
0.0109749876549306	0.591235812953366\\
0.0120450354025878	0.592027822827794\\
0.0132194114846603	0.592945119788789\\
0.0145082877849594	0.593919298727702\\
0.0159228279334109	0.594922302605279\\
0.0174752840000768	0.595920217429549\\
0.0191791026167249	0.597131232919769\\
0.0210490414451202	0.598297289035\\
0.0231012970008316	0.599470952558107\\
0.0253536449397011	0.600788930231828\\
0.0278255940220712	0.602067316395504\\
0.0305385550883342	0.603385481856075\\
0.0335160265093884	0.604651693999641\\
0.0367837977182863	0.605932283917195\\
0.0403701725859655	0.607480557730663\\
0.0443062145758388	0.608811302900449\\
0.0486260158006535	0.61008344192868\\
0.0533669923120631	0.611217345710682\\
0.0585702081805667	0.612410467715341\\
0.0642807311728432	0.613503278283383\\
0.0705480231071865	0.614469561857174\\
0.0774263682681127	0.615363526062935\\
0.0849753435908644	0.616208237212637\\
0.093260334688322	0.616766608106399\\
};

\addplot [color=blue, line width=0.8pt]
  table[row sep=crcr]{%
0.01	0.591310241961135\\
0.0109749876549306	0.592065730494898\\
0.0120450354025878	0.592937485194066\\
0.0132194114846603	0.593852780777407\\
0.0145082877849594	0.594744518463205\\
0.0159228279334109	0.595657825210977\\
0.0174752840000768	0.596833201842461\\
0.0191791026167249	0.597932127690322\\
0.0210490414451202	0.599187010415394\\
0.0231012970008316	0.600357391255563\\
0.0253536449397011	0.601613039703952\\
0.0278255940220712	0.603044829987727\\
0.0305385550883342	0.604514254097194\\
0.0335160265093884	0.605925231930356\\
0.0367837977182863	0.607318679525982\\
0.0403701725859655	0.608986128415986\\
0.0443062145758388	0.610144271023834\\
0.0486260158006535	0.611428885673662\\
0.0533669923120631	0.61315534847406\\
0.0585702081805667	0.614262190754223\\
0.0642807311728432	0.61567608319176\\
0.0705480231071865	0.616944713956435\\
0.0774263682681127	0.61816130148645\\
0.0849753435908644	0.619167917016755\\
0.093260334688322	0.619824664210875\\
0.102353102189903	0.62083717613628\\
0.112332403297803	0.621398808793095\\
0.123284673944207	0.622107184638397\\
0.135304777457981	0.622306987396336\\
0.148496826225446	0.622482889404866\\
0.162975083462064	0.622756131157089\\
0.178864952905744	0.622659757368226\\
0.196304065004027	0.623083035556624\\
0.215443469003188	0.623167560692171\\
0.236448941264541	0.623061936736099\\
0.259502421139974	0.623028605821838\\
0.28480358684358	0.622647368997952\\
0.312571584968824	0.622655443037461\\
0.343046928631492	0.622736234203865\\
0.376493580679247	0.622866411042311\\
0.413201240011534	0.622915825892388\\
0.453487850812858	0.622951470328754\\
0.497702356433211	0.622800447097357\\
0.546227721768434	0.622895558832893\\
0.599484250318941	0.622931541593303\\
0.657933224657568	0.622931541593303\\
0.722080901838546	0.622931541593303\\
0.792482898353917	0.622931541593303\\
0.869749002617783	0.622931541593303\\
0.954548456661834	0.622931541593303\\
};

\addplot [color=red, line width=0.8pt]
  table[row sep=crcr]{%
0.01	0.595823854912878\\
0.0109749876549306	0.596443730236157\\
0.0120450354025878	0.597191485990828\\
0.0132194114846603	0.59794778352735\\
0.0145082877849594	0.59875348708076\\
0.0159228279334109	0.599782062686897\\
0.0174752840000768	0.600716646774427\\
0.0191791026167249	0.601831582494737\\
0.0210490414451202	0.602770006923975\\
0.0231012970008316	0.604046751572616\\
0.0253536449397011	0.60515304231258\\
0.0278255940220712	0.606277246089782\\
0.0305385550883342	0.607495520236969\\
0.0335160265093884	0.608929806453024\\
0.0367837977182863	0.610265851183848\\
0.0403701725859655	0.611449007066829\\
0.0443062145758388	0.612913097021712\\
0.0486260158006535	0.614257740493664\\
0.0533669923120631	0.615778836313786\\
0.0585702081805667	0.617166927170291\\
0.0642807311728432	0.618380200326639\\
0.0705480231071865	0.620054089756727\\
0.0774263682681127	0.62114397073116\\
0.0849753435908644	0.62214528994721\\
0.093260334688322	0.62305234732713\\
0.102353102189903	0.624288016508926\\
0.112332403297803	0.62512376683998\\
0.123284673944207	0.625821891397354\\
0.135304777457981	0.626489365786891\\
0.148496826225446	0.626768293368972\\
0.162975083462064	0.62734297181593\\
0.178864952905744	0.627158332638061\\
0.196304065004027	0.627457060278111\\
0.215443469003188	0.628021236417554\\
0.236448941264541	0.628090102908767\\
0.259502421139974	0.628431712778275\\
0.28480358684358	0.62793947311082\\
0.312571584968824	0.628358563152099\\
0.343046928631492	0.628501920167349\\
0.376493580679247	0.628394436842729\\
0.413201240011534	0.628355679347693\\
0.453487850812858	0.6283314394028\\
0.497702356433211	0.628460331942952\\
0.546227721768434	0.628718436258349\\
0.599484250318941	0.6282620490526\\
0.657933224657568	0.628559790303189\\
0.722080901838546	0.628633317384976\\
0.792482898353917	0.628364144260642\\
0.869749002617783	0.628306093254884\\
0.954548456661834	0.628701552529998\\
1.04761575278967	0.62829807300807\\
1.14975699539774	0.628097065508168\\
1.26185688306602	0.628288845819571\\
1.38488637139387	0.628333224977709\\
1.51991108295293	0.628325168203897\\
1.66810053720006	0.628325168203897\\
1.83073828029537	0.628325168203897\\
2.00923300256505	0.628325168203897\\
2.20513073990305	0.628325168203897\\
2.42012826479438	0.628325168203897\\
2.65608778294669	0.628325168203897\\
2.91505306282518	0.628325168203897\\
3.19926713779738	0.628325168203897\\
3.51119173421513	0.628325168203897\\
3.85352859371053	0.628325168203897\\
4.2292428743895	0.628325168203897\\
4.64158883361278	0.628325168203897\\
5.09413801481638	0.628325168203897\\
5.59081018251222	0.628325168203897\\
6.13590727341318	0.628325168203897\\
6.73415065775082	0.628325168203897\\
7.39072203352578	0.628325168203897\\
8.11130830789687	0.628325168203897\\
8.90215085445038	0.628325168203897\\
9.77009957299226	0.628325168203897\\
};

\end{axis}
\end{tikzpicture}%